%% file: thesis.tex
\def\degree{Doctor of Philosophy}
\def\department{School of Computer Science \\ Faculty of Engineering}
\def\mydegrees{}

\def\supervisora{Sarvnaz Karimi}
\def\supervisorb{Ben Hachey}
\def\supervisorc{Cecile Paris}
\def\supervisord{Joachim Gudmundsson}

\documentclass[logo]{style/usydthesis}

\newcommand*{\noaddvspace}{\renewcommand*{\addvspace}[1]{}}
\addtocontents{lof}{\protect\noaddvspace}
\usepackage{amstext,amssymb,amsfonts,latexsym}
\usepackage[utf8]{inputenc}
\usepackage[UKenglish]{babel}
\usepackage{csquotes}
\usepackage{ccicons}
\usepackage{animate}
\usepackage{movie15}
\usepackage{float,lscape}
\usepackage{afterpage,rotating}

\usepackage{amsmath,amssymb,bm,booktabs,caption,graphicx,multirow,natbib,soul,subcaption,xspace}
\usepackage[inline]{enumitem}

\newcommand{\CONLL}{\textsc{CoNLL}\xspace}
\newcommand{\GLOVE}{\textsc{GloVe}\xspace}
\newcommand{\ELMO}{\textsc{ELMo}\xspace}
\newcommand{\FLAIR}{\textsc{Flair}\xspace}
\newcommand{\NCBIDISEASE}{\textsc{NCBI}-\textsc{Disease}\xspace}
\newcommand{\ONTONOTES}{\textsc{OntoNotes}\xspace}
\newcommand{\POS}{\textsc{PoS}\xspace}

\newcommand{\SHARECLEF}{\textsc{ShARe/CLEF}\xspace}
\newcommand{\SQUAD}{\textsc{SQuAD}\xspace}
\newcommand{\WORDNET}{\textsc{WordNet}\xspace}
\DeclareMathOperator*{\argmax}{arg\,max}
\newcommand{\dmatric}[1]{\bm{\uppercase{#1}}}
\newcommand{\dvector}[1]{\bm{#1}}

\newcommand{\drandvari}[1]{\textnormal{#1}}
\newcommand{\dscalar}[1]{#1}
\newcommand{\dset}[1]{\mathbb{#1}}

\newcommand{\bilstm}{\textsc{BiLSTM}\xspace}
\newcommand{\crf}{\textsc{CRF}\xspace}
\newcommand{\ffnn}{\textsc{FFNN}\xspace}
\newcommand{\lstm}{\textsc{LSTM}\xspace}

\newcommand{\softmax}{\textsc{SoftMax}\xspace}
\newcommand{\dtanh}{\textsc{Tanh}\xspace}
\usepackage[colorinlistoftodos]{todonotes}

\usepackage{setspace}
\usepackage{pdfpages}
\usepackage[pdftex,bookmarks=true]{hyperref}
\hypersetup{
    pdfauthor = {Xiang Dai},
    pdftitle = {Recognising Biomedical Names: Challenges and Solutions},
    colorlinks,
    linkcolor={black},
    citecolor={black}
    }

\RequirePackage{natbib}
\usepackage{microtype}

\begin{document}

\renewcommand{\thepage}{\roman{page}}	

\title{{\bf\Huge Recognising Biomedical Names: Challenges and Solutions}}
\author{Xiang Dai}
\date{2021}

\maketitle
\setstretch{1.5}

\input{src/ch0-statement.tex}

\input{src/ch0-abstract.tex}
\input{src/ch0-thanks.tex}

\input{src/ch0-authorship.tex}

\setcounter{tocdepth}{2}
\newpage
\addcontentsline{toc}{chapter}{Contents}
\tableofcontents
\listoffigures
\listoftables

\setcounter{page}{1}
\setcounter{chapter}{0}

\renewcommand{\thepage}{\arabic{page}}	
\setupParagraphs

\input{src/ch0-abbreviations.tex}


\chapter{Introduction~\label{chapter-introduction}}
\input{src/ch1-introduction.tex}


\chapter{Literature Review~\label{chapter-literature-review}}
\input{src/ch3-literature-review.tex}

\chapter{Data Augmentation for NER~\label{chapter-data-augmntation}}
\input{src/ch4-data-augmentation.tex}

\chapter{Cost-effective Selection of Pre-training Data~\label{chapter-select-pretraining-data}}
\input{src/ch5-select-pretraining-data.tex}

\chapter{Transition-based Model for Discontinuous NER~\label{chapter-discontinuous-ner}}
\input{src/ch6-discontinuous-ner.tex}

\chapter{Conclusions~\label{chapter-conclusion}}
\input{src/ch7-conclusion.tex}

\bibliography{thesis}
\bibliographystyle{acl_natbib}

\end{document}

%% file: src/ch0-statement.tex
\chapter*{Statement of Originality}

This is to certify that to the best of my knowledge, the content of this thesis is my own work. This thesis has not been submitted for any degree or other purposes.

I certify that the intellectual content of this thesis is the product of my own work and that all the assistance received in preparing this thesis and sources have been acknowledged.

\vspace{3cm}

Xiang Dai 

\par

2021-Feb-27

%% file: src/ch0-abstract.tex
\chapter*{Abstract}
The growth rate in the amount of biomedical documents---such as scholarly articles, clinical notes and health forum discussions---is staggering. Unlocking information trapped in these documents can enable researchers and practitioners to operate confidently in the information world. Biomedical Information Extraction (IE) system aims to automatically extract structured information---such as biomedical concepts, attributes, events, and their relations---from unstructured text. Within an IE system, the first step is called Biomedical Named Entity Recognition (NER), the task of recognising biomedical names.

NER has been heavily studied in the generic domain, recognising person, organisation, and location names in newspaper articles. However, the effectiveness of existing Biomedical NER model is still not satisfactory. In contrast to entity mentions in the generic domain which are usually short spans of text, biomedical names---surface forms that represent biomedical concepts, such as genes, proteins, symptoms, diseases, and drugs---pose unique challenges. For example, it is even common for an ordinary person to confuse \emph{`severe acute respiratory syndrome coronavirus 2'} (virus name), \emph{`severe acute respiratory syndrome'} (disease name), and \emph{`coronavirus disease 2019'} (disease name). The variety of language used for different communicative purposes makes biomedical NER even more challenging. Various groups of people use totally different languages to describe the same biomedical concept. For example, researchers tend to use standard names in biomedical vocabularies to make the description more comprehensible and less confused; hospital doctors, who write notes under time pressure, use abbreviations for efficient communication with their colleagues; and, ordinary people use linguistically noisy layman language to share their experiences. 

State-of-the-art NER models, based on sequence tagging technique, are good at recognising short entity mentions in the generic domain, especially when they are enhanced by pre-trained language representation models. However, there are several open challenges of applying these models to recognise biomedical names: 
\begin{itemize}
    \item Biomedical names may contain complex inner structure (discontinuity and overlapping) which cannot be recognised using standard sequence tagging technique;
    \item The training of NER models usually requires large amount of labelled data, which are difficult to obtain in the biomedical domain; and,
    \item Commonly used language representation models are pre-trained on generic data, such as the Wikipedia and books, a domain shift therefore exists between these models and target biomedical data.
\end{itemize}

To deal with these challenges, we explore several research directions and make the following contributions: (1) we propose a transition-based NER model which can recognise discontinuous mentions. Through experiments on three datasets from the biomedical domain, we show that our model can effectively recognise discontinuous entity mentions without sacrificing the accuracy on continuous mentions. Analysis also suggests that our model is good at recognising long mentions, resulting in higher recall than other baselines; (2) We develop a cost-effective approach that nominates the suitable pre-training data, via measuring the similarity between different pre-training data options and target task data. Through experiments on 56 source-target data pairs, we show that simple similarity measures are good predictors of the usefulness of pre-trained language representation models on downstream NER datasets; and, (3) We design several data augmentation methods which do not rely on any external trained models, for NER. Experimental results show that the proposed augmentation methods can improve performance over strong baselines, where large scale pre-trained language representation models are used.

Our contributions have obvious practical implications, especially when new biomedical applications are needed. Our proposed data augmentation methods can help the NER model achieve decent performance, requiring only a small amount of labelled data. Our investigation regarding selecting pre-training data can improve the model by incorporating language representation models, which are pre-trained using in-domain data. Finally, our proposed transition-based NER model can further improve the performance by recognising discontinuous mentions without sacrificing the accuracy on continuous mentions.

%% file: src/ch0-thanks.tex

\chapter*{Acknowledgements}
I enjoyed my journey of doing a PhD, and I am immensely thankful for the many people I met during this journey. Foremost thanks must go to my PhD supervisors Sarvnaz Karimi, Ben Hachey and Cecile Paris. Sarvnaz, thank you for your encouragement, without which I would never have started working on NLP. Also thank you for your detailed guidance, which shapes my research. Ben, thank you for your completely honest criticisms and feedback, which always help me revisit my work from a practical perspective. Cecile, thank you for sensible advice and patience. I learn a lot from people like you who want to make things to be perfect. I also want to thank my supervisor Joachim Gudmundsson, who helped me a lot with university's administration and funding.

I am very thankful to Dietrich Klakow from Saarland University. Because of the COVID-$19$ pandemic, I was stranded in Germany after I finished my internship at Bosch Center for Artificial Intelligence. Dietrich hosted me in his group, providing me a shelter where I could finish my thesis. Thanks also to Heike Adel, Matthew Honnibal, and Vera Demberg for their help.

I also want to thank my colleagues at Data61. In particular, I am grateful to Aditya Joshi, Chang Xu, Maciej Rybinski, Vincent Nguyen, Stephen Wan, Sunghwan Mac Kim, Wenyi Tay, and Sonit Singh for regular reading group meetings, and inspiring discussions. Thanks also to Lukas Lange and Michael A. Hedderich from Saarland University for feedback on drafts of this thesis, and exchange of ideas.

Finally, thank you to my family for being encouraging and patient. Special thanks to my dad, who explains me a lot about medications and medical procedures.

%% file: src/ch0-authorship.tex
\chapter*{Authorship Attribution Statement}
\begin{itemize}
    \item The Chapter~\ref{chapter-introduction} of this thesis relates to~\citep{Dai:Karimi:ALTA:2017}.
    
    \textbf{Xiang Dai}, Sarvnaz Karimi, and Cecile Paris. 2017. Medication and adverse event extraction from noisy text. In Proceedings of the Australasian Language Technology Association Workshop, pages 79–87, Brisbane, Australia.

    \item The Section~\ref{section-complex-ner} of this thesis relates to~\citep{Dai:ACL-SRW:2018}.
    
    \textbf{Xiang Dai}. 2018. Recognizing complex entity mentions: A review and future directions. Proceedings of ACL 2018, Student Research Workshop, pages 37–44, Melbourne, Australia.
    
    \item Chapter 3 of this thesis relates to~\citep{Dai:Adel:COLING:2020}. 
    
    \textbf{Xiang Dai}, Heike Adel. 2020. An Analysis of Simple Data Augmentation for Named Entity Recognition. In Proceedings of the 28th International Conference on Computational Linguistics, Online.
    
    \item Chapter 4 of this thesis relates to~\citep{Dai:Karimi:NAACL:2019,Dai:Karimi:EMNLP:2020}.
    
    \textbf{Xiang Dai}, Sarvnaz Karimi, Ben Hachey, and Cecile Paris. 2019. Using similarity measures to select pretraining data for NER. In Proceedings of the 2019 Conference of the North American Chapter of the Association for Computational Linguistics: Human Language Technologies, Volume 1 (Long and Short Papers), pages 1460–1470, Minneapolis, Minnesota. 
    
    \textbf{Xiang Dai}, Sarvnaz Karimi, Ben Hachey, and Cecile Paris. 2020. Cost-effective Selection of Pretraining Data: A Case Study of Pretraining BERT on Social Media. In Findings of the 2020 Conference on Empirical Methods in Natural Language Processing, Online.
    
    \item Chapter 5 of this thesis relates to~\citep{Dai:Karimi:ACL:2020}.
    
    \textbf{Xiang Dai}, Sarvnaz Karimi, Ben Hachey, and Cecile Paris. 2020. An effective transition-based model for discontinuous NER. In Proceedings of the 58th Annual Meeting of the Association for Computational Linguistics, pages 5860--5870, Online.
\end{itemize}

I contribute to every aspect of the above mentioned publications, including designing the method, conducting the experiment, and writing the paper.

%% file: src/ch0-abbreviations.tex
\chapter*{Notations}

\begin{table}[h]
	\centering
	\begin{tabular}{p{0.3\linewidth} p{0.5\linewidth}}
	$\dscalar{a}$ & A scalar \\
	$\dvector{a}$ & A vector \\
	$\dmatric{a}$ & A matrix \\
	$\dset{A}$ & A set \\
	\\
	$P(\drandvari{a})$ & A probability distribution over a discrete variable \\
	$f(x; \bm \theta)$ & A function of x parameterised by $\bm \theta$ \\
	\\
	$\bm 1_\mathrm{condition}$ & is 1 if the condition is true, 0 otherwise \\
	$\oplus$ & Concatenating two vectors \\
	\\
	$\{t_i\}_{i=1}^N$ & A sequence of $N$ elements, such as tokens or vectors \\
	\end{tabular}
\end{table}

\chapter*{Abbreviations}

\begin{table}[h!]
\centering
\begin{tabular}{p{0.3\linewidth} p{0.5\linewidth}}
	\\ \\
	\textbf{cf.} & \textbf{c}on\textbf{f}er/\textbf{c}on\textbf{f}eratur (compare) \\ \\
	\textbf{e.g.} & \textbf{e}xemplum \textbf{g}ratia/conferatur (example) \\ \\
	\textbf{et al.} & \textbf{et al}ia (and others) \\ \\
	\textbf{etc.} & \textbf{et c}etera (and so on) \\ \\
	\textbf{i.e.} & \textbf{i}d \textbf{e}st (that is) \\ \\
\end{tabular}
\end{table}

\begin{table}[h!]
\centering
\begin{tabular}{p{0.3\linewidth} p{0.5\linewidth}}
	\\ 	\\
	\textbf{ADE} & \textbf{A}dverse \textbf{D}rug \textbf{E}vent \\ \\
	\textbf{ADR} & \textbf{A}dverse \textbf{D}rug \textbf{R}eaction \\ \\
	\textbf{BERT} & \textbf{B}idirectional \textbf{E}ncoder \textbf{R}epresentations from \textbf{T}ransformers \\ \\
	\textbf{BIO} & \textbf{B}eginning-\textbf{I}nside-\textbf{O}utside \\ \\
	\textbf{EHR} & \textbf{E}lectronic \textbf{H}ealth \textbf{R}ecord \\ \\
	\textbf{ELECTRA} & \textbf{E}fficiently \textbf{L}earning an \textbf{E}ncoder that \textbf{C}lassifies \textbf{T}oken \textbf{R}eplacements \textbf{A}ccurately \\ \\
	\textbf{ELMo} & \textbf{E}mbeddings from \textbf{L}anguage \textbf{Mo}dels \\ \\
	\textbf{IE} & \textbf{I}nformation \textbf{E}xtraction \\ \\
	\textbf{LSTM} & \textbf{L}ong \textbf{S}hort-\textbf{T}erm \textbf{M}emory \\ \\
	\textbf{LM} & \textbf{L}anguage \textbf{M}odel \\ \\
	\textbf{NER} & \textbf{N}amed \textbf{E}ntity \textbf{R}ecognition \\ \\
	\textbf{NLP} & \textbf{N}atural \textbf{L}anguage \textbf{P}rocessing \\ \\
	\textbf{UMLS} & \textbf{U}nified \textbf{M}edical \textbf{L}anguage \textbf{S}ystem \\ \\
\end{tabular}
\end{table}

%% file: src/ch1-introduction.tex
The growth rate in the amount of biomedical documents---such as scholarly articles, clinical notes and health forum discussions---is staggering. Unlocking information trapped in these documents can enable researchers and practitioners to operate confidently in the information world. Biomedical Information Extraction (IE) system aims to automatically extract structured information---such as biomedical concepts, attributes, events, and their relations---from unstructured text. Within an IE system, the first step is called Biomedical Named Entity Recognition (NER), the task of recognising biomedical names.

In this chapter, we first use a real world application---recognising adverse drug events from social media---as an example, to illustrate how a NER model can be used to extract useful information (Section~\ref{section-introduction-example-applicatoin}). Next, we describe a unified architecture for the most popular sequence tagging based NER models, dividing sequence taggers into two components: (1) a mapping function that maps each token to a feature vector, and, (2) a classifier that predicts a sequence of tags given the input sequence of feature vectors (Section~\ref{section-background-unified-architecture}). Then, in Section~\ref{section-introduction-key-challenges}, we identify three open challenges of applying state-of-the-art sequence taggers, enhanced by pre-trained language representation models, to recognise biomedical names: (1) complex structures---overlapping and discontinuity---occur often in biomedical names; (2) the training of sequence taggers requires large training sets which are usually difficult to obtain in the biomedical domain; and, (3) there is a discrepancy between publicly available language representation models pre-trained on generic data and target biomedical data. To deal with these challenges, we explore different research directions and make the following contributions: we propose a transition-based model for discontinuous NER; we develop a cost-effective approach that nominates the suitable pre-training data; and we design several data augmentation methods for NER (Section~\ref{section-introduction-about-the-thesis}).

\section{Recognising Adverse Drug Events from Social Media---A Motivating Application}
\label{section-introduction-example-applicatoin}
An \emph{Adverse Drug Reaction} (ADR) is an injury occurring after a drug is used at the recommended dosage, for recommended symptoms~\citep{Karimi:Wang:Survey:2015}. Detecting ADRs as early as possible can potentially have a major impact, because ADRs are among the leading causes of death in many countries, and ADR-related costs have exceeded the cost of medications~\citep{WHO:ADR:2020}. \citet{Bonn:1998,Hadi:Neoh:Pharm:2017}; and \citet{Khalil:Huang:Health:2020} estimate that ADRs account for more than $100,000$ deaths per year in the United States, and $197,000$ deaths annually in Europe. The situation in developing countries may be more severe. For example, \citet{Mouton:Mehta:Pharm:2015} estimate that in South Africa ADRs contribute to the death of $2.9\%$ of medical admissions, and $16\%$ of deaths are ADR-related.

Different from controlled clinical trials which are mainly conducted \textit{before} drugs are licensed for use, pharmacovigilance---the practice of monitoring the ADRs of pharmaceutical products---focuses on identifying previously unreported adverse reactions \textit{after} the drugs are marketed. Establishing causality---whether the given adverse reaction is caused by the drug---is often done by domain experts. Causality assessment needs to investigate the statistical association in laboratory parameters and exclude other causes, such as alcohol, disease-related causes, other drugs and so on~\citep{Anderson:Borlak:2011}. Surveillance systems, both passive and active, play an important role in collecting potential \emph{Adverse Drug Events} (ADEs). Note that, when causality between an adverse reaction and a drug is not known, it is referred to as an adverse drug event.

\emph{Passive surveillance} of ADEs relies on spontaneous reporting systems which allow health professionals and patients to voluntarily report observed or suspected ADEs to regulatory agencies. For example, the MedWatch system has been built by the Food and Drug Administration since the early $1990$s~\citep{Piazza-Hepp:Kennedy:1995}. However, under-reporting is severe. Studies estimate that more than $90\%$ of ADEs are not reported to these systems due to various obstacles, such as lack of suspicions, lack of information about reporting utility, lack of time, and difficulties in filling out forms~\citep{Vallano:Cereza:2005,Hazell:Shakir:2006}. 

\emph{Active surveillance}, in contrast, aims to discover ADEs automatically from multiple sources, including Electronic Health Records (EHRs), medical literature, search engine logs, and even social media. Since such information is often trapped in free text representation, IE systems can be used to extract information of interest. NER is usually employed at the very beginning of the IE system. The sentence, represented as a sequence of tokens, is taken as input of the NER model, and entity mentions, each of which is represented as a set of token positions, are outputted. In addition, one entity category, such as drug, disease, symptom, ADE and so on, is assigned to each entity mention. 

\paragraph{A simple example}
In this section, we describe a simple example of a post from a patient forum and explain how the NER model recognises biomedical names. 

\begin{figure}[tb] 
	\centering
	\includegraphics[width=0.85\textwidth]{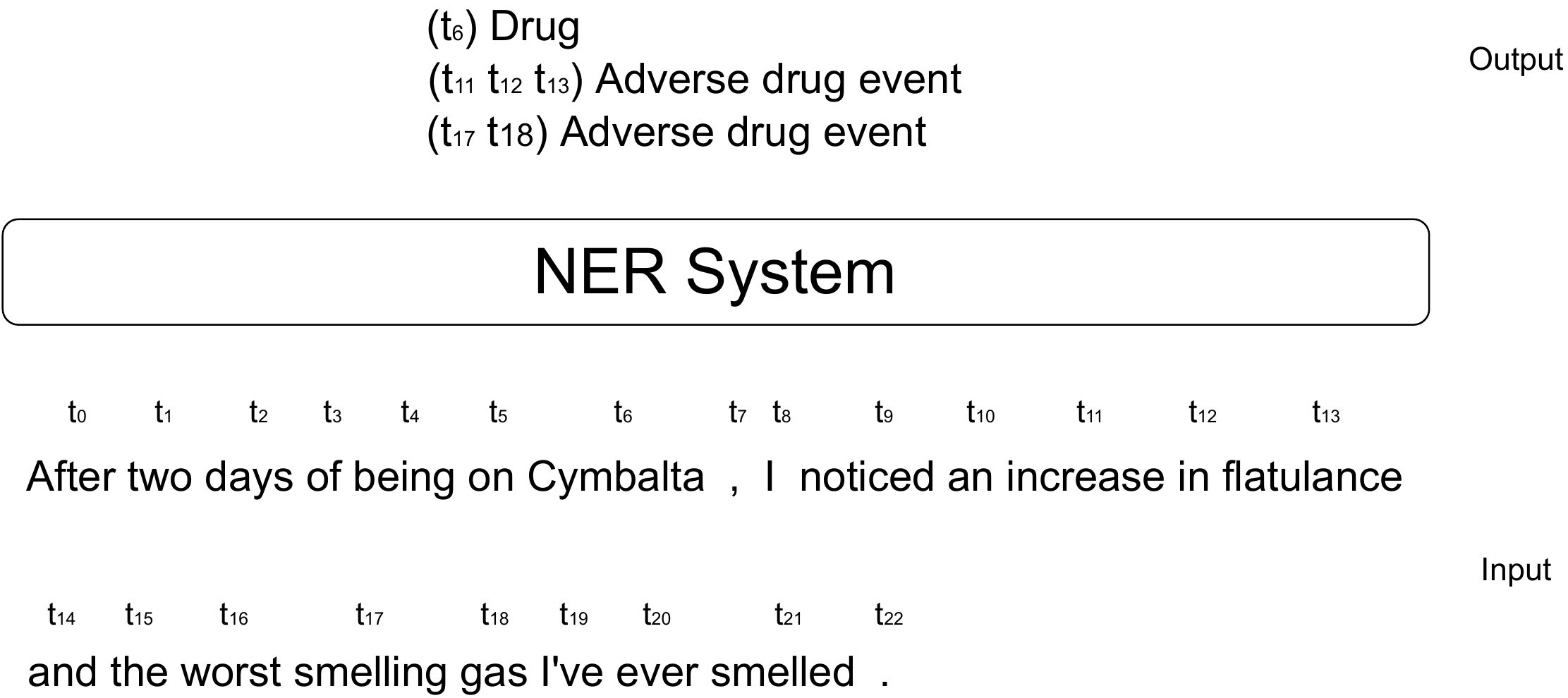}
	\caption{An example input sentence and the entity mentions which are supposed to be recognised by the NER model.~\label{figure-background-example-sentence}}
\end{figure}

Given a sequence of tokens: 
\begin{quote}
	After two days of being on Cymbalta , I noticed an increase in flatulance\footnote{The spelling error is from the original post.} and the worst smelling gas I've ever smelled .
\end{quote}
the NER model is supposed to recognise three entity mentions: \textit{`Cymbalta'}, as a drug mention, \textit{`increase in flatulance'} and \textit{`smelling gas'}, as ADEs (Figure~\ref{figure-background-example-sentence}).

\begin{figure}[tb] 
	\centering
	\includegraphics[width=0.8\textwidth]{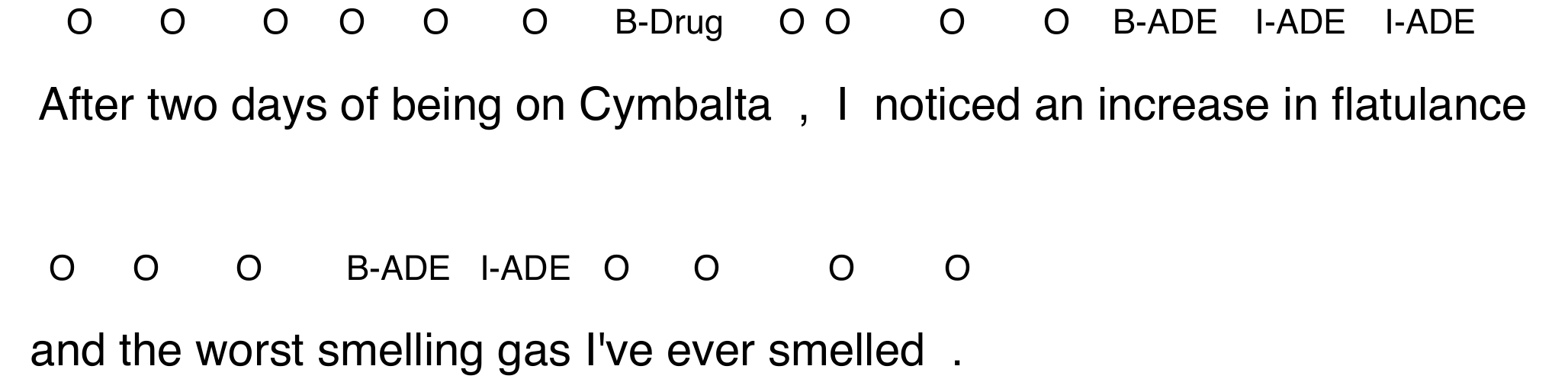}
	\caption{State-of-the-art NER model is based on sequence tagging technique that assigns a tag to each token. Token positions of mentions can be extracted from the output tag sequence.~\label{figure-background-bio-example}}
\end{figure}

\paragraph{Sequence tagging based NER model}
The state-of-the-art NER model is based on sequence tagging technique that assigns a tag to each token. The tag is usually composed of a position indicator and an entity category. The position indicator is used to represent the token’s role in a mention. For example, in the BIO schema~\citep{Sang:Meulder:CONLL:2003}, B stands for the Beginning of a mention, I for the Inside of a mention, and O for Outside a mention. Figure~\ref{figure-background-bio-example} is an example of input sequence of tokens and the corresponding output sequence of tags. Taking the token \textit{`smelling'} as an example, its tag `B-ADE' indicates that the token is the beginning token of an ADE mention. 

Once the sequence of tags is outputted, token positions of mentions can be extracted from the tag sequence via finding all sub tag sequences starting with `B-$\star$' tag, and including all succeeding `I-$\star$' tags. Put another way, for each token whose tag starts with `B', there is a mention starting at this token position, and ending before the next token position where the corresponding tag is `O' or starts with `B'. Note that it is possible for the sequence tagger to predict an invalid sequence of tags, for example, a tag `B-ADE' followed by a tag `I-Drug'. Therefore, post-processing steps, such as changing the tag's position indicator `I' to `B' if its entity category is different from the preceding category, are usually employed before the tag sequence is decoded into mentions. 

\subsection{A unified architecture for sequence taggers}
\label{section-background-unified-architecture}

\begin{figure}[tb] 
	\centering
	\includegraphics[width=0.75\textwidth]{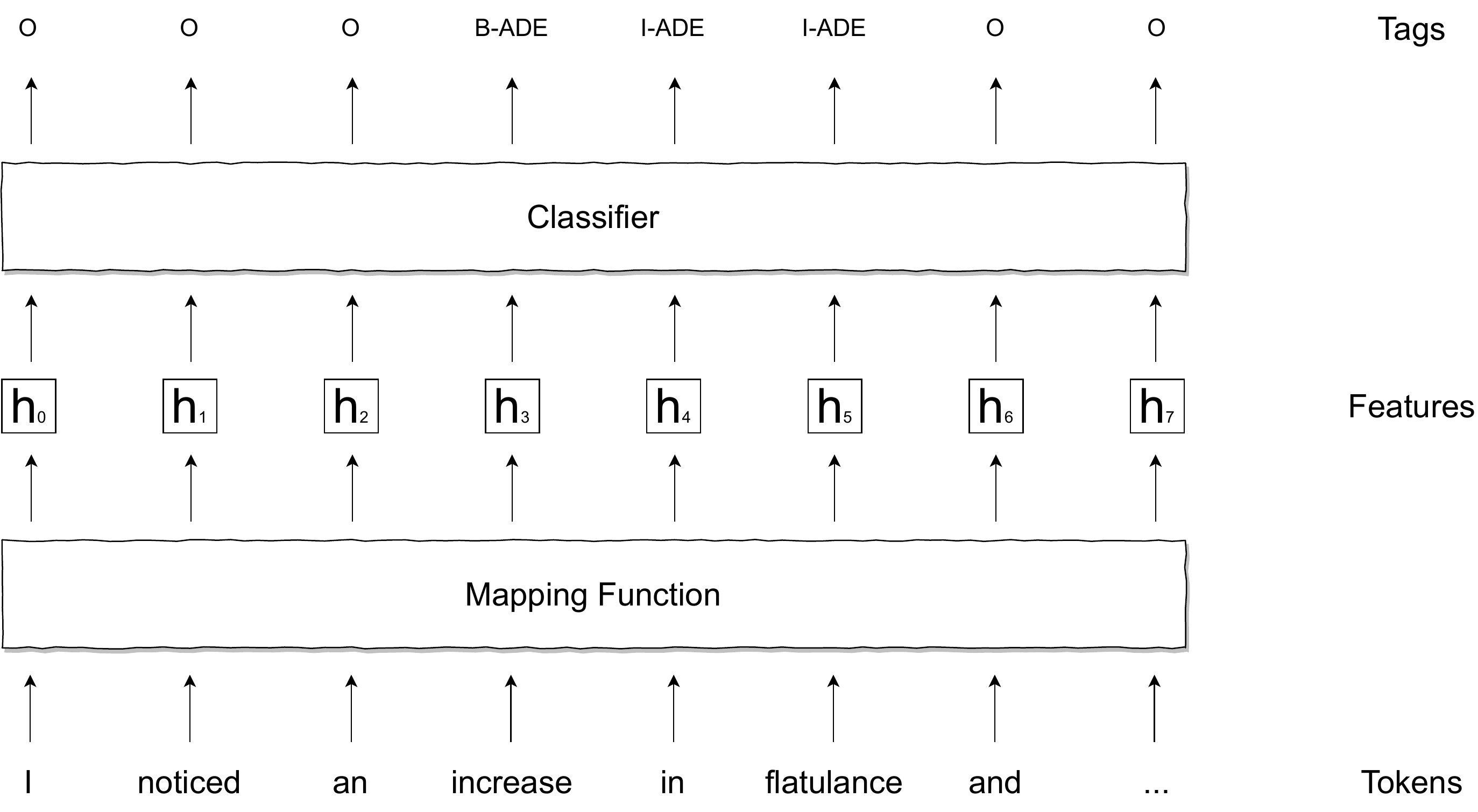}
	\caption{A unified architecture for sequence tagging models, consisting of a mapping function and a classifier.~\label{figure-background-ner}}
\end{figure}

In general, sequence taggers can be divided into two components (Figure~\ref{figure-background-ner}): 
\begin{description}
	\item[Mapping function] it converts the input sequence of tokens into a sequence of features vectors, each of which represents the corresponding token-in-the-context; and,
	\item[Classifier] it predicts a sequence of tags given the input sequence of feature vectors.
\end{description}

The key to supervised machine learning based techniques is optimising the model so that they can fit the labelled training data. In other words, for each training instance consisting of the input sequence of tokens and the output sequence of tags, the tagger aims to predict a sequence of tags as close as possible to the grounding truth. The main advantage of recent deep learning based techniques over conventional feature based machine learning techniques is that the former optimises mapping function and the classifier jointly, whereas the mapping function in the latter is usually handcrafted and fixed during the model training stage. 

Current state-of-the-art approaches~\citep{Huang:Xu:arXiv:2015,Lample:Ballesteros:NAACL:2016,Ma:Hovy:ACL:2016,Yang:Liang:COLING:2018} for sequence tagging use the Bidirectional Long Short-Term Memory (\bilstm) as the mapping function, and a subsequent linear-chain Conditional Random Field (\crf) as the classifier. \FLAIR~\citep{Akbik:Blythe:COLING:2018} is a variant of \bilstm-\crf sequence tagger, which achieves the state-of-the-art performance in multiple sequence tagging datasets, including the \CONLL2003 English and German NER datasets. In this section, we detail its components in a top-down manner.

\paragraph{Linear-chain \crf}
Given a sequence of feature vectors: $\{\dvector{h}_i\}_{i=1}^n$, each of which representing a token-in-the-context, the simplest classifier can take each feature vector as input and makes the prediction independently. That is, for each feature vector at position $\dscalar{i}$, 
\begin{equation} 
\dvector{o}_i = \mathrm{softmax} (\dmatric{W} \dvector{h}_i + \dvector{b}).
\end{equation}
This classifier ignores the relationship between neighbouring tags. For example, if the tag at a position is \textit{`B-ADE'} (beginning token of an ADE), it is impossible for the succeeding tag to be \textit{`I-Drug'} (inside token of a drug name), because the tag \textit{`I-Drug'} should always follow a \textit{`B-Drug'} (beginning token of a drug name) or another \textit{`I-Drug'}. 

CRF is a classifier that predicts the output sequence jointly, taking the dependency between neighbouring outputs into consideration. That is, it aims to predict a sequence of tags $\dmatric{\hat{O}} = \{\dvector{o}_i\}_{i=1}^n$ which has the maximum probability over all possible tag sequences: 

\begin{equation} 
\label{equation-background-crf-best-sequence}
\dmatric{\hat{O}} = \argmax_{\dmatric{O}} \, P \, (\dmatric{O}\mid\dmatric{H}),
\end{equation}
where 
\begin{equation} 
P \, (\dmatric{O}\mid\dmatric{H}) \propto \prod_{i=1}^n \psi(\dvector{o}_{i-1}, \dvector{o}_i, \dvector{h}_i)
\end{equation}
and 
\begin{equation} 
\label{equation-background-crf}
\psi(\dvector{o}_i, \dvector{o}_j, \dvector{h}) = \exp \, ( \dmatric{W} \dvector{h} + \dmatric{A}_{\dvector{o}_i, \dvector{o}_j} ).
\end{equation}
In Equation~\ref{equation-background-crf}, $\dmatric{A}_{i,j}$ is the compatibility score of a transition from the tag i to tag j. 

\begin{figure}[tb] 
	\centering
	\includegraphics[width=0.75\textwidth]{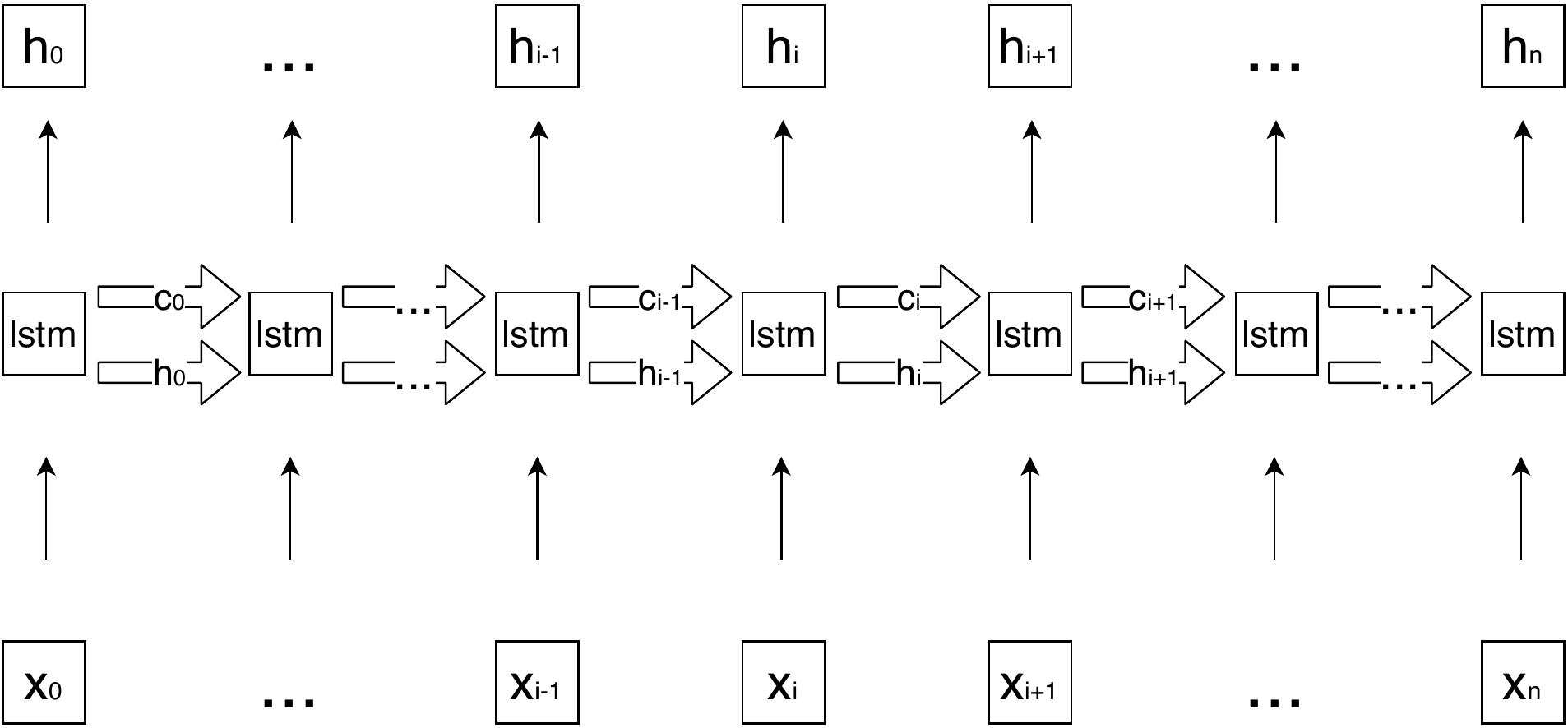}
	\caption{An \lstm computes the output state by taking the entire past (left context) of the input sequence into consideration.~\label{figure-background-lstm}}
\end{figure}

\paragraph{Token level \bilstm layer}
The \lstm variant of recurrent neural networks~\citep{Hochreiter:Schmidhuber:1997,Graves:Mohamed:ICASSP:2013} is widely used by recent work to create the contextual representation, due to its ability to flexibly encode long-term dependencies via a memory cell. In the \lstm architecture (Figure~\ref{figure-background-lstm}), the output state at each position ($\dvector{h}_i)$ is computed by taking the input at the current position ($\dvector{x}_i)$ as well as the hidden state and cell state from the previous position ($\dvector{h}_{i-1}$ and $\dvector{c}_{i-1}$, respectively) into consideration: \begin{equation} 
\label{equation-background-forward-lstm}
	\dvector{h}_i = f(\dvector{t}_i,\dvector{h}_{i-1}, \dvector{c}_{i-1}; \bm \theta).
\end{equation}

By computing recursively the hidden state, the entire past history---left context---of each position is incorporated. The term \emph{bidirectional} indicates that there are two models---forward and backward models---used to capture both left and right contexts. The backward model works in the same way but in the reversed direction: 
\begin{equation} 
\label{equation-background-backward-lstm}
\dvector{h}_i = f(\dvector{t}_i,\dvector{h}_{i+1}, \dvector{c}_{i+1}; \bm \theta).
\end{equation}

In the following, we use the superscript $f$ to define states relating to the forward model and $b$ to the backward model. 
For example, $\dvector{h}_i^f$ indicates the contextual representation of the $i$-th token from the output in Equation~\ref{equation-background-forward-lstm} and $\dvector{h}_i^b$ the contextual representation of the $i$-th token obtained from output in Equation~\ref{equation-background-backward-lstm}. 
A convention of employing \bilstm is that the final contextual representation at each position $\dvector{h}_i$---the contextual representation of the $i$-th token---is usually extracted by concatenating the hidden states for each position from both forward and backward models: 
\begin{equation} 
\dvector{h}_i = \begin{bmatrix} 
\dvector{h}_i^f \oplus \dvector{h}_i^b
\end{bmatrix}.
\end{equation}

\paragraph{Contextual string embeddings}
\FLAIR introduces a novel type of token embeddings based on character level encoder. The input token is first treated as a sequence of characters. Table~\ref{table-background-contextual-string-embeddings} is an example sequence of tokens and the corresponding character level start indices. 

\begin{table}[tb] 
	\centering
	\begin{tabular}{r | c c c c c c c c}
		\toprule
		& I & noticed & an & increase & in & flatulance & and & ... \\
		Token level indices & 0 & 1 & 2 & 3 & 4 & 5 & 6 & 7 \\
		Character level start indices & 0 & 2 & 10 & 13 & 22 & 25 & 37 & 41 \\
		\bottomrule
	\end{tabular}
	\caption{The sequence of tokens is treated as a sequence of characters.~\label{table-background-contextual-string-embeddings}}
\end{table}

\begin{figure}[tb] 
	\centering
	\includegraphics[width=0.95\textwidth]{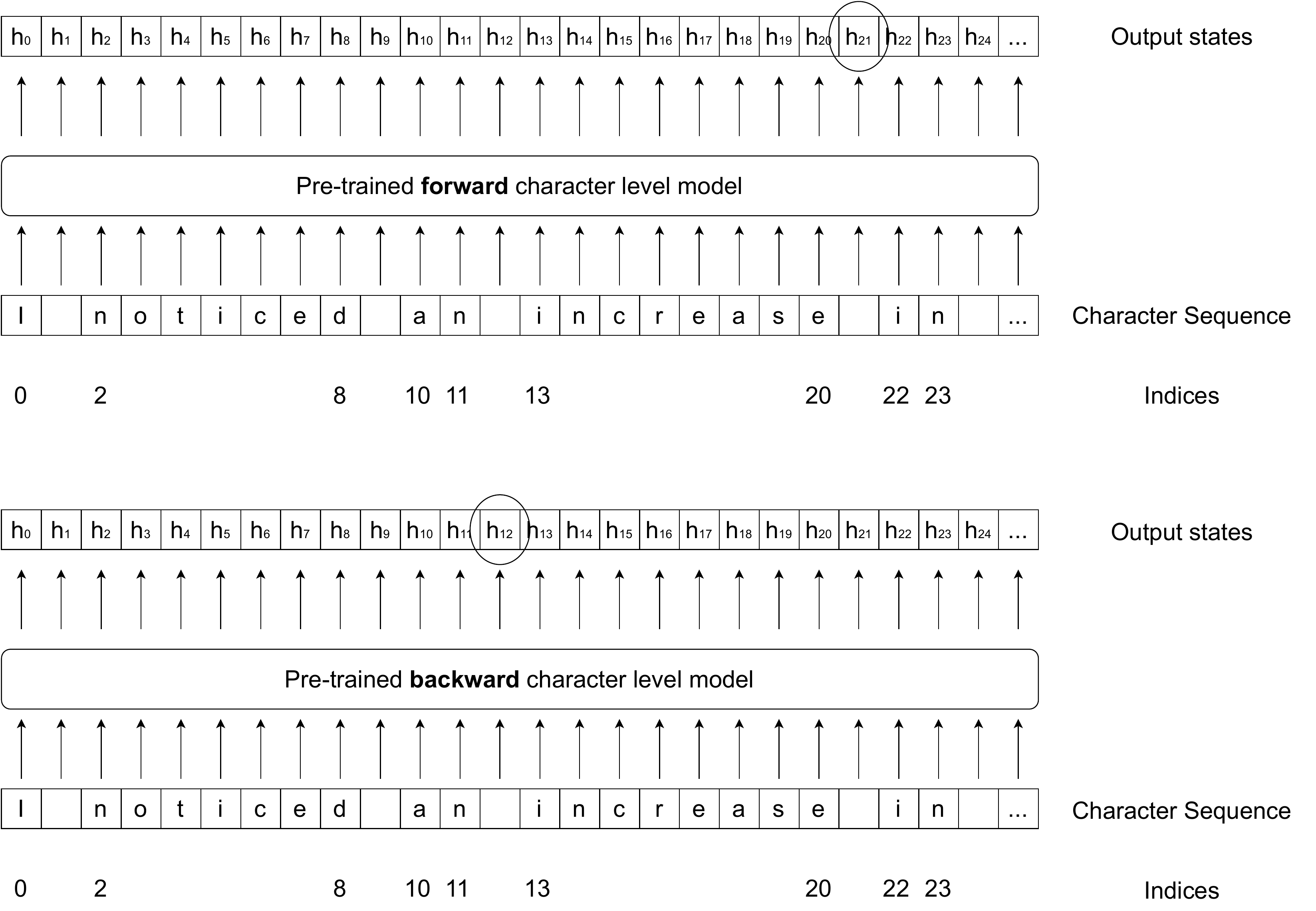}
	\caption{An forward model computes the output state by taking the entire past (left context) of the input sequence into consideration, whereas the backward model considers the entire right context. Both output hidden states are concatenated to form the final contextual string embedding and capture the information of the token itself as well as its surrounding tokens.~\label{figure-background-char-lm}}
\end{figure}

Then the sequence of characters is taken as input of two---forward and backward---pre-trained character level \bilstm models. The final contextual string embeddings for each token can be extracted by concatenating outputs from these two \bilstm models. Specifically, for the forward model, the output hidden state after the last character in the token is used, and the output hidden state before the token's first character from the backward model is used. 
Taking the token \textit{`increase'} in Table~\ref{table-background-contextual-string-embeddings} as an example, the output state of the $12$-th character from the backward model and the output state of the $21$-st character from the forward model are concatenated as the contextual string embedding (Figure~\ref{figure-background-char-lm}). 

Finally, the stacking embeddings, a concatenation of contextual string embedding and pre-computed \GLOVE embedding~\citep{Pennington:Socher:EMNLP:2014}, are used as the final token embedding and taken as input to the previous described token-level \bilstm layer. 

\paragraph{Pre-trained language representation models}
In contrast to supervised machine learning that optimises model using labelled data only, semi-supervised learning aims to make use of both labelled data and unlabelled data. Pre-training language representation models on unlabelled data and then adapting pre-trained model to the downstream supervised task is one type of semi-supervised learning. It has demonstrated its effectiveness in NLP during the past decade~\citep{Mikolov:Sutskever:NIPS:2013,Dai:Le:NIPS:2015,Howard:Ruder:ACL:2018,Peters:Neumann:NAACL:2018,Devlin:Chang:NAACL:2019}. In this section, we briefly describe the design in \FLAIR and more options are discussed in Section~\ref{section-embeddings-pretrain-tasks}.

\begin{figure}[p] 
	\centering
	\includegraphics[width=0.95\textwidth]{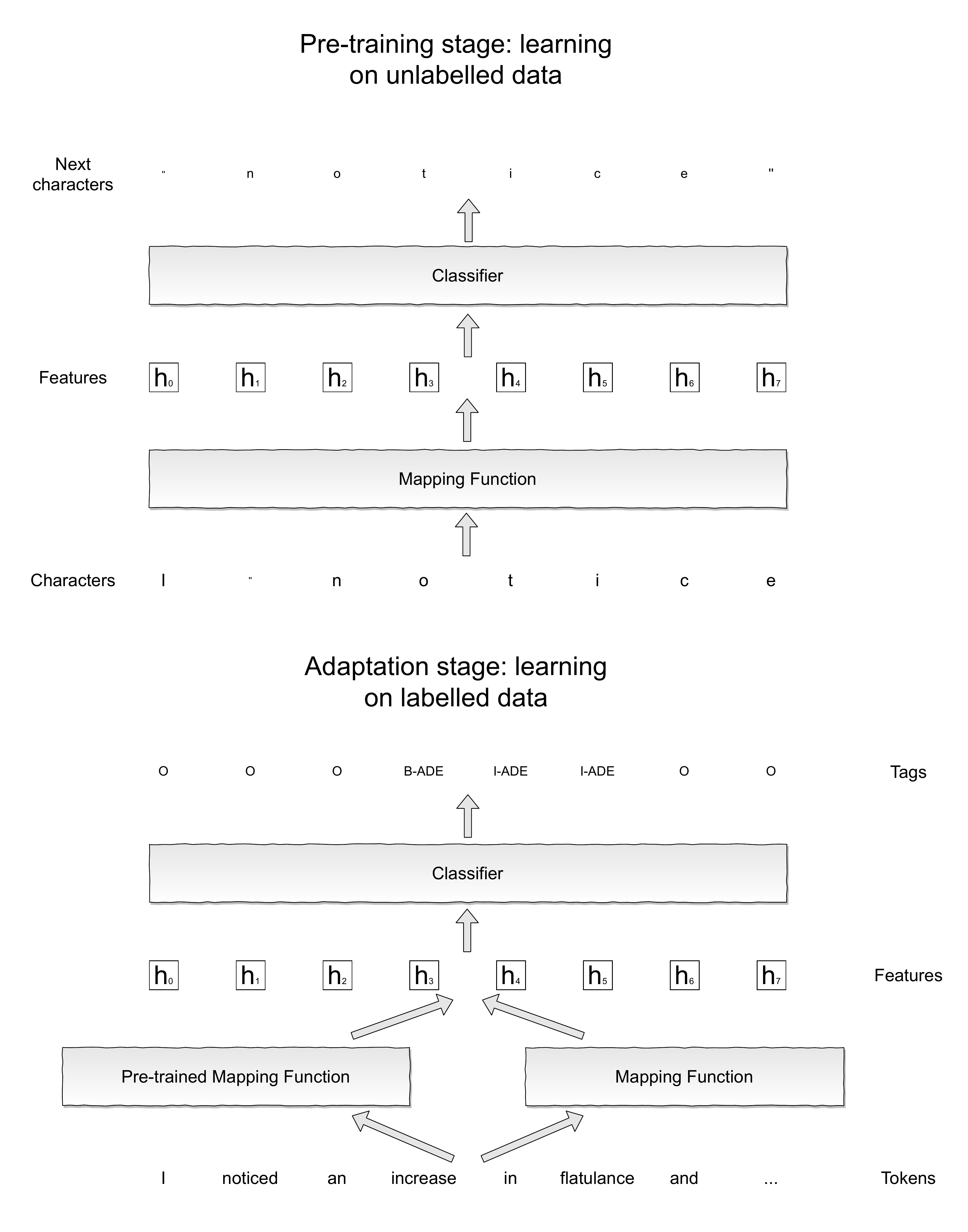}
	\caption{The semi-supervised learning approach used in \FLAIR: pre-training two---forward and backward---character level language models, and using the pre-trained model as part of the mapping function in the downstream supervised tasks. For the sake of brevity, we show only the forward model and the backward model is omitted. The whitespace character is represented using \textit{''} in this figure.~\label{figure-background-flair-transfer}}
\end{figure}

During the pre-training stage, \citet{Akbik:Blythe:COLING:2018} train two separate models---forward and backward---on the 1-billion word corpus~\citep{Chelba:Mikolov:arXiv:2013}. The pre-training task is a standard character level language modelling task that predicts the next character given a sequence of characters. Taking the backward model illustrated in Figure~\ref{figure-background-char-lm} as an example, the output state of the $12$-nd character is taken as input to a classifier to predict the next character, whose ground truth in this example is the character \textit{`n'}. Once the pre-training finishes, the pre-trained models are frozen and used as part of the mapping function for downstream supervised task (Figure~\ref{figure-background-flair-transfer}).

\section{Key Open Challenges}
\label{section-introduction-key-challenges}
Although \FLAIR, enhanced by pre-trained language representation models, has achieved state-of-the-art performance in multiple NER datasets in the generic domain, we find there are several open challenges of applying \FLAIR to recognise biomedical names. 

\subsection{Biomedical names are complex}
Different from entity mentions in the generic domain, which are usually short spans of text, biomedical names may contain more complex inner structure. 
Considering the following sequence of tokens: 
\begin{quote} 
	have much muscle pain and fatigue .
\end{quote}
it contains two biomedical names, \textit{`muscle pain'} and \textit{`muscle fatigue'}, that share the token \textit{`muscle'}. 
In this case, we call them \emph{overlapping} biomedical names. 
In addition, \textit{`muscle fatigue'} is a \emph{discontinuous} mention, consisting of two components that are separated from each other. 

The main motivation for recognising these biomedical names with complex inner structure is that they usually represent compositional concepts that differ from concepts represented by individual components. 
Specifically, each of these two names in the example sentence---\textit{`muscle pain'} and \textit{`muscle fatigue'}---describes a disorder which has its own CUI (Concept Unique Identifier) in UMLS (Unified Medical Language System), whereas \textit{`muscle'}, \textit{`pain'}, and \textit{`fatigue'} also have their own CUIs. 
In downstream applications, such as pharmacovigilance, extracting these compositional concepts, such as symptoms or ADEs, is often more useful than extracting individual components which may refer to body locations or general feelings. 

\subsection{Labelling data is difficult}
A well known limitation of deep neural models is that training these models usually requires large amount of labelled data~\citep{LeCun:Bengio:Nature:2015}. In other words, the advantages of deep learning may diminish when working with small training sets. For example, \citet{Shen:Yun:ICLR:2018} observe that a deep neural model outperforms the best shallow model by absolute $F_1$ score of $2.2$, when a large NER training set---\ONTONOTES 5.0, containing more than $1$ million tokens ---is available. In contrast, this advantage becomes only $0.4$, when training on a comparatively small training set (\CONLL 2003, containing around $0.2$ million tokens). 

Labelling large amount of generic NER data is time-consuming, because the annotation needs to be done at the token level. 
Labelling large amount of biomedical NER data is even more difficult due to the following reasons: 
\begin{itemize}
	\item The previously described complex structure increases the difficulty of annotation. Standard NER annotation is usually done at the token level: annotators need to scrutinise every token to decide whether it is part of \emph{one} entity mention. 
	However, due to the complex structure---overlapping and discontinuity---in biomedical names, one token may belong to multiple biomedical names, and tokens that are far away from each other may form one biomedical name. 
	Exhaustive enumeration of possible names, including discontinuous and overlapping ones, is exponential to sentence length. 
	
	\item Domain knowledge is required to annotate biomedical NER datasets. Different from the task of annotating generic entity mentions, such as person names or locations, with which ordinary people are familiar, recognising biomedical names, such as biological substances or disorders, requires the annotators to have at least basic domain knowledge. 
	
	Worse still, the same entity category may have subtle meanings in different biomedical applications.	This may even require annotators to have expert level knowledge in a specific application. For example, \emph{family history extraction} is a task that focuses on the detection of family history related disorders. Therefore, it also pays attention to some behaviour patterns which may be caused by genetic factors, and these behaviour patterns are usually overlooked by popular disorder recognition tasks~\citep{rybinski-2021-jmir-family-history}. In other words, a behaviour pattern is usually not defined as a disorder of interest in most of biomedical applications, but needs to be labelled as disorder in family history extraction application, once the behaviour---such as a pattern of alcohol use---may put people health at risk, and it may be influenced by genetic factors. 
	
	
	\item Some annotation tasks in the biomedical domain may cause negative impacts on annotators. For example, annotators may feel uncomfortable after continuing annotating online posts about adverse drug events for a long time. These posts are written by patients, containing complains about their sufferings after drug usage. Proper protective arrangements need to be made to protect the annotators, and they usually lead to longer annotation task duration. 
	
	\item The last, but not the least, reason relates to the cost-benefit analysis widely used in project management activities. That is, a project for building biomedical applications usually starts from defining target performance specifications, and then estimates the cost of achieving the target performance. 
	Labelling training data is often the most expensive part of the project, and, unfortunately, we do not have practical methods to estimate how much training data is required to achieve the target performance~\citep{Johnson:Anderson:ACL:2018}. 
	So a more practical strategy is that domain experts usually first annotate a small set of training data, on which NLP practitioners need to build pilot models. 
	After the persuasive results are obtained using limited amount of training data, domain experts and project managers are more likely to commit more resources to create more labelled training data. 
\end{itemize}


\subsection{Unlabelled biomedical data are limited}
The main strengths of \FLAIR come from the use of stacking embeddings, that consist of two types of embeddings: pre-trained \GLOVE embedding~\citep{Pennington:Socher:EMNLP:2014} and contextual string embeddings based on pre-trained language models. \citet{Akbik:Blythe:COLING:2018} show that the use of pre-trained \GLOVE embedding increases average $F_1$ score by $1.1$, and the use of contextual string embeddings brings even larger improvements, around $4.5$ absolute $F_1$ score on the English NER dataset. However, both of these two types of embeddings are pre-trained on generic data. For example, \GLOVE embeddings are pre-trained on the English Wikipedia and Gigaword dataset (archive of news stories). These generic data usually have very different characteristics from the biomedical data. 

\begin{table}[tb] 
	\centering
	\begin{tabular}{c | c | c}
		\toprule
		& \CONLL 2003 & \SHARECLEF 2013 \\
		\midrule
		\multirow{2}{*}{\GLOVE (Vocabulary coverage)} & 87.6 \% & 37.2 \% \\ 
		& 18,415 / 21,089 & 5,282 / 14,172 \\ \hline
		String embeddings (Perplexity) & 7.746 & 29.839 \\
		\bottomrule
	\end{tabular}
	\caption{Discrepancy between the pre-trained model used in \FLAIR and the target datasets: \SHARECLEF 2013 (clinical notes) and \CONLL 2003 (news stories).~\label{table-background-discrepancy}}
\end{table}

Pre-trained models used in \FLAIR usually have sub-optimal performances on biomedical datasets, such as \SHARECLEF 2013, which is sourced from clinical notes.

We measure this discrepancy between the pre-trained model and the target data using two measures: vocabulary coverage and perplexity. Vocabulary coverage indicates the ratio of target data's vocabulary existing in the pre-trained model. For example, there is only 37.2 \% of \SHARECLEF 2013's vocabulary covered by \GLOVE (Table~\ref{table-background-discrepancy}). Perplexity is a way of evaluating the language model, which is used to generate contextual string embeddings in \FLAIR. The pre-trained character level language models achieve higher perplexity---a measurement of how well a language model predicts a test sentence---on \SHARECLEF 2013. Note that high perplexity indicates the language model is bad at predicting the test sentence, assigning low probability. The result suggests that there is a higher discrepancy between pre-trained models and the \SHARECLEF 2013 than \CONLL 2003, which is sourced from news stories.

Unfortunately, the access to unlabelled data in the biomedical domain can be restricted due to privacy and regulatory reasons. Documents with privacy sensitive contents, such as electronic health records, are usually available only after applying anonymisation operations. For example, the Health Insurance Portability and Accountability Act (HIPAA) of the United States defines that $18$ types of Protected Health Information (PHI), such as patient names, ages, phone numbers etc., need to be removed from the documents before they can be shared with third parties. Selecting proper pre-training data which are large enough and also similar to target task data is a non-trivial problem.


%

\section{About the Thesis}
\label{section-introduction-about-the-thesis}

\begin{figure}[tb]
\centering
\includegraphics[width=0.7\textwidth]{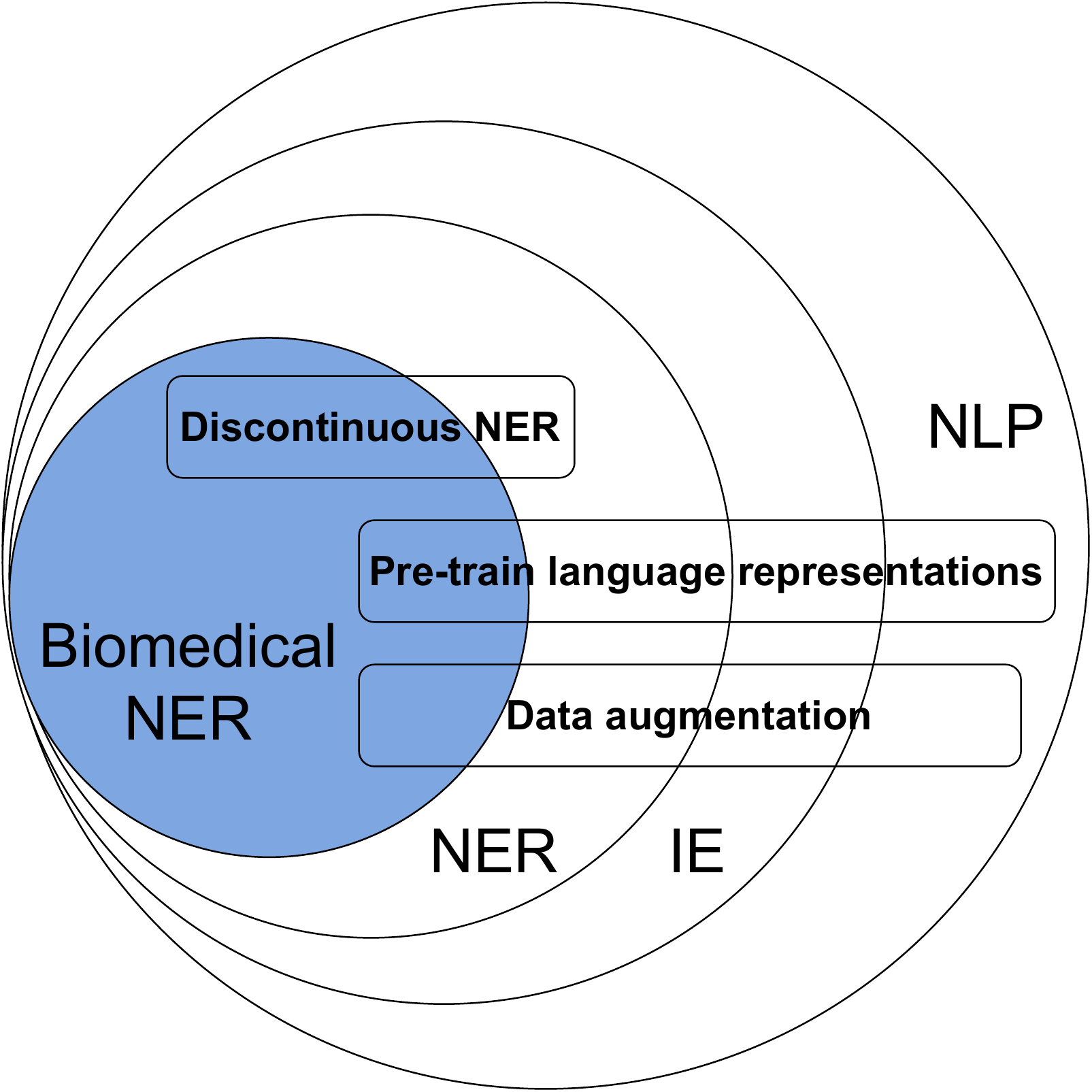}
\caption{We focus on Biomedical NER, and explore the following three research directions: recognising discontinuous entity mentions; pre-training domain-specific language representation models; and enhancing the effectiveness of NER models using data augmentation.\label{figure-introduction-scope}}
\end{figure}

To deal with these open challenges, we explore the corresponding research directions, aiming to improve the Biomedical NER. Figure~\ref{figure-introduction-scope} is a high-level overview of concepts we cover in this thesis. Although we focus on Biomedical NER in this thesis, some of these contributions, including proposed discontinuous NER models and new discoveries regarding the selection of pre-training data, can be potentially applied to other NLP tasks in other domains. Also it is worthy noting that we consider English text only.

\subsection{Publications}
The work in the thesis primarily relates to the following peer-reviewed articles (sorted by publication date):

\begin{enumerate}
	\item \textbf{Xiang Dai}, Sarvnaz Karimi, and Cecile Paris. 2017. Medication and adverse event extraction from noisy text. In Proceedings of the Australasian Language Technology Association Workshop, pages 79–87, Brisbane, Australia. (Chapter 1)
	
	\item \textbf{Xiang Dai}. 2018. Recognizing complex entity mentions: A review and future directions. Proceedings of ACL 2018, Student Research Workshop, pages 37–44, Melbourne, Australia.  (Section~\ref{section-complex-ner}) 
	
	\item \textbf{Xiang Dai}, Sarvnaz Karimi, Ben Hachey, and Cecile Paris. 2019. Using similarity measures to select pretraining data for NER. In Proceedings of the 2019 Conference of the North American Chapter of the Association for Computational Linguistics: Human Language Technologies, Volume 1 (Long and Short Papers), pages 1460–1470, Minneapolis, Minnesota.  (Chapter 4)
	
	\item \textbf{Xiang Dai}, Sarvnaz Karimi, Ben Hachey, and Cecile Paris. 2020. An effective transition-based model for discontinuous NER. In Proceedings of the 58th Annual Meeting of the Association for Computational Linguistics, pages 5860--5870, Online.   (Chapter 5)
	
	\item \textbf{Xiang Dai}, Sarvnaz Karimi, Ben Hachey, and Cecile Paris. 2020. Cost-effective Selection of Pretraining Data: A Case Study of Pretraining BERT on Social Media. In Findings of the 2020 Conference on Empirical Methods in Natural Language Processing, Online. (Chapter 4)
	
	\item \textbf{Xiang Dai}, Heike Adel. 2020. An Analysis of Simple Data Augmentation for Named Entity Recognition. In Proceedings of the 28th International Conference on Computational Linguistics, Online. (Chapter 3)
\end{enumerate}

The following articles are related, but will not be extensively discussed in this thesis:
\begin{enumerate}[resume]
	\item Nicky Ringland, \textbf{Xiang Dai}, Ben Hachey, Sarvnaz Karimi, Cecile Paris, and James R. Curran. 2019. NNE: A dataset for nested named entity recognition in English newswire. In Proceedings of the 57th Annual Meeting of the Association for Computational Linguistics, pages 5176–5181, Florence, Italy.
	
	\item Aditya Joshi, \textbf{Xiang Dai}, Sarvnaz Karimi, Ross Sparks, Cecile Paris, and C Raina MacIntyre. 2018. Shot or not: Comparison of NLP approaches for vaccination behaviour detection. In Proceedings of the 2018 EMNLP Workshop SMM4H: The 3rd Social Media Mining for Health Applications Workshop and Shared Task, pages 43–47, Brussels, Belgium.
	
	\item Lukas Lange, \textbf{Xiang Dai}, Heike Adel, Jannik Str\"otgen. 2020. NLNDE at CANTEMIST: Neural Sequence Labeling and Parsing Approaches for Clinical Concept Extraction. In Iberian Languages Evaluation Forum (IberLEF 2020), Online.
\end{enumerate}

\subsection{Definition and clarification of used terms}
The usage of technical terminologies in the literature is usually confusing and inconsistent, especially when researchers from different communities use the same term to refer to different concepts, or when some conventions are only shared by a small group of people. For the sake of brevity, we define and clarify some frequently used terms in this thesis.
\begin{description}
	\item[Token] An individual occurrence of a linguistic unit in text. We use a token to refer to an individual word unless specified otherwise. If a word is further split into several pieces, we use sub-tokens to refer to these pieces.
	\item[Span] A consecutive sequence of tokens, or an individual token.
	\item[Biomedical concept] Conceptual objects, events, and procedures in the biomedical ontologies. We use entity and biomedical concept interchangeably.
	\item[Biomedical name] An instance where a biomedical concept is referenced to in text. We use mention and biomedical name interchangeably. Following~\citep{McDonald:Crammer:EMNLP:2005}, we denote the mention by the set of token positions that belong to the mention. Therefore, a mention may consist of several spans and mentions may overlap.
	\item[Embedding] A mapping function that converts a token into a dense vector. It can also be considered as a look-up dictionary, where the token is the key and the vector is the value.
	\item[Encoder] A mapping function that converts a sequence of tokens into a sequence of dense vectors:
	\[
	[\dvector{h}_1, \dvector{h}_2, \cdots, \dvector{h}_N] = \textbf{Text encoder} \left( [t_1, t_2, \cdots, t_N] \right),
	\]
	where $N$ is the sequence length. 
	
	Usually, the encoder is a trainable neural network, and the output vectors are contextualised in the sense that they reflect both the corresponding token and its contexts.
	\item[Attention] A mechanism that is widely used in sequence models to allow the current state `attending' to context states to obtain a context vector. The resulting context vector is usually be used together with the current state for the downstream layers. Given the current state $\dvector{h}_i$ and a sequence of context states $\dmatric{H} = \{\dvector{h}_i\}_{i=1}^N$, we can calculate the context vector $\dvector{c}_i$ as the average of context states weighted with attention scores:
    \begin{equation}
    \dvector{c}_i = \sum_{j=1}^N \dscalar{a}_{ij} \dvector{h}_j,
    \end{equation}
    where $\dscalar{a}_{ij}$ is the $j$-th element in the attention vector $\dscalar{a}_{i}$:
    \begin{equation}
    \dscalar{a}_{i} = \mathrm{softmax} ( f ( \dvector{h}_i, \dmatric{H} ) ).
    \end{equation}
    We use the function $f$ proposed by~\citet{Luong:Pham:EMNLP:2015}, unless specified otherwise:
    \begin{equation}
    f ( \dvector{h}_i, \dmatric{H} ) = \dvector{h}_i^\top \cdot \dmatric{W} \cdot \dmatric{H},
    \end{equation}
    where $\dmatric{W}$ is a trainable weight matrix.
    \item[Language representation model] Similar to text encoder, a language representation model can generate a contextual vector representation for each input token. The subtle difference between a text encoder and a language representation model is that the former focuses more on generating task-oriented representations, and the latter emphasises the general semantic and syntactic representations. For example, a text encoder in an NER model may assign similar vectors to words belonging to the same entity category, even though their semantic meaning are dissimilar.
    \item[Domain-specific vs. Generic domain] The term `domain' is loosely used in both machine learning and NLP communities~\citep{Ramponi:Plank:COLING:2020}. In this thesis, we do not attempt to define what constitutes a domain but assume the domain exists in a covert way.
    
    We use domain-specific model to indicate that the model focuses on a specific domain. For example, we call the language representation model trained on biomedical corpora a domain-specific model. In contrast, we use generic model to refer to a model which is supposed to capture any kind of knowledge. For example, we call the language representation model trained on Common Crawl corpus a generic domain model.
    \item[Supervised vs. Unsupervised] We use supervised learning to indicate that human annotators are required to label the training set. In contrast, if training does not require labels annotated by human annotators, then we call it unsupervised learning, or self-supervised learning. For example, we call pre-training language models unsupervised learning. The task is to predict the next token, given a sequence of tokens, and this task does not require human annotated labels.
\end{description}

\section{Summary}
In this chapter, we first illustrate how an NER model can be used to extract useful information, improving applications in the biomedical domain. Next, we describe a state-of-the-art NER model, \FLAIR, which is based on sequence tagging techniques. Then, we identify three open challenges of applying current techniques to recognise biomedical names.

In the following chapters, we organise our related work and content chapters into three groups, each of which focuses on solving one particular challenge: Section~\ref{section-low-resource} and Chapter~\ref{chapter-data-augmntation} focusing on the lack of labelled data problem; Section~\ref{section-domain-specific-langauge} and Chapter~\ref{chapter-select-pretraining-data} on selecting suitable pre-training data given the downstream task; and Section~\ref{section-complex-ner} and Chapter~\ref{chapter-discontinuous-ner} on recognising biomedical names containing complex inner structure. 

%% file: src/ch3-literature-review.tex
Named Entity Recognition (NER) techniques have developed gradually from dictionary based and rule based to machine learning based approach during the last several decades. Motivated by previously discussed challenges of training supervised models---e.g., labelling biomedical NER dataset can often be expensive and time-consuming---we provide an overview of promising approaches to overcome the lack of training data problem, with a special focus on data augmentation (Section~\ref{section-low-resource-data-augmentation}) as well as transfer learning (Section~\ref{section-domain-specific-langauge}). Additionally, complex structures---overlapping and discontinuity---are common in biomedical names. We review the existing methods for complex entity recognition, and group these methods into token-level, span-level and sentence-level approaches (Section~\ref{section-complex-ner}). 

\section{A Brief History of NER}
Information Extraction (IE) is an important Natural Language Processing (NLP) task that aims to automatically extract structured information from unstructured text. It has been widely used in many applications. For example, a successful email system can identify messages that contain event information, extract the attributes of the event (i.e., time, location, and participants), and insert the extracted event to the calendar~\citep{Laclavik:Dlugolinsky:2012}. In the biomedical domain, IE has been widely used to extract biomedical concepts, attributes, events, and their relations from scholarly articles, clinical notes, and social media data~\citep{Sarawagi:2008,Wang:Wang:JBI:2018}. 

One of the common practices in IE is to separate processing into several stages, among which NER is typically employed as the first step~\citep{Hobbs:JBI:2002}. On the one hand, NER requires a deeper analysis than key word searches, because the semantics of entity mentions are influenced by their contexts. For example, \textit{`Washington'} may refer to a person, a city, a state, or an organisation, depending on the contexts. On the other hand, NER does not seek to fully understand every aspect of the text, such as the writer's communicative intent~\citep{Bender:Koller:ACL:2020}. Therefore, it focuses only on relevant words and ignores the rest. Because of its location at a midpoint on this spectrum, NER is a fundamental task, and it has received considerable attention in the last several decades. 

\citet{Grishman:Sundheim:1996} use the term \emph{Named Entity}, referring to possible persons, organisations, and locations mentioned in text, and they aim to recognise structured information of company activities and defence related activities from newspaper articles. \citet{Florian:Hassan:NAACL:2004} extend the task to recognise mentions of textual references to conceptual objects, which can be either named (e.g. \textit{`George Washington'}), nominal (e.g. \textit{`The president'}) or pronominal (e.g, \textit{`He'}). The entity categories studied in the generic domain are mainly person, organisation, and location. 

Biomedical NER, focusing on identifying and classifying biomedical names, whose surface forms can represent biomedical concepts, has its unique characteristics comparing to NER in the generic domain. Early stage efforts, e.g., GENIA project~\citep{Collier:Park:EACL:1999,Kim:Ohta:Bioinformatics:2003} and BioCreAtIvE (Critical Assessment for Information Extraction in Biology) challenges~\citep{Hirschman:Yeh:BMC:2005}, focus on automatically extracting genome information from biochemical papers written by domain specialists. i2b2 (Informatics for Integrating Biology \& the Bedside) and n2c2 (National NLP Clinical Challenges) projects~\citep{Kohane:Masys:JAMA:2006,Brownstein:Murphy:2010} start to bring Electronic Health Records (EHRs) to researchers' attention, by releasing publicly available de-identified clinical notes. Additionally, the value of informal sources, such as user generated text on the web and search engine logs, have also been recognised by researchers. They start to use these data for mining health related information, such as predicting epidemic events~\citep{Joshi:Karimi:Survey:2019}, and monitoring adverse drug events~\citep{Sarker:Ginn:JBI:2015}.

In this section, we provide a brief overview of the development of NER techniques. Instead of exhaustively surveying different approaches and discussing design variants, we describe representative work and focus on identifying what are the strengths and limitations of different approaches. For more detailed surveys of NER techniques in both generic and biomedical domains, we refer the reader to \citep{Nadeau:Sekine:2007,Campos:Matos:2012,Yang:Liang:COLING:2018,Yadav:Bethard:COLING:2018,Li:Sun:TKDE:2020}. 

\subsection{Dictionary based approach}
\citet{Mikheev:Moens:EACL:1999} build a minimal NER system equipped with dictionaries, also known as gazetteers or name lists. They collect person names, organisation names and location names from the MUC-7 training data, as well as several external resources, including the CIA World Fact Book, financial web sites, etc. Despite its simplicity, evaluation results on MUC-7 test set show that pure list lookup---finding occurrences of exact matches with items from dictionaries---performs reasonably well for locations (precision of $0.90$ and recall of $0.86$), but not for the organisation and person categories (recall of lower than $0.50$, precision of around $0.80$). 

One serious limitation of this approach is that it cannot recognise unseen entity mentions, i.e., entities not in the dictionaries. In addition, maintaining large dictionaries requires great efforts. For example, there are around 1.5 million unique family names, just in the United States. The dictionary of company names, if at all available, would be much larger and out of date quickly, because new companies emerge all the time. 

Naming variation is another issue that needs to be overcome. For example, the organisation dictionary might contain \textit{`University of Sydney'}, but this organisation may also be referred to as \textit{`Sydney Uni'}. In the biomedical domain, this problem is even more severe. For example, the drug \textit{`Acetylcysteine'}, usually used for cough and other lung conditions, is also known as \textit{`Acetyl Cysteine'}, \textit{`Cysteine Hydrochloride'}, \textit{`Cystine'}, \textit{`N-acetyl cysteine'}, \textit{`N-acetylcysteine'}, \textit{`N-acetyl-L-cysteine'}, \textit{`N-Ac\'etyl-L-Cyst\'eine'}, etc. 

Finally, ambiguity may be caused by the overlapping between dictionaries belonging to different entity categories. For example, \textit{`J. P. Morgan'} could belong to both the person name dictionary and the organisation name dictionary. Ambiguity can also be caused by the usage of abbreviations and acronyms. For example, \textit{`CRF'} may refer to \textit{`Conditional Random Field'} in the context of natural language processing. However, the possible number of meanings of the term \textit{`CRF'} in the context of biomedical is much larger, including \textit{`Cardiorespiratory fitness'}---relating to heart health, \textit{`Clinical risk factors'}, \textit{`Controlled Rate Freezer'}---a medical equipment, \textit{`Chronic renal failure'}---a type of kidney disease, etc. Note that clinical notes are usually written by practitioners under time pressure. So abbreviations and acronyms are used frequently. All of these limitations make the dictionary based approach more difficult to be widely employed. 

\subsection{Rule based approach}
To overcome the previously mentioned limitations of dictionary based approaches, efforts were made to handcraft a set of rules to alleviate the reliance on the completeness of dictionaries. Rules can be created to expand the dictionaries to identify previously unseen mentions. For example, MetaMap~\citep{Aronson:AMIA:2001,Aronson:Lang:AMIA:2010} makes use of external knowledge sources of biomedical terms---the SPECIALIST lexicon, and it employs complex rules to identify all possible mention variants of an entity, including acronyms, abbreviations, synonyms, or derivational variants. Table~\ref{table-background-metamap-variant} is an example that illustrates how expansion rules are used to generate variants given a word. Expansion rules include `i' (inflection), `p' (spelling variant), `a' (acronym/abbreviation), `e' (expansion of acronym/abbreviation), `s' (synonym) and `d' (derivational variant). For example, the expansion rule of variant `ophthalmia'---`ssd'---indicates that it is a derivational variant of a synonym (\textit{`ophthalmic'}) of a synonym (\textit{`eye'}) of \textit{`ocular'}. 

\begin{table}[tb] 
	\centering
	\begin{tabular}{r | c | c | c | c | c}
		\toprule
		Origin & \multicolumn{3}{c|}{Variant} & \POS & Expansion Rule \\ \hline
		ocular & \multicolumn{3}{c|}{} & adj & -- \\ \hline
		& eye & & &  noun & s \\
		& & eyes & & noun & si \\
		& & optic & & adj & ss \\
		& & ophthalmic & & adj & ss \\
		& & & ophthalmia & noun & ssd \\
		& oculus & & &  noun & d \\
		& & oculi & &  noun & di \\
		\bottomrule
	\end{tabular}
	\caption{The variants of word \textit{`ocular'} and the corresponding rules to generate them. The indentation reflects the hierarchical structure of these variants according to the history of how they are generated.~\label{table-background-metamap-variant}}
\end{table}

Rules can also be triggered by characteristic attributes of known entity mentions, including their spellings and the contexts in which they appear. For example, a spelling rule can be a simple look up for the string, such as \emph{any string containing `Mr.' is a person}; or a spelling pattern, such as \emph{any all capitalised string is an organisation (e.g., `IBM')}. A contextual rule gets clues from surrounding words and their syntactic relationships, such as \emph{any proper name modified by an appositive whose head is `president' is a person (e.g., `Maury Cooper' in the context of `... says Maury Cooper, a vice president at ...')}. 

One advantage of a rule based approach is that rules can be derived using unlabelled text only, which are much easier to obtain. For example, \citet{Collins:Singer:EMNLP:1999} build a named entity classifier using $90,000$ unlabelled examples. They start from $7$ seed rules (\emph{`New York', `California' and `U.S.' are locations}; \emph{any name containing `Mr.' is a person}; \emph{any name containing `Incorporated' is an organisation}; \emph{`I.B.M.' and `Microsoft' are organisations}), which is the only supervision in their approach. The classifier, automatically inducing new spelling rules and contextual rules, finally achieves over $91\%$ accuracy when evaluated on a test set of $1,000$ manually labelled instances. 

However, applying these rules is challenging, when the number of rules become large. That is, it is difficult to prioritise one particular rule over others, especially when some of these rules may conflict with each other. For example, the following three rules may be used to represent the same example: \emph{any all capitalised string is an organisation (e.g., `IBM')}; \emph{any string which is all capitalised or full periods, and contains at least one period is a location (e.g., `N.Y.')} and \emph{any string has an appositive modifier whose head is a singular noun (`player') is a person (e.g., `L.J., the greatest basketball player')}. Iterating over all possible applicable rules and arranging them in order of importance, even if at all possible, will cause heavy computations. 

Another difficulty of applying these rule based systems is that they usually rely on other NLP tools, such as a syntactic parser. For example, \citet{Zhang:Elhadad:JBI:2013} use a noun phrase chunker to first identify candidate entity mentions; and context rules used by~\citet{Collins:Singer:EMNLP:1999} involve finding the head word of the appositive modifier for the entity mention. Building syntactic analysis tools itself is a challenge task, especially for syntactically noisy text, such as clinical notes and social media data. 

\subsection{Statistical machine learning based approach}
Statistical machine learning approaches replace `hard' rules with `soft' features and estimate the importance (weights) of features using labelled training data. Tokens are typically represented by vectors, each of which can consist of boolean, numeric and nominal values, representing each token-in-the-context. For example, a boolean value can be used to indicate whether the token is capitalised, and a nominal attribute can be used to represent the stem of the token. The feature creating function, mapping from a token to a sparse vector, is called a feature template. It controls the length of the token vector and the meaning of each element in the vector. Once the feature template is fixed, feature vectors---created via the same mapping function---can be taken as input of any supervised classifier, including Decision Tree, Maximum Entropy Models, Support Vector Machines, Hidden Markov Models and Conditional Random Fields. 

Despite the successful applications of machine learning based NER, its main shortcoming is the requirement of sophisticated feature templates. These features should be informative and generalisable for unseen data. This is challenging because such high quality feature engineering requires expert domain knowledge and is usually tailored to specific entity categories or text types. Learning from these features may also suffer from the sparsity problem. For example, if a stem appears only one time in the training data, it is impossible to estimate its importance---the weight associated with the feature---from such a rare observation. 

To alleviate the burden of manually building feature templates, deep learning models enable automated feature extraction. Distributed representations are usually employed to solve the sparsity issue. In other words, the mapping function is a neural network. It takes a token as input, and it outputs a dense vector instead of a sparse vector. Feature vectors---created via the neural network---can be combined with almost any previous mentioned classifier, except for those which are better at sparse input vectors, such as Decision Tree.

\section{NLP for Low Resource Scenarios~\label{section-low-resource}}
\input{src/ch3-low-resource.tex}

\section{Transfer Learning~\label{section-domain-specific-langauge}}
\input{src/ch3-transfer-learning}

\section{Complex Entity Recognition~\label{section-complex-ner}}
\input{src/ch3-complex-ner.tex}

%% file: src/ch3-low-resource.tex
Although supervised neural models have achieved state-of-the-art performance on numerous benchmark NER datasets in the generic domain, due to the availability of large amount of labelled training data (\emph{high resource}), they does not cover all applications. On one hand, the trained model usually does not generalise well across different types of text, let alone to recognise mentions belonging to new entity categories (Table~\ref{table-low-resource-generization}). On the other hand, re-annotation for a new task, domain or language requires considerable effort. 

In this section, we describe several approaches---except transfer learning which will be detailed in Section~\ref{section-domain-specific-langauge}---to deal with the lack of labelled training data problem (\emph{low resource}). We describe data augmentation approaches in details, because our methods described in Chapter~\ref{chapter-data-augmntation} are built on top of such related work. 

\begin{table}[tb] 
	\centering
	\begin{tabular}{r | c | p{10cm} }
		\toprule
		\bf Evaluation Dataset & $\mathbf{F_1}$ & \bf Task Description \\
		\midrule
		\textsc{NCBI-Disease} & 87.5 & Recognise disease names in biomedical publications \\
		\textsc{i2b2-2010} & 43.6 & Recognise disease names (labelled as \emph{problem}) in clinical notes \\
		\textsc{n2c2-2019} & 60.2 & Recognise genetic disease names (labelled as \emph{observation}) in clinical notes \\
		\bottomrule
	\end{tabular}
	\caption{Decline in effectiveness of a model trained on \textsc{NCBI-Disease}~\citep{Dogan:Leaman:JBI:2014}, when evaluated on other datasets: \textsc{i2b2-2010}~\citep{Uzuner:South:AMIA:2011}, and \textsc{n2c2-2019}~\citep{n2c2-2019-shared-task}. The mention-level $F_1$ score is reported.}
	\label{table-low-resource-generization}
\end{table}

\subsection{Distant supervision~\label{section-low-resource-distant-supervision}}
Instead of manually labelling data, one research direction---often called \emph{distant} or \emph{weak} supervision~\citep{Hoffmann:Zhang:ACL:2011}---aims to automatically create training data by exploring existing knowledge base (e.g., Wikipedia, MeSH\footnote{Medical Subject Headings: \text{https://www.nlm.nih.gov/mesh/meshhome.html}. Accessed date: \today}, CTD\footnote{Comparative Toxicogenomics Database: \text{http://ctdbase.org/downloads/}. Accessed date: \today}) or heuristic rules. 

\citet{Nothman:Curran:ALTA:2008} transform Wikipedia into named entity annotations by (1) classifying Wiki articles into common entity categories; (2) finding all possible inter-article links; and (3) assigning the entity category of the target page to the anchor text. Because the authors of Wikipedia are dictated to link only the first mention of an entity in each article, \citeauthor{Nothman:Curran:ALTA:2008} use several heuristic rules to infer additional links from shorter referential forms. For example, the first or last word of a person name found later in the article may also refer to the same person. In addition, heuristic rules are employed to adjust link boundaries. For example, linked text may contain the possessive \textit{'s} at the end of a name, and it should be removed from the entity name. 

\begin{figure}[tb] 
	\centering
	\includegraphics[width=0.9\textwidth]{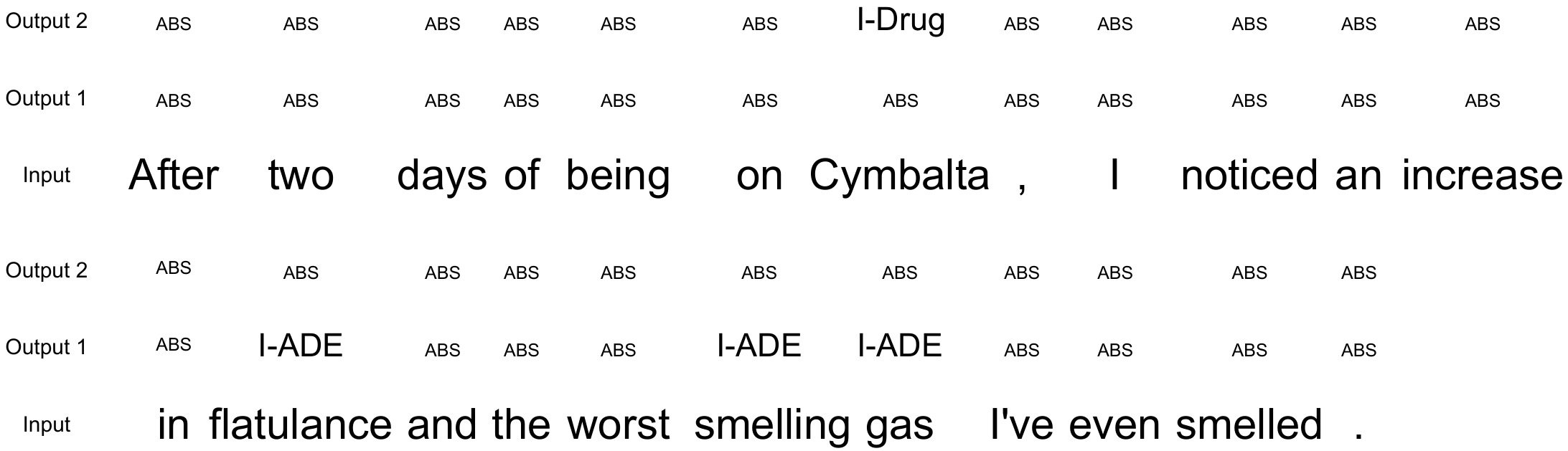}
	\caption{Example sequence of tags generated by a rule and two domain-specific dictionaries. For example, \textit{`Cymbalta'} is assigned the tag \textit{`I-Drug'} because it appears in the drug dictionary.~\label{figure-low-resource-distant-supervision-tag}}
\end{figure}

\citet{Safranchik:Luo:AAAI:2020} describe a framework, which takes unlabelled data and a set of rules as input, for creating labelled training data. Rules, which are implemented as functions, can take unlabelled data as input and output heuristic information about tags. For example, a simple rule can be \textit{`tagging any token that appear in a dictionary of known entity category as I-$\star$, and all other tokens as ABS, indicating that the rule abstains from assigning a tag'}. Taking the sentence in Figure~\ref{figure-background-bio-example} as an example, this rule---combined with a drug dictionary and an adverse drug event dictionary---may create the sequence of tags in Figure~\ref{figure-low-resource-distant-supervision-tag}, if there are some tokens appearing in these dictionaries. 

Note that it is possible that different rules output conflicting tags, if one token appears in multiple dictionaries. Also, it is possible that the identified span is incomplete. For example, \textit{`increase in flatulance'} should be identified as an adverse drug event, but because \textit{`increase in'} may not appear in the adverse drug event dictionary, these two tokens are labelled as \textit{`ABS'} (the rule abstains from assigning a tag). To reconcile the conflicting and incomplete information, \citet{Safranchik:Luo:AAAI:2020} introduce a set of \emph{linking rules} that decide whether adjacent tokens should be grouped into one span, and which tag is used for the span. For example, \textit{`increase in flatulance'} can be grouped into a span and share the tag assigned to \textit{`flatulance'}. These linking rules are usually implemented based on automatic phrase mining, or with the help of language models that predicts which words may co-occur. 

Although these described automatic labelling methods provide a cheap way to obtain a large amount of labelled training data, the obtained labelled data are usually \emph{noisy}. Automatically annotated labels usually contain more errors than the manual annotations (Table~\ref{table-low-resource-noisy-label}), which are in contrast called \emph{clean data}. \citet{Liang:Yu:KDD:2020} point out that there is a trade-off between recall and precision using automatic labelling. That is, setting strict rules can generate high precision labels, but may not generalise well and thus have low recall. In contrast, relaxed rules can increase the coverage of annotation, leading to high recall and low precision. 

\begin{table}[tb] 
	\centering
	\begin{tabular}{l | c | c | c}
		\toprule
		\bf Evaluation Dataset & \bf Precision & \bf Recall & $\bm F_1$ \\
		\midrule
		\CONLL \textsc{Dutch}~\citep{Sang:CONLL:2002} & 32.4 & 21.1 & 25.5 \\
		\CONLL \textsc{Spanish}~\citep{Sang:CONLL:2002} & 51.0 & 24.7 & 33.3 \\
		\CONLL \textsc{English}~\citep{Sang:Meulder:CONLL:2003} & 39.9 & 30.1 & 34.3 \\
		\CONLL \textsc{German}~\citep{Sang:Meulder:CONLL:2003} & 23.2 & \phantom{0}9.2 & 13.2 \\		
		\textsc{Estonian}~\citep{Tkachenko:Petmanson:BSNLP:2013} & 59.7 & 49.3 & 54.0 \\
		\bottomrule
	\end{tabular}
\caption{Evaluation results, as reported by~\citet{Lange:Hedderich:EMNLP:2019}, of automatically annotated labels against manual annotations.~\label{table-low-resource-noisy-label}}
\end{table}

\paragraph{Learning in the presence of noisy labels}
Training a supervised model on noisy labels can sometimes result in negative results. \citet{Fang:Cohn:CoNLL:2016,Hedderich:Klakow:DeepLo:2018} show that training on the combination of noisy training data and a small amount of clean training data performs worse than training on clean data only. Therefore, efforts are made to solve the noisy labelled data problem~\citep{Han:Yao:NIPS:2018,Liang:Yu:KDD:2020}. 

One popular approach of training with noisy labels is to model the true label as a latent variable and learn a noisy model that relate the true and noisy labels~\citep{Hedderich:Klakow:DeepLo:2018,Lange:Hedderich:EMNLP:2019}. We use $P(y|x)$ to represent the probability distribution of a small set of clean instances $(x,y) \in \dset{C}$, and use $P(\tilde{y}|x)$ to represent the distribution of a large set of noisy instances $(x, \tilde{y}) \in  \dset{N}$. Then, the noisy distribution can be calculated using:
\begin{equation}
	P ( \tilde{y}=j | x ) = \sum_{i=1}^k P(\tilde{y}=j|y=i) P (y=i|x).
\end{equation}
To estimate the relationship between true and noisy labels, i.e., $P(\tilde{y}=j|y=i)$, \citet{Hedderich:Klakow:DeepLo:2018} first apply the same auto-labelling operations on clean data $\dset{C}$ to obtain pairs of clean $y$ and corresponding noisy label $\tilde{y}$. Then, a simple noisy layer is used to model the relationship between true and noisy labels using these label pairs:
\begin{equation}
	P(\tilde{y}=j \mid y=i)=\frac{\exp \left(b_{i j}\right)}{\sum_{l=1}^{k} \exp \left(b_{i l}\right)},
\end{equation}
where
\begin{equation}
	b_{i j}=\log \left(\frac{\sum_{t=1}^{|C|} \bm 1_{\left\{y_{t}=i\right\}} \bm 1_{\left\{\tilde{y}_{t}=j\right\}}}{\sum_{t=1}^{|C|} \bm 1_{\left\{y_{t}=i\right\}}}\right).
\end{equation}
\citet{Lange:Hedderich:EMNLP:2019} further extend this method by taking the input features into consideration. That is, they first cluster contextual token vectors, and then build different distributions for each cluster, i.e., $P(\tilde{y}=j|y=i;x)$. Experimental results show that this method improves the $F_1$ score up to 36\% over methods without noise handling when evaluate on low-resource NER settings. 


\subsection{Active learning~\label{section-low-resource-active-learning}}
Active learning is a promising approach for efficient annotation, based on the hypothesis that the learning algorithm can perform better with less training if it is allowed to choose the data from which it learns~\citep{Settles:2009}. It can be used when expert annotators are available during the development cycle, but the number of instances they can annotate under budget is far less than the usual number of labelled instances needed to train a supervised model, to reach satisfactory performance. Instead of asking annotators to annotate a set of randomly sampled (\emph{passive}) instances, \emph{active} learning uses algorithms to choose a small set of \emph{informative} instances to annotate. 

\begin{figure}[tb] 
	\centering
	\includegraphics[width=0.9\textwidth]{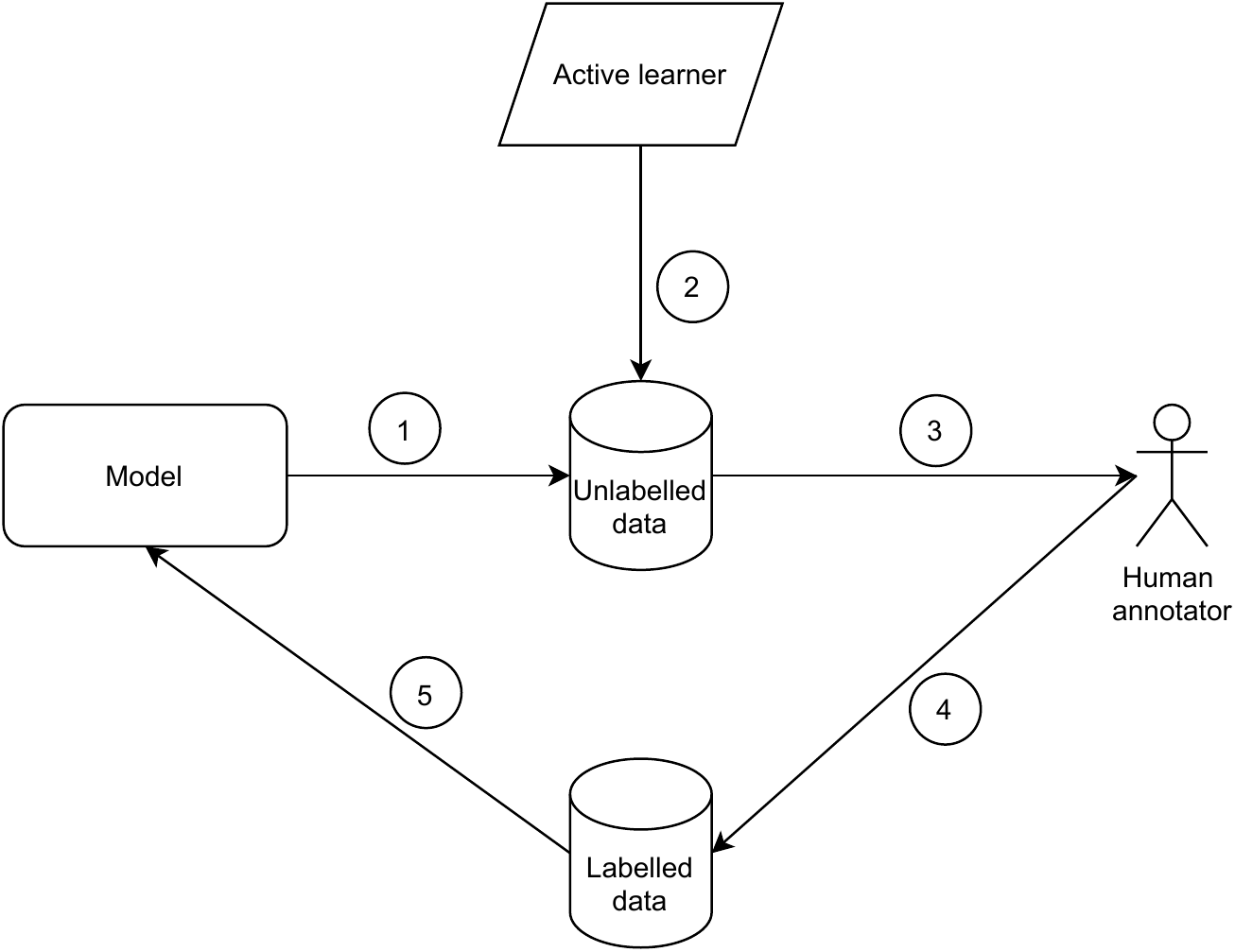}
	\caption{The active learning process usually have multiple rounds, each of which consists of five steps: (1) applying model on unlabelled data; (2) querying on unlabelled data; (3) presenting informative instances; (4) annotating instances; and (5) updating the model.~\label{figure-low-resource-active-learning}}
\end{figure}

A series of events in active learning is shown in Figure~\ref{figure-low-resource-active-learning}. They are repeated until the annotation budget has run out or the model performance has reached the satisfactory level. At the beginning, a model that may be trained on a small number of labelled instances or transferred from other tasks is available to make predictions on unlabelled data. The active learner chooses a small number of instances, which are considered most informative, and presents them to the expert annotators. After receiving human annotations, the model parameters can be either retrained from scratch using all available labelled data, or incrementally updated by training only on the newest batch of labelled data~\citep{Shen:Yun:ICLR:2018}. 

Although many variants exist during each step of the active learning cycle~\citep{Settles:2009}, the key component in active learning is assessing how \emph{informative} each unlabelled instance is. In the following, we describe two widely used approaches with sequence models, including \emph{uncertainty sampling} and \emph{query-by-committee}, and refer the reader to~\citep{Settles:Craven:EMNLP:2008,Settles:2009,Olsson:2009} for more options. 

\paragraph{Uncertainty sampling}
Active learner employing uncertainty-based functions chooses the instance whose label is most uncertain given the existing model. 

\citet{Culotta:McCallum:AAAI:2005} choose the instance for which the existing model has the least confidence in its best prediction: 
\begin{equation}
	\phi ( \mathbf{x} ) = 1 - P (\mathbf{y}^\star | \mathbf{x}; \bm\theta),
\end{equation}
where $\mathbf{y}^\star$ is the most likely label sequence, and $\bm \theta$ represents the existing model. 

\citet{Scheffer:Decomain:IDA:2001} choose the instance with the smallest margin between its two best predicted label sequences: 
\begin{equation} 
\phi ( \mathbf{x} ) = - (P (\mathbf{y}_1^\star | \mathbf{x}; \bm \theta) - P (\mathbf{y}_2^\star | \mathbf{x}; \bm \theta)),
\end{equation}
where $\mathbf{y}_1^\star$ and $\mathbf{y}_2^\star$ are the first and second most likely labelling, respectively. 

\paragraph{Query-By-Committee}
In contrast to uncertainty sampling where only one model is used, methods belonging to query-by-committee category use multiple models, known as a committee of models. The active learner chooses the instance over which a committee of models are in most disagreement. Note that the committee needs to be comprised of diverse models. 

\citet{Settles:Craven:EMNLP:2008} use the bagging technique to train different models. Several subsets are first randomly sampled with replacement from the original labelled training set. The same base model is then trained on each subset to create a committee of diverse models. Similarly, \citet{Shen:Yun:ICLR:2018} draw a committee of models via applying independently sampled dropout masks---thus different subsets of the neural network---to the same CNN-LSTM model. 

To measure disagreement among a set of $C$ models, \citet{Argamon-Engelson:Dagan:JAIR:1999} introduce a measure called vote entropy: 
\begin{equation} 
\phi ( \mathbf{x} ) = - \frac{1}{T} \sum_{t=1}^T \sum_{j=1}^J \frac{F(\mathbf{y}_t, j)}{C} \log \frac{F(\mathbf{y}_t, j)}{C},
\end{equation}
where $\mathbf{y}_t$ be a list of $C$ labels predicted by all the committee models at sequence position $t$, and $F(\mathbf{y}_t, j)$ is the frequency of label $j$ in $\mathbf{y}_t$. \citet{Shen:Yun:ICLR:2018} first find the most popular choice $\mathbf{y}_t^{\star}$ in $\mathbf{y}_t$, and then measure the disagreement by calculating the ratio of models which disagree with $\mathbf{y}_t^{\star}$: 
\begin{equation} 
\phi ( \mathbf{x} ) = - \frac{1}{T} \sum_{t=1}^T (1 - \frac{|{c: \mathbf{y}_t^{(c)} \ne \mathbf{y}_t^{\star}}|}{C}),
\end{equation}
where $|\cdot|$ denotes cardinality of a set. 

Instead of measure disagreement on the token-level, \citet{Settles:Craven:EMNLP:2008} describe two sequence-level measures, which consider the label sequence as a whole. Given the posterior probability of a label sequence based on a particular model, $P(\mathbf{\hat{y}} | \mathbf{x}; \bm \theta^{(c)})$, they first calculate the probability given the committee of models via: 
\begin{equation} 
P(\mathbf{\hat{y}} | \mathbf{x}; C) = \frac{1}{C} \sum_{c=1}^C P(\mathbf{\hat{y}} | \mathbf{x}; \bm \theta^{(c)}).
\end{equation}
Then a set of predicted label sequences, $\mathcal{N}^C$, is obtained by taking the union of the $N$-best predictions from all models. Finally, the disagreement can be measured by calculating sequence Kullback-Leibler: 
\begin{equation}
	\phi ( \mathbf{x} ) = \frac{1}{C} \sum_{c=1}^C \sum_{\mathbf{\hat{y}} \in \mathcal{N}^C} P(\mathbf{\hat{y}} | \mathbf{x}; \bm \theta^{(c)}) \log \frac{P(\mathbf{\hat{y}} | \mathbf{x}; \bm \theta^{(c)})}{P(\mathbf{\hat{y}} | \mathbf{x}; C)},
\end{equation}
or sequence entropy: 
\begin{equation} 
	\phi ( \mathbf{x} ) = - \sum_{\mathbf{\hat{y}} \in \mathcal{N}^C} P(\mathbf{\hat{y}} | \mathbf{x}; C) \log P(\mathbf{\hat{y}} | \mathbf{x}; C).
\end{equation} 



\subsection{Data augmentation~\label{section-low-resource-data-augmentation}}
Data augmentation, expanding the training set by transforming training instances without changing their labels, is heavily studied in the field of computer vision~\citep{Shorten:Khoshgoftaar:BigData:2019}. Simple augmentations, such as cropping, resizing, rotating and flipping, have become standard practices in vision tasks. However, data augmentation is still under exploration in NLP. In this section, we survey data augmentations for sentence level NLP tasks, such as text classification, natural language inference and machine translation, and group them into four categories based on how they generate augmented instances: (1) word replacement, (2) mention replacement, (3) word position swapping, and (4) using generative models. 

\subsubsection{Word replacement}
Various word replacement approaches have been explored to generate augmented instances for text classification tasks. \citet{Zhang:Zhao:NIPS:2015} generate augmented instances by replacing words in the original instance with their synonyms, which are retrieved from an English thesaurus---\textsc{WordNet}~\citep{Miller:Beckwith:1990}. They first extract all replaceable words from the original instance, and randomly choose $n$---determined by a geometric distribution---of them to be replaced. Then a random synonym given a word is chosen to replace the original word. Similarly, \citet{Wei:Zou:EMNLP:2019} randomly choose $n$ words that are not stop words and replace each of them with one of its synonyms chosen at random. \citeauthor{Wei:Zou:EMNLP:2019} show that, when the number of original training instances is small (i.e., $500$),  randomly choosing and replacing $10\%$ of words from the sentence can increase the classification accuracy by $2\%$ on average. However, when replacing too many words, for example more than $20\%$ of words in the sentence, performance gain diminishes. 

\citet{Kobayashi:NAACL:2018} proposes context-aware augmentation that replace words with other words which are predicted by a language model at the word positions. Specifically, the author pre-trains a \bilstm language model on \textsc{WikiText-103}~\citep{Merity:Xiong:arXiv:2016} -- a subset of English Wikipedia articles. Then, given the surrounding words, denoted as $S$, at each word position $i$, replacement is sampled from an annealed distribution, $P(\cdot | S)^{1/\tau}$, using the language model. The parameter $\tau$ is used to control the strength of the language model. That is, when $\tau$ becomes infinity, the words are sampled from a uniform distribution. When it becomes zero, the augmentation word is always the one with the highest probability. One problem of context-aware augmentation is that the predicted word may not be compatible with the original label. For example, in a sentiment analysis dataset, the original instance \textit{`the actors are fantastic'} is labelled as \emph{positive}. Given the word position of \textit{`fantastic'}, the language model often assigns high probabilities to words such as \textit{`bad'} or \textit{`terrible'}. To solve this problem, \citeauthor{Kobayashi:NAACL:2018} concatenates the embedded label $y$ with surrounding words and use it as input to the \bilstm language model. In other words, when training the model, \citeauthor{Kobayashi:NAACL:2018} calculate a label-conditional language model: $P(\cdot | y, S)$ instead of $P(\cdot | S)$. Evaluation results on several classification datasets show that context-aware augmentation slightly outperforms synonym-based augmentation, by accuracy of $0.5\%$ on average. 

For machine translation, word replacement has also been used to generate augmented parallel sentence pairs. \citet{Wang:Pham:EMNLP:2018} replace words in both the source and the target sentence by other words uniformly sampled from the source and the target vocabularies. \citet{Fadaee:Bisazza:ACL:2017} search for contexts where a common word can be replaced by a low-frequency word, relying on recurrent language models. Similarly, \citet{Gao:Zhu:ACL:2019} use a monolingual language model to obtain the replacement for a randomly chosen word. Instead of predicting a single replacement word, they propose to replace the word by a soft word, which is a probabilistic distribution over the vocabulary, represented using a weighted sum of the corresponding word vectors. Experimental results show that, on both low-resource and high-resource machine translation datasets, the soft data augmentation can achieve more than $1.0$ BLEU score improvement over the baseline without using data augmentation. 

\subsubsection{Mention replacement}
Instead of creating augmented instances by replacing individual words, replacement can be employed at the mention level, usually with the help of an external knowledge base and heuristic rules. 

\begin{figure}[tb] 
	\centering
	\includegraphics[width=0.9\textwidth]{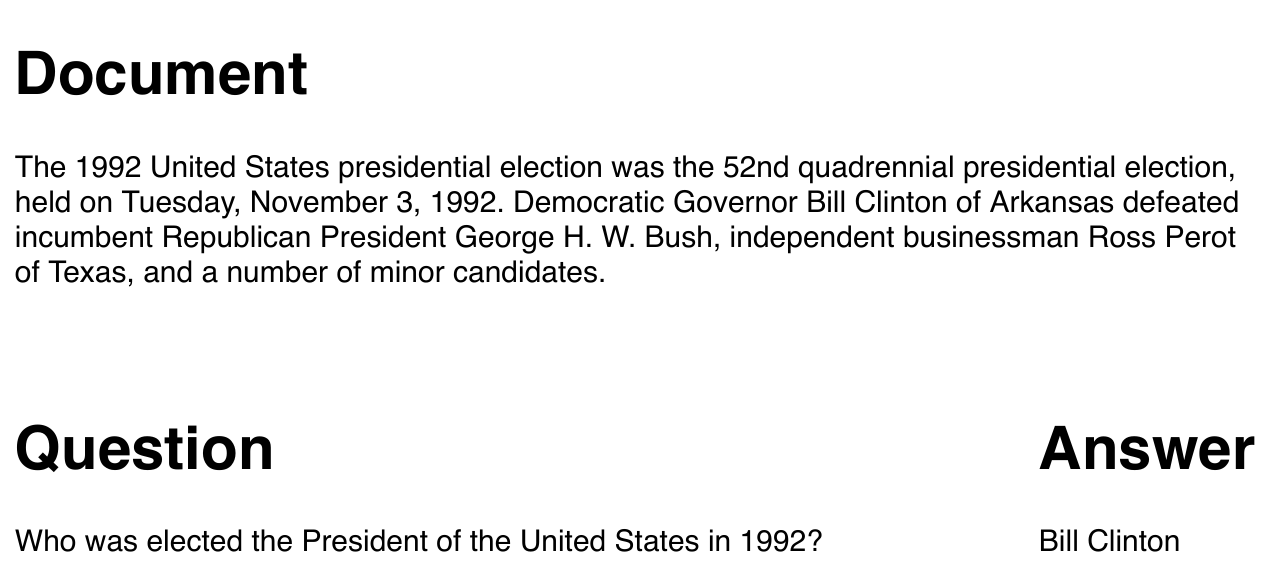}
	\caption{The extractive question answering model tends to use the question type (e.g., Who) and select the spans whose nature agrees with the question type (e.g., `Bill Clinton', `George H. W. Bush', and `Ross Perot'), without the necessity to understand the question.}
	\label{figure-low-resource-qa-mention-replacement}
\end{figure}

After observing that question answering models tend to astray by selecting a text span that shares the answer's type but has the wrong underlying entity (Figure~\ref{figure-low-resource-qa-mention-replacement}), \citet{Raiman:Miller:EMNLP:2017} design an augmentation strategy to make the model more robust to surface form variation. It includes three steps: 
\begin{enumerate}
	\item Extract nominal groups in the training set using a part of speech tagger. 
	\item Perform string matching with entities in Wikidata. 
	\item Randomly replace matched entities in the training set with other entities of the same category in Wikidata. 
\end{enumerate} 

Specifically, they extract $47,598$ entities in \SQUAD that fall under $6,380$ Wikidata \emph{instance of} types. During each training epoch, $T$---a hyperparameter, tuned from a range $[0, 10^5]$---augmented instances are generated and used in combination with the original training set. Experimental results on \SQUAD show that the proposed data augmentation improves the performance by $F_1$ of $1.0$. 

In order to remove gender bias from coreference resolution systems, \citet{Zhao:Wang:NAACL:2018} propose to generate an augmented set where all male entities are replaced by female entities, and vice versa, and train the model on both original and augmented sets. They use a rule based approach consisting of two steps. First, named entities are anonymised. For example, \textit{`John went to his house'} would be anonymised to \textit{`E1 went to his house'}. Then a dictionary of gendered terms and their realisation as the opposite gender is used to change all matching tokens. For example, \textit{`she'} is changed to \textit{`he'}, \textit{`Mr.'} is changed to \textit{`Mrs.'}. Finally, the augmented instance \textit{`E1 went to her house'} is generated and added to the training set. Evaluation results on a benchmark dataset focused on gender bias show that this data augmentation can effectively remove the gender bias without significantly affecting the model performance on other coreference benchmark datasets. 

\subsubsection{Word position swapping}
\citet{Wei:Zou:EMNLP:2019} randomly choose two words in the sentence and swap their positions to augment text classification training sets. They use only one parameter to control the number of words changed based on the sentence length. Experimental results show that random swap can yield high performance gains when less than $20\%$ of words in the sentence are swapped, but decline when more than $30\%$ of words are swapped. 

\citet{Min:McCoy:ACL:2020} explore syntactic transformations (e.g., subject/object inversion, passivisation) to augment the training data for Natural Language Inference (NLI) to mitigate over-fitting. This transformation does not attempt to ensure the naturalness of the generated examples, neither the correctness of labels. For example, in the subject/object inversion transformation, the sentence \textit{`The carriage made a lot of noise'} is transformed into \textit{`A lot of noise made the carriage'}, and the gold label of the augmented instance is set to \emph{neutral} if the original label is \emph{entailment}. Experimental results show that the proposed augmentation does not harm overall performance on the MNLI test set, but it can help the model achieve better generalisation, evaluated on HANS. 


\subsubsection{Generative models}
Instead of creating an augmented instance by manipulating one or several tokens in the original instance, some approaches aim to create a new instance via generative models. 

\citet{Yu:Dohan:ICLR:2018} train a question answering model with data augmented by back-translation from a neural machine translation model. Specifically, they use two translation models, one model from English to French and another model from French to English. They feed the document from an original instance into the English-to-French model to obtain $k$ French translations via the decoder using beam search. Then each of the French translation is passed through a French-to-English model with beam decoder, and can thus obtain $k^2$ paraphrased instances in total. Experimental results on \SQUAD show that the proposed data augmentation can improve the performance by $F_1$ of $1.1$, when the training data is made three times as large by adding augmented instances. 

Similarly, \citet{Xia:Kong:ACL:2019} convert data from a high-resource language to a low-resource language, using a bilingual dictionary and an unsupervised machine translation model in order to expand the machine translation training set for the low-resource language. Results show that, under extreme low resource settings, the proposed data augmentation can improve translation quality measured by BLEU compared to supervised back-translation baselines. 

\subsection{Summary}
In this section, we reviewed three promising approaches---distant supervision, active learning and data augmentation---that aim to achieve high accuracy with as little annotating efforts as possible. Different approaches make use of different types of resources (Table~\ref{table-low-resource-requirement-summary}), and therefore can be suitable for different scenarios. For example, active learning requires expert-in-the-loop, and distant supervision makes use of domain-specific knowledge base or domain knowledge for designing heuristic rules. They are good options once these resources are available. In contrast, data augmentation is the most flexible approach, since some augmentation methods can be applied without the requirement of any domain-specific resources, e.g., word replacement. Encouraged by its adaptability and existing data augmentation methods for sentence-level NLP tasks, we investigate easy to use data augmentation methods for NER, which will be detailed in Chapter~\ref{chapter-data-augmntation}. 

\begin{table}[tb] 
	\centering
	\begin{tabular}{c | c | c | c | c}
		\toprule
		& Labelled data & Unlabelled data & Knowledge base & Domain expert \\
		\hline
		Active learning & & $\checkmark$ &  & $\checkmark$ \\
		Data augmentation & $\checkmark$ & & & \\
		Distant supervision & & $\checkmark$ & $\checkmark$ & $\checkmark$ \\ 
		\bottomrule
	\end{tabular}
	\caption{Requirement of different types of resources by each approach.}
	\label{table-low-resource-requirement-summary}
\end{table}

We note that there are other approaches to overcome the low resource problem, such as unsupervised learning~\citep{Collins:Singer:EMNLP:1999,Etzioni:Cafarella:AI:2005,Zhang:Elhadad:JBI:2013}, as well as transfer learning, which we describe in the following section. These approaches are not mutually exclusive, therefore we can combine them. For example, transfer learning---pre-training language representation models on unlabelled data, and then fine-tuning on target labelled data---has become a standard practice in NLP. Methods belonging to other approaches can be combined with transfer learning, such as using off-the-shelf pre-trained models as the baseline model. 

%% file: src/ch3-transfer-learning.tex
The standard supervised learning requires sufficient labelled data to train a decent performing model, given a particular task, domain and language. In other words, each model is trained individually for a combination of task, domain and language. In contrast, transfer learning explores the relatedness between tasks, domains and languages. The knowledge gained in solving a \emph{source} task in a \emph{source} domain and a \emph{source} language is applied to solve the \emph{target} task in the \emph{target} domain and \emph{target} language~\citep{Ruder:PhD:2019}. 

\begin{figure}[p]
	\centering
	\begin{subfigure}{.45\textwidth}
		\centering
		\includegraphics[width=1\linewidth]{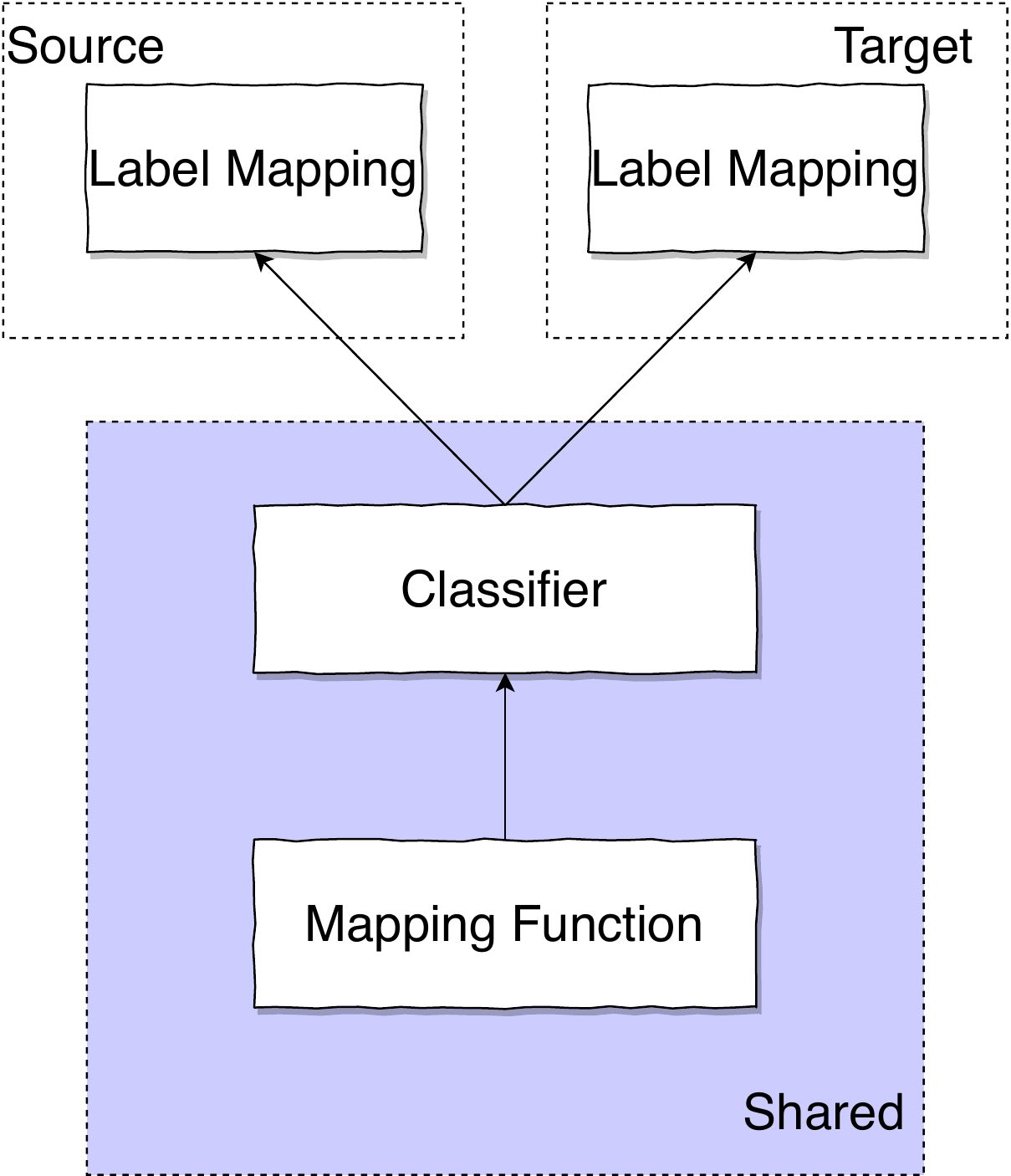}
		\caption{Cross-domain transfer when the label sets are mappable.}
		\label{figure-transfer-learning-cross-domain-a}
	\end{subfigure}%
	\hfill
	\begin{subfigure}{.45\textwidth}
		\centering
		\includegraphics[width=1\linewidth]{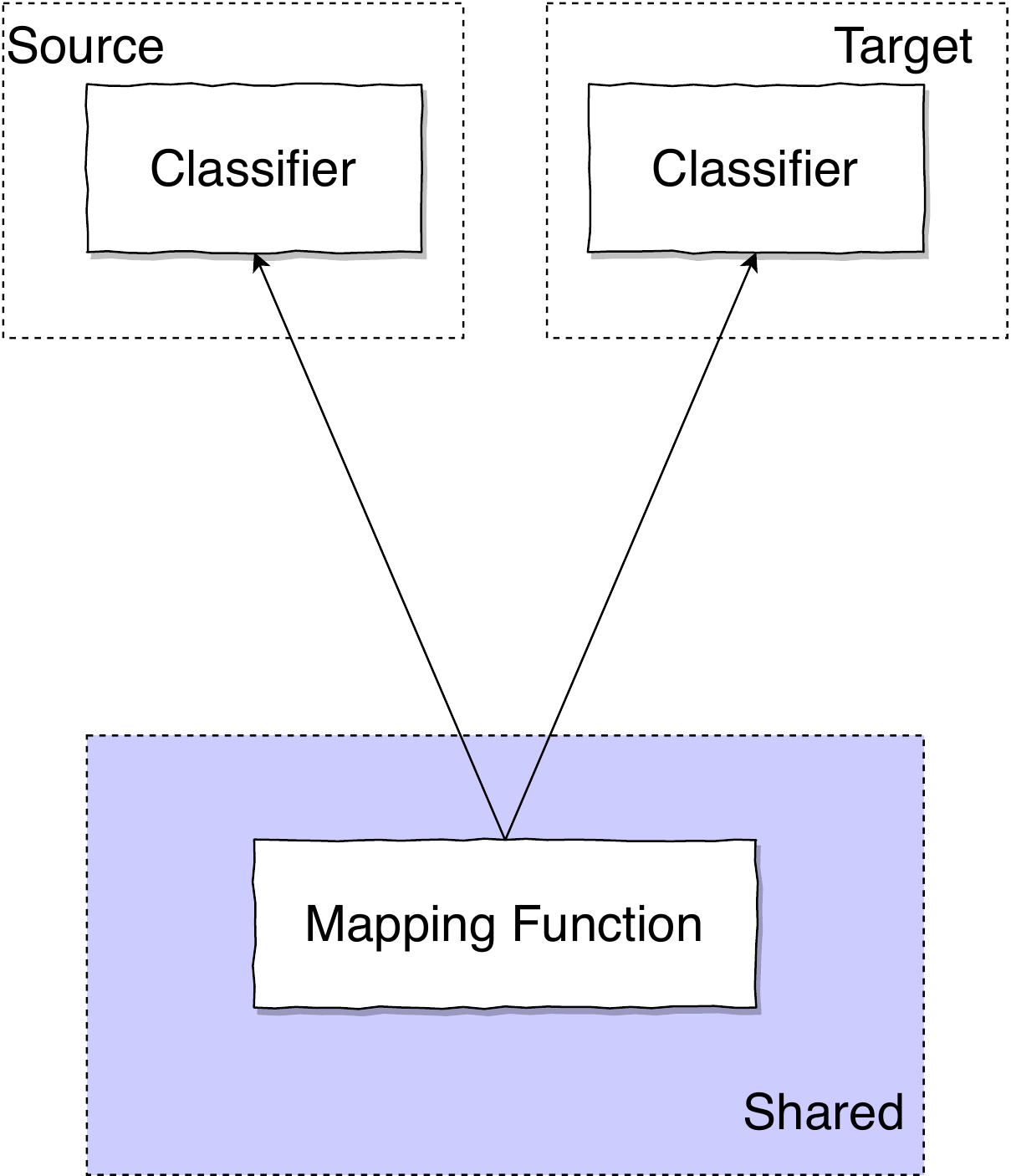}
		\caption{Cross-domain transfer when the label sets are disparate, and cross-task transfer.}
		\label{figure-transfer-learning-cross-domain-b}
	\end{subfigure}
	
	\begin{subfigure}{1\textwidth}
		\centering
		\includegraphics[width=1\linewidth]{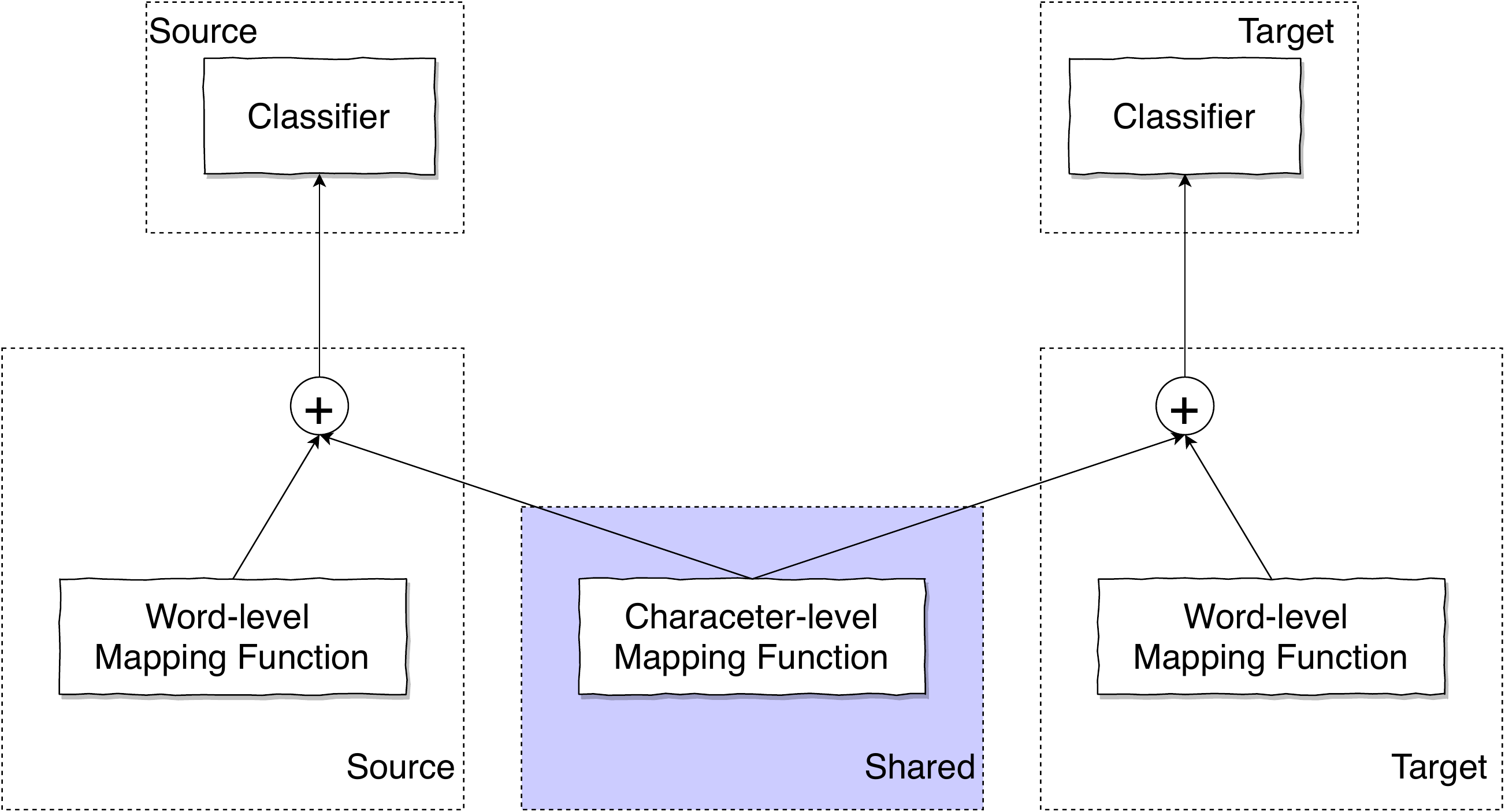}
		\caption{Cross-lingual transfer. The $\oplus$ also represents a neural network that takes both character-level and word-level feature vectors as input and creates the final token feature vector.}
		\label{figure-transfer-learning-cross-domain-c}
	\end{subfigure}
	\caption{Neural architectures for the settings of cross-domain, cross-task, cross-lingual transfer proposed in~\citep{Yang:Salakhutdinov:ICLR:2017}.}
\end{figure}

\citet{Yang:Salakhutdinov:ICLR:2017} develop a transfer learning approach for sequence tagging and design different neural architectures for cross-domain, cross-task, and cross-lingual transfer settings. In the cross-domain transfer, the authors share all parameters of the model---\textsc{BiLSTM-CRF}---and perform a label mapping on top of the classifier (Figure~\ref{figure-transfer-learning-cross-domain-a}). Note that, cross-domain transfer typically has mappable label sets that labels in different domains can be mapped to each other. For the unmappable setting, \citeauthor{Yang:Salakhutdinov:ICLR:2017} consider it the same as cross-task transfer, and each task learns a separate classifier (Figure~\ref{figure-transfer-learning-cross-domain-b}). The cross-lingual transfer is achieved by exploiting the morphologies shared by different languages. For example, the morphological similarity between \textit{`Canada'} in English and \textit{`Canad\'a'} in Spanish can be exploited for NER. The transfer learning architecture shares only the character level mapping function, which takes a sequence of characters as input, building a token feature vector (Figure~\ref{figure-transfer-learning-cross-domain-c}).
    
Although improvements have been reported by using cross-domain, cross-task, and cross-lingual transfers, a big challenge in these approaches is finding related source task, domain, and language. In other words, the knowledge learned in solving a source task in a source domain and language can be transferred, only if the knowledge is indeed shared between the source and target. In this thesis, we focus on English datasets; therefore, we do not discuss cross-lingual in details; we refer readers to ~\citep{Rahimi:Li:ACL:2019,Ruder:Vulic:JAIR:2019,Conneau:Khandelwal:ACL:2020} for more discussions.

\subsection{Cross-task transfer}
\label{section-embeddings-pretrain-tasks}
\citet{Patra:Moniz:EMNLP:2019} point out that it is easier and faster for an annotator to answer a yes/no question than to recognise all entity mentions. That is, the cognitive load of selecting whether an entity mention is present or not in the sentence is less than that of highlighting and annotating mentions with their entity categories. Therefore, they propose to use a model which is trained on a sentence level multi-label classification task---whether an entity mention is present or not---and transfer it to the entity recognition task. Evaluation results on \CONLL2003 show that the proposed method works surprisingly well, achieving $F_1$ score of 81.1.

\citet{Ruder:Bingel:AAAI:2019} propose a meta-architecture for multi-task learning. They use part-of-speech tagging---a fundamental syntactic task---as the auxiliary task, and observe that chunking, NER, and semantic role labelling tasks can benefit from the auxiliary task, outperforming the single task learning baseline. Similar ideas have also been explored by~\citet{Collobert:Weston:JMLR:2011,Sogaard:Goldberg:ACL:2016}. For example, \citet{Sogaard:Goldberg:ACL:2016} use \emph{low level} NLP task, such as part-of-speech tagging, to improve the \emph{higher level} tasks, such as chunking and CCG Supertagging. They design specialised multi level LSTM networks that have part-of-speech supervision at the innermost layer, and other tasks the outermost layer.

In contrast to transferring from a source task that requires labelled data, transfer learning techniques can make the most of limited labelled data by incorporating language representation models pre-trained on a large amount of unlabelled data~\citep{Mikolov:Chen:arXiv:2013,Pennington:Socher:EMNLP:2014,Peters:Neumann:NAACL:2018,Devlin:Chang:NAACL:2019}. Many pre-training tasks have demonstrated their effectiveness for different downstream tasks. In this section, we briefly review three pre-training tasks and refer readers to~\citep{Wang:Hula:ACL:2019} for more options.

\begin{figure}
    \centering
    \includegraphics[width=0.8\textwidth]{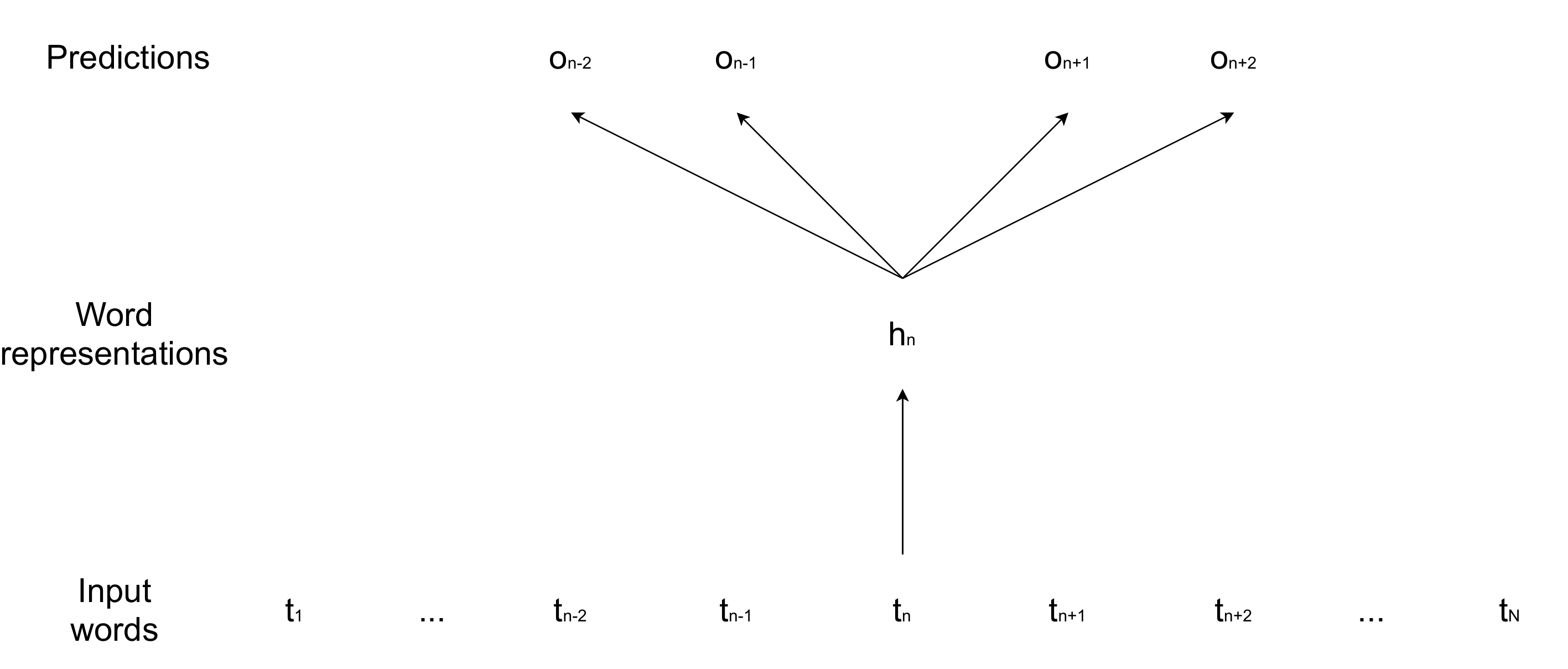}
    \caption{The Skip-gram model aims to learn word representations that can be used to predict the surrounding words.}
    \label{figure-transfer-learning-skip-gram}
\end{figure}

\paragraph{Skip-gram model}
The Skip-gram model, introduced by~\citep{Mikolov:Chen:arXiv:2013}, is an efficient method for learning static word representations from unlabelled text. The training objective of the Skip-gram model is to build word representations that can be used to predict the surrounding words in a sentence (Figure~\ref{figure-transfer-learning-skip-gram}). Given a sequence of tokens $\{t_i\}_{i=1}^N$, the Skip-gram model aims to maximise the objective:
\[
\frac{1}{N} \sum_{i=1}^{N} \sum_{-c \leq j \leq c, j \neq 0} \log p\left(t_{i+j} \mid t_{i}\right),
\]
where $c$, a hyper-parameter, is the size of the context window.

Skip-gram model has been shown to effectively capture syntactic and semantic information of words~\citep{Mikolov:Sutskever:NIPS:2013}. However, its main disadvantage is that it always assigns the same vector to the word, no matter what the context of the word is. The word representations learned using Skip-gram model are therefore called \emph{static} word representations.

\paragraph{Masked language modelling}
In contrast to \emph{static} word representations, where one word is always assigned the same vector, \emph{contextual} word representations can assign different vectors to the same word, depending on its context.

\begin{figure}
    \centering
    \includegraphics[width=0.95\textwidth]{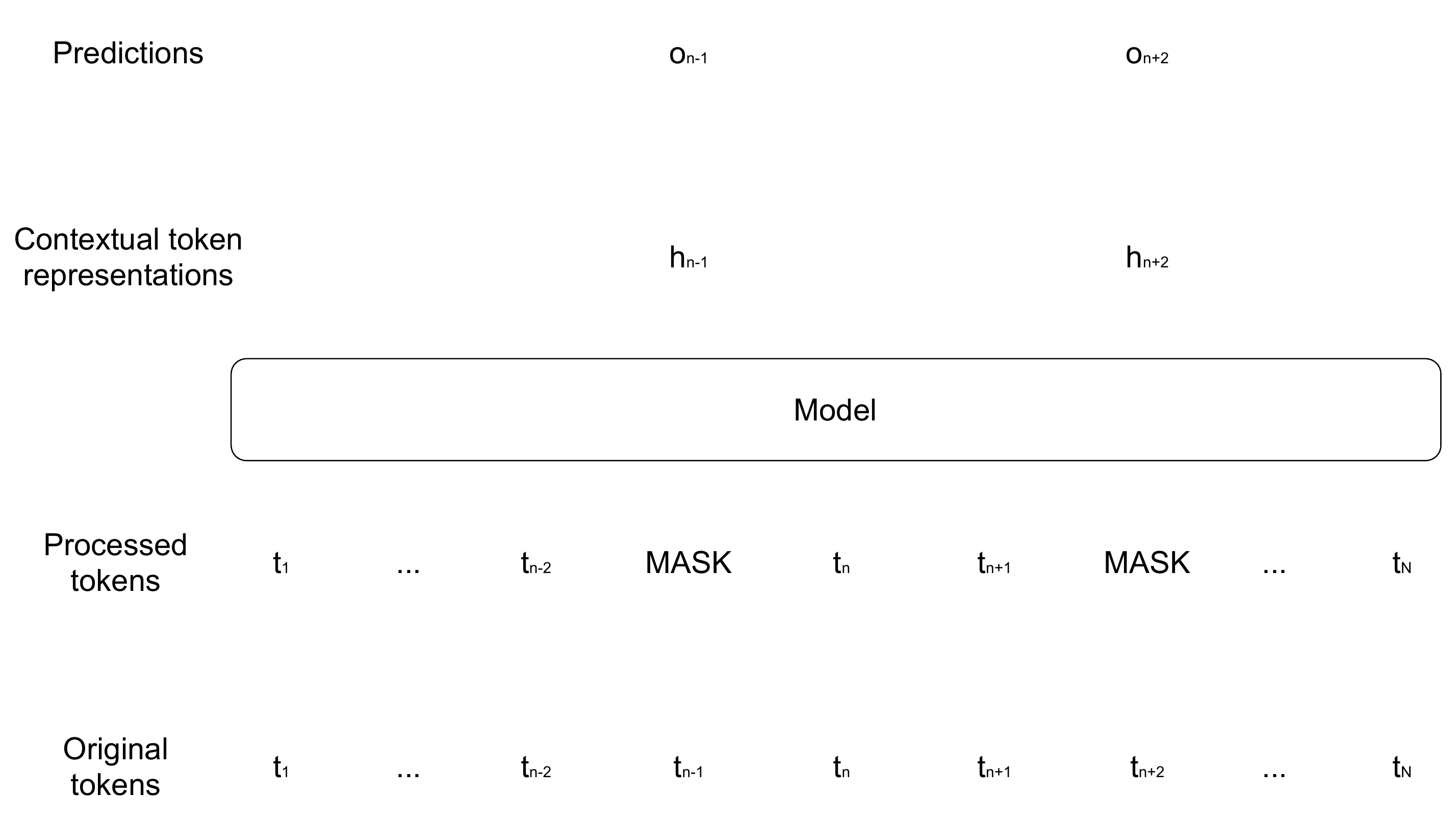}
    \caption{The mask language modelling pre-training task aims to learn contextual word representations that can be used to predict what the masked token is.}
    \label{figure-transfer-learning-masked-lm}
\end{figure}

\citet{Dai:Le:NIPS:2015} explored the idea of pre-training recurrent language model and transferring it to the downstream supervised models. They use unlabelled data from Amazon reviews to pre-train the language model and find that it can improve classification accuracy on the Rotten Tomatoes dataset. \citet{Peters:Ammar:ACL:2017} extend the single direction language model to bidirectional. Based on these efforts, \citet{Devlin:Chang:NAACL:2019} propose the masked language modelling pre-training task to better capture contexts from both sides. Different to~\citet{Peters:Neumann:NAACL:2018} who build two language models---left-to-right and right-to-left---which are trained separately, the masked language modelling is a fill-in-the-blank task. That is, a small set of tokens are masked, and the model needs to use the context tokens to try to predict what the masked tokens should be (Figure~\ref{figure-transfer-learning-masked-lm}). 

\begin{figure}
    \centering
    \includegraphics[width=0.95\textwidth]{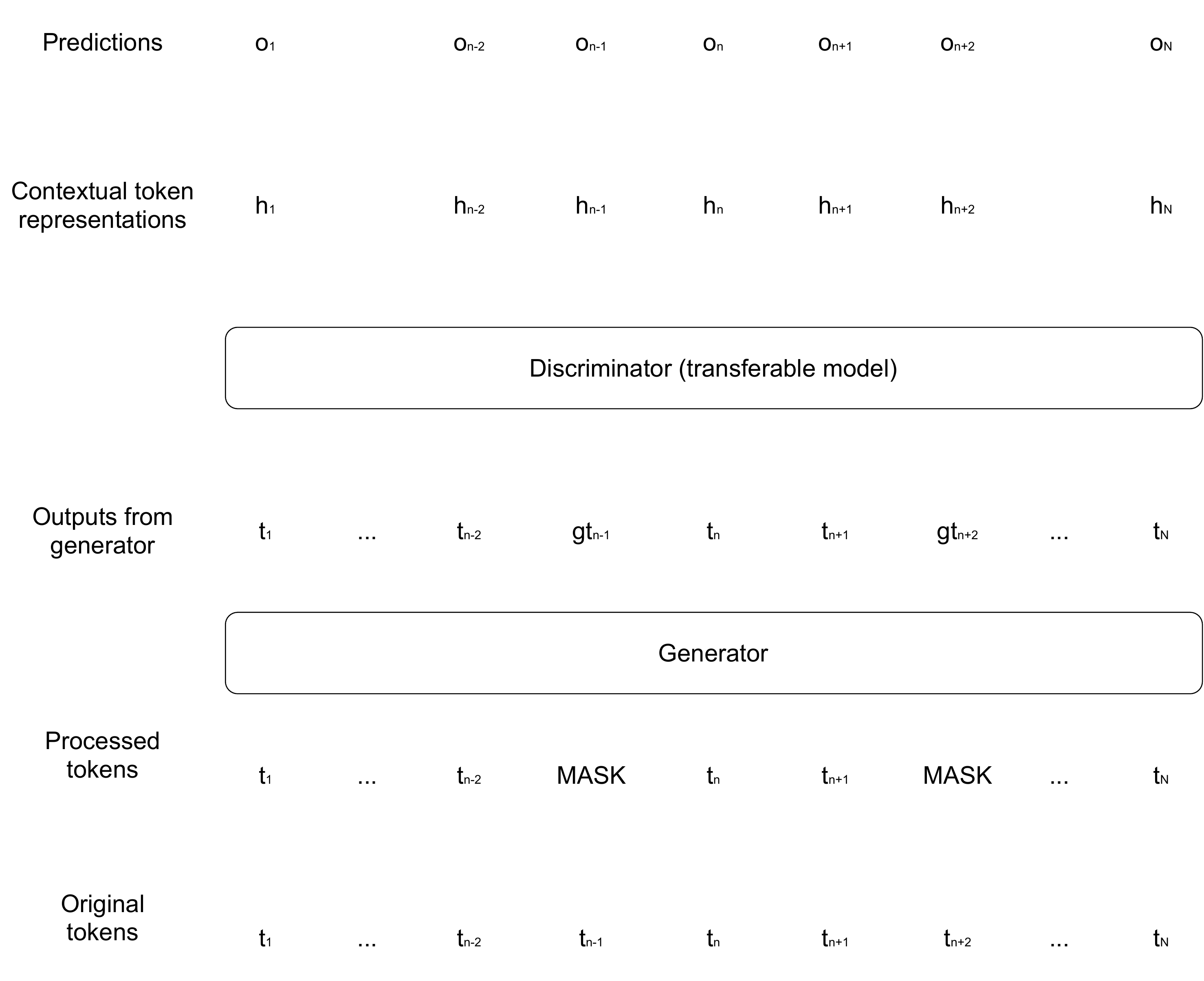}
    \caption{The replace token detection task aims to train the discriminator to predict whether the token is the original token or a fictional token.}
    \label{figure-transfer-learning-replaced-token-detection}
\end{figure}

\paragraph{Replace token detection}
The \emph{replace token detection} task, proposed by~\citet{Clark:Luong:ICLR:2020}, is a sample efficient variant of masked language modelling (Figure~\ref{figure-transfer-learning-replaced-token-detection}). Instead of replacing some tokens as the special [MASK] token, \citet{Clark:Luong:ICLR:2020} employ a small generator network to generate plausible alternatives. Then the discriminator network predicts whether each token in the input is replaced by a generator sample or not. 

Another difference between the masked language modelling and replace token detection pre-training tasks is that the former is performed on only masked tokens, whereas the latter is defined over all input tokens. Therefore, a replace token detection pre-training task requires less compute, measured using floating point operations. \citet{Clark:Luong:ICLR:2020} show that it performs comparably to masked language modelling pre-training task while using less than 1/4 of their compute and outperforms masked language modelling when using the same amount of compute. Because pre-training a language representation model using replace token detection task can be done within several days using a single GPU, we thus use it for our investigation in Chapter~\ref{chapter-select-pretraining-data}.

\subsection{Cross-domain transfer}

\citet{Han:Eisenstein:EMNLP:2019} propose \emph{domain-adaptive fine-tuning}, in which the language representation models are adapted by masked language modelling on text from the target domain. They evaluate this approach on sequence labelling in two challenge domains: Early Modern English and Twitter. Results show that domain-adaptive fine-tuning yields substantial improvements over strong BERT baselines, with particularly strong results on out-of-vocabulary words. Similarly, \citet{Gururangan:Marasovic:ACL:2020} investigate whether it is helpful to tailor a pre-trained model to the domain of a target task. They show that domain-adaptive pre-training---continue pre-training on a large corpus of unlabelled domain-specific text---leads to performance gains. Moreover, task-adaptive pre-training---continue pre-training on the unlabelled set for a given task---improves performance even after domain-adaptive pre-training. \citet{Gururangan:Marasovic:ACL:2020} consider domain vocabularies containing the top 10K most frequent uni-grams and use the vocabulary overlap as the measure of domain similarity. They find that pre-training data used in RoBERTa---over 160GB of uncompressed text, consisting of Wikipedia, books, stories, news articles, and web content extracted from URLs shared on Reddit---have a very low vocabulary overlap with datasets sampled from biomedical scholarly articles (27.3\%) and computer science scholarly articles (19.2\%).

\citet{Moore:Lewis:ACL:2010} propose a cross-entropy difference selection method to select in-domain training data to build auxiliary language models for use in tasks such as machine translation and speech recognition. Given the target data set $\mathcal{T}$ and a generic source $\mathcal{S}$, they aim to select a subset of the available source data as language model training data. Let $H_T(s)$ be the per-word cross-entropy, according to a language model trained on $\mathcal{T}$, of a sentence $s$ drawn from $\mathcal{S}$. Let $H_S(s)$ be the per-word cross-entropy of $s$ according to a language model trained on a random sample of $\mathcal{S}$. \citet{Moore:Lewis:ACL:2010} score the sentences from $\mathcal{S}$ according to $H_T(s) - H_S(s)$, and all sentences whose score is less than a threshold are selected as in-domain training data. A similar idea was explored by~\citet{Klakow:ICASSP:2000}, who estimates a language model from the
entire $\mathcal{S}$, and scores the subset of $\mathcal{S}$ by the change in the log likelihood of $\mathcal{T}$ according to another language model, where that subset is removed from training data. Those subsets whose removal would decrease the log likelihood of $\mathcal{T}$ more than a threshold are selected.

\citet{Plank:Noord:ACL:2011} evaluate measures of domain similarity and their impact on dependency parsing accuracy. Given a target article to parse and a collection of annotated articles, they want to select the most similar articles to train the parser which is then evaluated on the target article. Both probabilistically motivated similarity functions---such as Jensen-Shannon divergence, and skew divergence---and geometrically motivated distance functions---such as cosine, euclidean, and variational distance functions---are evaluated on different features. \citet{Plank:Noord:ACL:2011} find that comparing article topic distributions estimated by Latent Dirichlet Allocation (LDA)~\citep{Blei:Ng:JMLR:2003} using variational distance function or Jensen-Shannon divergence can effectively find the most similar source, and using these automatic measures can outperform using human annotated genre labels. In addition to above mentioned similarity measures, \citet{VanAsch:Daelemans:DANLP:2010} explore to use Rényi entropy and Bhattacharyya coefficient to estimate the impact of domain similarity on cross-domain transfer.


\subsection{Summary}
Inspired by these efforts that use domain similarity to nominate suitable data for labelling or training statistical language models. We explore whether these similarity measures can also be used to nominate in-domain data for pre-training large scale neural language representation models.

Our work is also inspired by several lines of work that aim to link the known to
the unknown, studying its impact. \citet{Ramponi:Plank:COLING:2020} study the implications of variations of language on model performance. They argue that treating text as just input data to machine learning is problematic, and it is important to study how covert and overt factors, such as genre, social-demographic aspect, stylistic and data sampling strategy, impact results, and take these factors into consideration in modelling and evaluation. \citet{Johnson:Anderson:ACL:2018} predict a system's accuracy using larger training data from its performance on much smaller pilot data. In Chapter~\ref{chapter-select-pretraining-data}, we aim to link the similarity between source pre-training data and target task data to the effectiveness of pre-trained models. In other words, we aim to design a cost-effective approach that predicts the usefulness of pre-trained models for target datasets based on the similarity between the source pre-training data and the target task data.




%% file: src/ch3-complex-ner.tex
Commonly, the NER problem is framed as: given a sequence of tokens, output a list of spans, each of which is an entity mention in text. Recall that a span is a consecutive sequence of tokens, or an individual token. The mention can therefore be represented using the starting and ending indices of the span: $I_s$, $I_e$. Additionally, each mention is assigned to an entity category. This perspective imposes two constraints: 
\begin{enumerate}[series=MyList] 
    \item An entity mention consists of a continuous sequence of tokens, where all the tokens indexed between $I_s$ and $I_e$ are part of the entity mention; and, 
    \item These linear spans do not overlap with each other. In other words, no token can belong to more than one entity mention.
\end{enumerate}

Most of the existing NER datasets in the generic domain, for example \CONLL 2003~\citep{Sang:Meulder:CONLL:2003}, or \ONTONOTES 5.0~\citep{Weischedel:Hovy:2011}, are annotated satisfying these two constraints. Therefore, conventional sequence taggers achieve state-of-the-art effectiveness in these datasets \citep{Lample:Ballesteros:NAACL:2016,Ma:Hovy:ACL:2016,Akbik:Blythe:COLING:2018,Baevski:Edunov:EMNLP:2019}. 

\begin{figure}[tb] 
\centering
\begin{subfigure}[b]{0.65\textwidth}
    \includegraphics[width=0.9\textwidth]{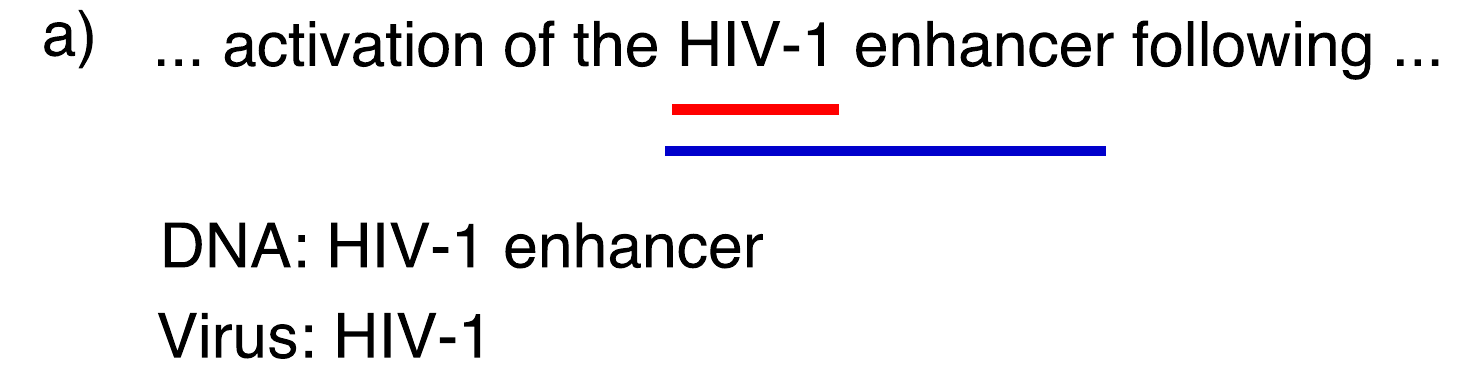}
\end{subfigure}
~

~

~
\begin{subfigure}[b]{0.68\textwidth}
    \includegraphics[width=0.78\textwidth]{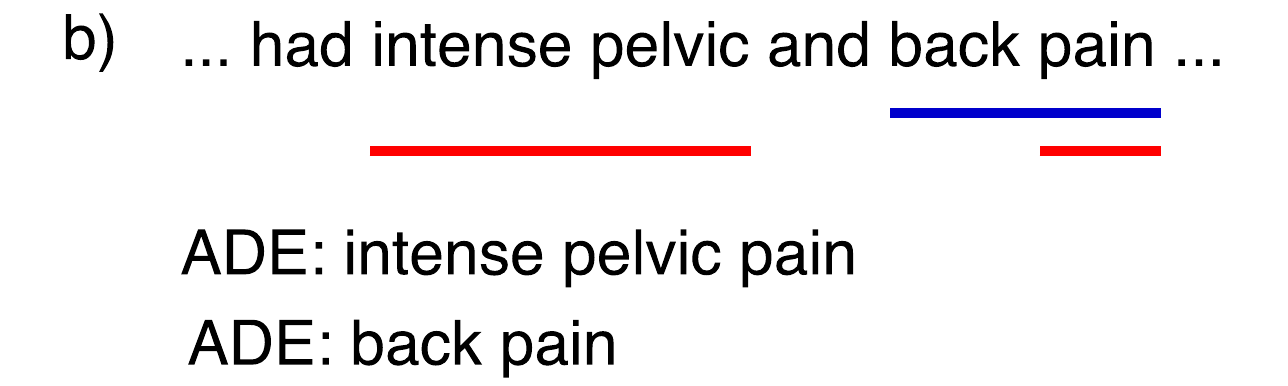}
\end{subfigure}
\caption{\label{figure-complex-ner-example}Examples involving nested, overlapping and discontinuous entity mentions. In (a), `\textit{HIV-1 enhancer}' and `\textit{HIV-1}' are nested entity mentions. In (b), `\textit{intense pelvic pain}' and `\textit{back pain}' overlap, and `\textit{intense pelvic pain}' is a discontinuous mention.}
\end{figure}

However, in practice, there are domains, such as the biomedical domain, in which there can be entity mentions nested, overlapping, and discontinuous (see examples in Figure~\ref{figure-complex-ner-example}). These \emph{complex entity mentions} cannot be directly recognised by conventional sequence taggers because they break the above mentioned constraints based on which sequence tagging techniques are built. 

In this section, we first describe these complex entity mentions in details (Section~\ref{section-complex-ner-definitions}). We then review the existing methods which are proposed to recognise complex entity mentions and categorise them into token-level (Section~\ref{section-complex-ner-token-level}), span-level (Section~\ref{section-complex-ner-span-level}), and sentence-level (Section~\ref{section-complex-ner-sentence-level}) approaches. Finally, we identify the research gap, that our proposed method (described in Chapter~\ref{chapter-discontinuous-ner}) is going to fill. 


\subsection{Definitions of complex entity mentions~\label{section-complex-ner-definitions}}
\paragraph{Nested entity mentions}
One entity mention is completely contained by the other. We call both of the mentions involved nested entity mentions. Figure~\ref{figure-complex-ner-example} a) is an example taken from the GENIA corpus \citep{Kim:Ohta:Bioinformatics:2003}. Here, `\textit{HIV-1 enhancer}' is a DNA mention, and it contains another mention `\textit{HIV-1}', which is a virus. 

\subparagraph{Multi-type entity mentions}
An extreme case of nested entity mentions is one in which a span corresponds to multiple mentions. For example, in the EPPI corpus~\citep{Alex:Haddow:BioNLP:2007}, proteins can also be annotated as drug/compound, indicating that the protein is used as a drug to affect the function of a cell. Such a mention should be classified as both protein and drug/compound. In this case, we consider this mention as two mentions of different categories, and these two mentions contain each other. 

\paragraph{Overlapping entity mentions}
Two entity mentions overlap, but neither is completely contained by the other. Figure~\ref{figure-complex-ner-example} b) is an example taken from the CADEC corpus \citep{Karimi:Metke-Jimenez:JBI:2015}, which is annotated for adverse drug events (ADE) and relevant concepts. In this example, two ADEs: `\textit{intense pelvic pain}' and `\textit{back pain}', share a common token `\textit{pain}', and neither is contained by the other. 

\paragraph{Discontinuous entity mentions}
The mention consists of a discontinuous sequence of tokens. In other words, the mention contains at least one interval. In Figure~\ref{figure-complex-ner-example} b), `\textit{intense pelvic pain}' is a discontinuous entity mention since it is interrupted by `\textit{and back}'. 

Recognising complex entity mentions is important because these mentions can hold very useful information~\citep{Ringland:Dai:ACL:2019}. First, the nested and overlapping structures themselves are already good indicators of the relationship between different entities involved. For example, an \textsc{Organisation} mention `\textit{University of Sydney}' contains a \textsc{Location} mention `\textit{Sydney}'. This structure implies the location of the organisation, and recognition of these mentions can potentially speed up the construction of a knowledge base~\citep{Ringland:Dai:ACL:2019}. Second, recognising complex entity mentions can simplify the design of downstream tasks. For example, separating overlapping mentions rather than identifying them as a single mention is important for a downstream entity linking task, where the assumption is that the input mention refers to one entity, and the task can thus be regarded as one-to-one mapping~\citep{Shen:Wang:TKDE:2014}. Third, recognising complex entity mentions can improve the performance of other NLP tasks. For example, entity mentions often have fixed representations in different languages. Therefore, recognising entity mentions, especially those discontinuous entity mentions, can improve the performance of a machine translation system~\citep{Klementiev:Roth:ACL:2006}. Last but not least, we notice that similar complex structures also exist in other NLP tasks, such as multi-word expressions recognition~\citep{Baldwin:Kim:2010,Rohanian:Taslimipoor:NAACL:2019} and constituency parsing~\citep{Coavoux:Crabbe:TACL:2019,Coavoux:Cohen:NAACL:2019}. We believe the ideas proposed for recognising complex entity mentions should also apply to similar complex structures in other tasks. 

\subsection{Token-level Approach~\label{section-complex-ner-token-level}}
\begin{figure}[tb] 
\centering
\includegraphics[width=0.9\textwidth]{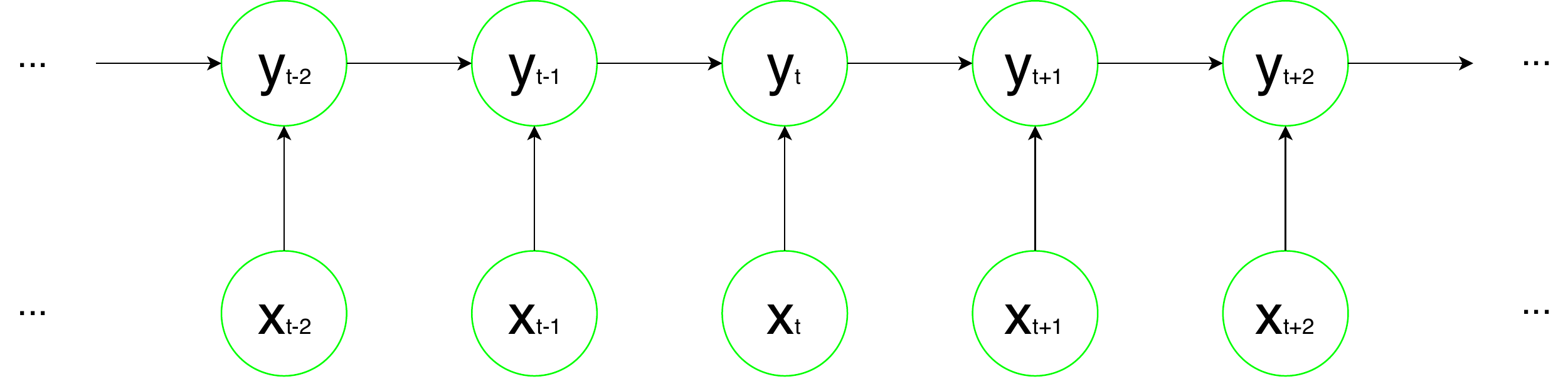}
\caption{In a linear-chain CRF model, the output for each token depends on the representation of that token in context and the output for the previous token.}
\label{figure-complex-ner-crf}
\end{figure}

The main component of sequence tagging techniques is a structural prediction model, which takes a sequence of contextual token representations as input and outputs a tag for each token. Figure~\ref{figure-complex-ner-crf} is an illustration of such a model. That is, in a linear-chain CRF model, the tag of one token depends on both the token representation and the tag of the previous token. These local decisions are then chained together to perform joint inference, and the tag sequence predicted by the tagger is finally decoded into entity mentions using explicit rules. We categorise the methods based on conventional sequence tagging as token-level approach. 

\begin{figure}[tb] 
\centering
\includegraphics[width=0.5\textwidth]{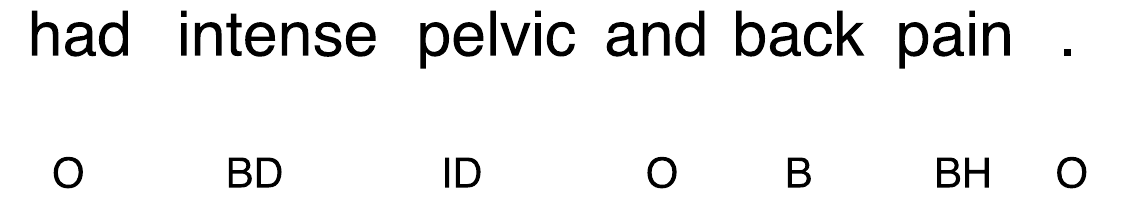}
\caption{\, An encoding example of two adverse drug event mentions: `\textit{intense pelvic pain}' and `\textit{back pain}'.}
\label{figure-complex-ner-bio-variant}
\end{figure}

In vanilla sequence tagging models, the intermediate outputs for each token are usually BIO schema tags. However, since the BIO tags cannot effectively represent complex entity mentions, the first natural direction is to expand the BIO tag set so that different kinds of complex entity mentions can be captured. \citet{Metke-Jimenez:Karimi:BMDID:2016} introduce a BIO variant schema to represent discontinuous and overlapping entity mentions. That is, in addition to the BIO prefixes, four new position indicators, BD, ID, BH, and IH are proposed to denote \textbf{B}eginning of \textbf{D}iscontinuous body, \textbf{I}nside of \textbf{D}iscontinuous body, \textbf{B}eginning of \textbf{H}ead, and \textbf{I}nside of \textbf{H}ead. Here, the token sequences which are shared by multiple mentions are called head, and the remaining parts of the mention are called body. Figure~\ref{figure-complex-ner-bio-variant} is an encoding example using this schema. `\textit{pain}' is the beginning of the head that is shared by two mentions, and therefore tagged as \textit{BH}. `\textit{intense pelvic}' is the body of a discontinuous mention, while `\textit{back}' is the beginning of a continuous mention. Here, we keep only the position indicator and remove the entity category `ADE', since this schema can only represent overlapping mentions of the same entity category. Note that, even in this simple example, it is still impossible to represent several mentions unambiguously. For example, this encoding can also be decoded as having three mentions: `\textit{intense pelvic pain}', `\textit{back pain}' and `\textit{pain}'. 

\begin{figure}[tb] 
\centering
\includegraphics[width=0.78\textwidth]{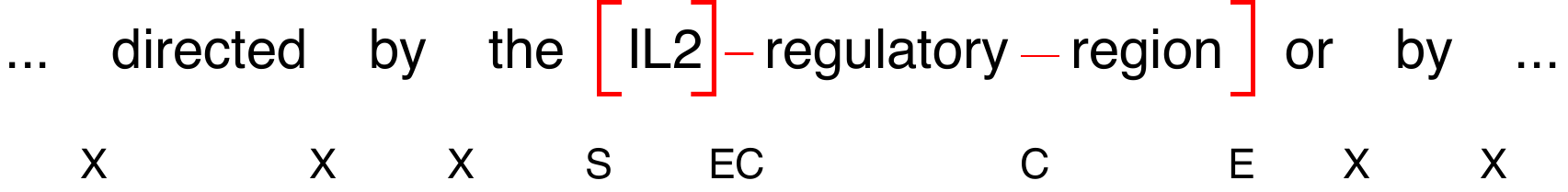}
\caption{\, An example of mention separators encoding two nested entity mentions: \textit{`IL2'} and \textit{`IL2 regulatory region'}. \citet{Muis:Lu:EMNLP:2017} design three mention separators: S, also denoted as \textcolor{red}{[}, indicating a mention is starting at the next token; E (\textcolor{red}{]}), indicating a mention is ending at the previous token; and C (\textcolor{red}{-}), indicating a mention is continuing to the next token. X means none of the three separators applies. The standard sequence tagger, which takes as input a sequence of $N$ tokens and outputs a sequence of $N$-1 mention separators, can be used to recognise nested NER.}
\label{figure-complex-ner-mention-separators}
\end{figure}

Tag variants are also proposed to deal with complex structures with specialised constraints. To deal with nested NER (one mention is completely contained by the other mention), \citet{Alex:Haddow:BioNLP:2007} propose a \emph{joined labelling} variant that each token is assigned a tag by concatenating the tags of all levels of nesting. For example, the token \textit{`HIV-1'} in Figure~\ref{figure-complex-ner-example} is assigned a tag `B-DNA+B-Virus', indicating that the token is the beginning token within a DNA mention and the beginning token within a \textsc{Virus} mention. Then the tagger is trained on the data containing the joined labels. During the inference stage, the joined labels are decoded into their original BIO format for each entity category. \citet{Rohanian:Taslimipoor:NAACL:2019} introduce BIOG tags for discontinuous structure without overlapping involved. The new G position indicator is used for tokens in between the components. \citet{Muis:Lu:EMNLP:2017} propose to assign tags to the gaps between tokens, while still regarding the problem as a sequence labelling problem. In other words, they model the mention boundaries instead of the role of tokens in forming mentions (Figure~\ref{figure-complex-ner-mention-separators}). 

Instead of elaborating schema to encode entity mentions with complex structures, another direction based on sequence tagging techniques is to employ multiple sequence tagging models or layers, that are arranged in a series. \citet{Alex:Haddow:BioNLP:2007} employ several sequence tagging models, each of which is used to recognise entity mentions belonging to a group of several entity categories without nested structures. Similarly, \citet{Ju:Miwa:NAACL:2018} stack several \bilstm-\crf layers together, each of which is used to recognise entity mentions belonging to a particular nesting layer. Note that, although these two methods achieve decent results in nested NER benchmarks, they have some difficulties in dealing with special nested structures. The cascade approach proposed by~\citet{Alex:Haddow:BioNLP:2007} cannot deal with nested mentions of the same entity category. For example, one DNA mention might contain another DNA mention. The layered approach proposed by~\citet{Ju:Miwa:NAACL:2018} cannot deal with multi-type entity mentions. For example, one mention might be annotated as both \textsc{Protein} and \textsc{Drug/Compound}. 

\subsection{Span-based Approach~\label{section-complex-ner-span-level}}
\label{section-ch3-span-based}
The vanilla idea of span-based approach enumerates all possible spans -- up to a certain length in a sentence -- as potential entity mentions. It then determines whether a span is a valid entity and what is its entity category. These candidate spans do not need to exclude each other, so the predicted entity mentions can also overlap with each other. This advantage makes a span-based approach a strong option for nested NER, and it has been extensively investigated. 

Given vector representations of each token $\dvector{h}_{i}$ in the sentence $ \dmatric{H} = \dvector{h}_{1}, \cdots, \dvector{h}_{i}, \cdots, \dvector{h}_{n} $ and a candidate span $(\dscalar{i}, \dscalar{j})$, the key decision of span-based approaches is how to build the span representation and score the span for each entity category. Once all candidate spans are represented as fix-length vectors and scored for each entity category, they are ranked by the scores and the top-ranked spans are outputted as the final predictions. 

A summary of representative techniques to build span representations and to score span for each entity category is shown in Table~\ref{litreivew-table-span-representation}. Building span representations via directly using boundary token representations as well as tokens within the span is the simplest solution. \citet{Sohrab:Miwa:EMNLP:2018} represent the span by concatenating the boundary token representations and the average of all token representations within the span ($\frac{1}{j-i+1} \sum_{k=i}^{j} \dvector{h}_{k}$). The span representation is then passed to a softmax output layer to classify the span into a specific entity category or non-entity. Similarly, \citet{Luan:He:EMNLP:2018,Luan:Wadden:NAACL:2019} construct span representation by concatenating the boundary token representations ($\mathbf{h}_{i}$ and $\mathbf{h}_{j}$), an attention-based soft \emph{headword}, and embedded span width features, and then use a feed-forward neural network to produce per-class scores for span. 


\begin{table}[tb]
	\small
	\centering
	\begin{tabular}{r | l }
		\toprule
\bf Model & \bf Representing and scoring spans \\
		
\midrule
		
\citep{Xu:Jiang:ACL:2017} & 

$\begin{aligned}
\dvector{h}(i,j) & = \begin{bmatrix} f(1,i-1) \\ f(1,j) \\ f(i,j) \\ f(i,n) \\ f(j + 1,n) \end{bmatrix} \\
& \text{where } f(i,j) = \left\{ \begin{array}{l}
\mathbf{h}_{i} \qquad \qquad \qquad \qquad \text{if } i = j \\
\alpha \cdot f(i,j-1) + \mathbf{h}_{j} \; \; \text{ otherwise}
\end{array}\right.   \\

\text{score}(i,j) & = \, \softmax (\dmatric{W} \cdot \dvector{h}(i,j))
\end{aligned}$ 
		
\\ \hline
		
\citep{Luan:He:EMNLP:2018,Luan:Wadden:NAACL:2019} & 
$\begin{aligned}
\dvector{h}(i,j) & = \begin{bmatrix} \dvector{h}_i \\ \dvector{h}_j \\\textsc{Self Attention}(\dmatric{H}) \\ \textsc{Span width feature} \end{bmatrix} \\

\text{score}(i,j) & = \dmatric{W} \cdot \dvector{h} (i, j)
\end{aligned}$ 
		
\\ \hline

\citep{Sohrab:Miwa:EMNLP:2018} & 
$\begin{aligned}
\dvector{h}(i,j) & = \begin{bmatrix} \dvector{h}_i \\ \frac{1}{j-i+1} \sum_{k=i}^{j} \dvector{h}_k \\ \dvector{h}_j \end{bmatrix} \\
\text{score}(i,j) & = \softmax \Big( \dmatric{W} \cdot \dvector{h}(i,j) \Big)
\end{aligned}$

\\ \hline
		
\citep{Xia:Zhang:ACL:2019} & 
$\begin{aligned}
\dvector{e} & = \bilstm (\dvector{h}_i \cdots \dvector{h}_j) \\
\dvector{a} & = \softmax \big( \dmatric{H}\dmatric{W}\dvector{e}^T \big) \\
\dmatric{C} & = \dvector{a} \star \dmatric{H} \\
\dvector{m} & = \bilstm (\dmatric{C}) \\
\dvector{h}(i,j) & = \begin{bmatrix} \dvector{m} \\ \dvector{e} \end{bmatrix} \\
\text{score}(i,j) & = \softmax \Bigg(\dmatric{W}_2 \cdot \bigg(\sigma \Big(\dmatric{W}_1 \cdot \dvector{h}(i,j) + \dvector{b}_1 \Big) \bigg) + \dvector{b}_2 \Bigg)
\end{aligned}$ 
		
\\ \hline		

\citep{Yu:Bohnet:ACL:2020} & 
$\begin{aligned}
\dvector{h}_s  & = \ffnn_{s} (\dvector{h}_{i} ) \\
\dvector{h}_e & = \ffnn_{e} (\dvector{h}_{j} ) \\
\text{score}(i,j) & = {\dvector{h}_s}^{\top} \dmatric{W}_1 \dvector{h}_e + \dmatric{W}_2 \cdot \big(\dvector{h}_s \oplus \dvector{h}_e \big) + \dvector{b}
\end{aligned}$ 
		
\\ \bottomrule
	\end{tabular}
	\caption{\, A summary of techniques to represent and score span, given a sequence of token representations $\dvector{h}_{1}, \cdots, \dvector{h}_{n} $. $\dvector{h}(i,j)$, being a fixed-length vector representation of the span, with its dimension being a hyper-parameter. $\text{score}(i,j)$ is the (normalised) score for the span from $i$ to $j$ inclusive, where $1 \leq i \leq j \leq n$. $\text{score}(i,j)$ is usually a $c$-dimension vector, where $c$ is the number of entity categories, including a special category for non-entity.~\label{litreivew-table-span-representation}}
\end{table}

\citet{Xu:Jiang:ACL:2017} employ Fixed-sized Ordinarily Forgetting Encoding (FOFE) to encode the span and its contexts into a fixed-size vector and then use a feed-forward neural network to predict its entity category. FOFE mimics bag-of-words but incorporates a forgetting factor ($\alpha$ in Table~\ref{litreivew-table-span-representation}) to capture positional information~\citep{Zhang:Jiang:ACL:2015}. \citeauthor{Xu:Jiang:ACL:2017} create both word-level and character-level features for each span and its left and right contexts: FOFE code of the span ($f(i,j)$); FOFE code for left context including the span ($f(1,j)$), FOFE code for left context excluding the span ($f(1,i-1)$); FOFE code for right context including the span ($f(i,n)$); FOFE code for right context excluding the span ($f(j+1,n)$). 

\citet{Yu:Bohnet:ACL:2020} argue the contexts of the start and end of the span are different. They apply two separate feed-forward neural networks to create different boundary representations ($\dvector{h}_s$ and $\dvector{h}_e$, in Table~\ref{litreivew-table-span-representation}) for the start and end of the span. Then they use a biaffine model~\citep{Dozat:Manning:ICLR:2017} to score the span. \citet{Xia:Zhang:ACL:2019} run an additional \bilstm on top of the token representations and use an attention mechanism to let tokens within the span attend to contexts to get the span representation. Finally, a two-layer feed-forward neural network is used to score the span. 


One shortcoming of span-based approaches is that all candidate spans are scored independently. The exhaustive enumeration of possible spans creates a large number of negative instances. That is, the majority of candidate spans belong to a non-entity category. Also, interactions among mentions are not explored, because all span representations are built in parallel on top of the same underlying token representations. 

We describe efforts on overcoming this problem from different perspectives: 


\paragraph{Solving class imbalance problem}
Given a sequence of $\dscalar{n}$ tokens, if we enumerate all possible spans in the sentence, there are in total $\frac{n \times (n + 1)}{2}$ candidates spans. These candidate spans belong to one of three categories:
\begin{enumerate*}
	\item exact match with a gold entity mention;
	\item partial overlap with a gold mention; and, 
	\item disjoint with any mention.
\end{enumerate*}
The latter two (negative instances) significantly outnumber the first exact match ones (positive instances).
This class imbalance problem may result in low predictive accuracy. 

To solve this problem, \citet{Xu:Jiang:ACL:2017} and \citet{Xia:Zhang:ACL:2019} use a down-sampling strategy. That is, they fix the total number of candidate spans in each training batch, including all positive spans and sampled negative spans. 

\citet{Sun:Ji:IEEE:2019} remove those negative spans that highly overlap with spans corresponding to gold mentions. The negative span $b$ is used for training, only if
\begin{equation}
    \max \bigg( \bigg[ \textnormal{IoU} (b, g) \, \textnormal{for} \, g \, \textnormal{in} \, G \bigg] \bigg) \leq \Gamma,
\end{equation}
where $G$ is the set of gold entity mentions. $\textnormal{IoU} (b, g)$ is a function measuring how many tokens are shared between two spans:
\begin{equation}
    \textnormal{IoU} (b, g) = \frac{\textnormal{length} (b \cap g)}{\textnormal{length} (b \cup g)}
\end{equation}
and $\Gamma$ is a hyperparameter tuned on different datasets. 

\paragraph{Reducing search space}
Instead of exhaustive classifying over all possible spans, a two-stage paradigm is investigated to reduce the size of candidate mentions. \citet{Zheng:Cai:EMNLP:2019} propose a boundary-aware model, where first sequence labelling models are used to detect possible span boundaries, and then the span based models are used to predict entity categories of a small number of candidate spans. Similarly, \citet{Xia:Zhang:ACL:2019} separate the task into two stages: deciding whether the candidate span is an entity mention or not via a detector, and then classifying detected candidates into predefined entity categories via a classifier. 

\citet{Lin:Lu:ACL:2019} detect entity mentions by using what they call \emph{head-driven phrase structures}. They first identify possible head words of entity mentions, and then recognise the mention boundaries by exploiting phrase structures. They argue that although entity mentions might nest with each other, they cannot share the same head words, and the head words are informative to decide the entity category. They also propose an objective function---bag loss---which does not require gold head word annotations. This is done by exploiting the association between words and entity categories. 

\paragraph{Modelling surrounding mentions}
To take the surrounding mentions of a given span into consideration, \citet{Xu:Jiang:ACL:2017} introduce a \emph{2nd-pass} mechanism. They train two models: one standard model, and the other model using outputs from the first model, where the predicted entity categories are used to replace the entity mentions. During inference, the span score is the linear interpolation between scores from these two models. 

\citet{Luan:He:EMNLP:2018} propose a multi-task learning framework where entity recognition, relation extraction, and coreference resolution are treated as classification problems with shared span representations. By sharing low-level LSTM encoder, information about relation types with surrounding mentions and coreferences can be used to create input span representations to entity classifier. Instead of sharing only LSTM encoder, \citet{Luan:Wadden:NAACL:2019} further extend the multi-task model using dynamically constructed span (node) graphs. At each training step, the most confident entity spans are treated as nodes in a graph structure, and arcs are confidence-weighted relation types and coreferences. Then, the span representations are refined using updates which are propagated from neighbouring relation types and co-refeerred entities.


\subsection{Sentence-level Approach~\label{section-complex-ner-sentence-level}}
Instead of predicting whether a token belongs to an entity mention and its role in the mention (token-level approach) or whether a consecutive sequence of tokens form an entity mention (span-level approach), some methods predict directly a combination of entity mentions within a sentence. We call these methods sentence-level approach. 

\begin{figure}[tb] 
\centering
\includegraphics[width=0.55\textwidth]{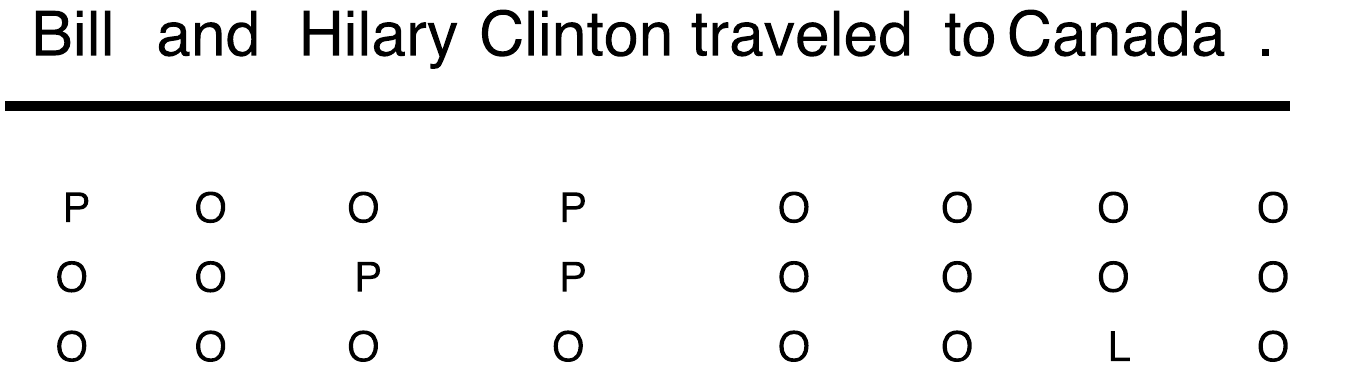}
\caption{\, An example of a sentence with three entity mentions: \textit{`Bill Clinton'} and \textit{`Hilary Clinton'} are \textsc{Person} mentions, and \textit{`Canada'} is a \textsc{Location} mention. P and L refer to the entity categories: \textsc{Person} and \textsc{Location}, respectively.}
\label{figure-complex-ner-multilabel}
\end{figure}

\citet{McDonald:Crammer:EMNLP:2005} consider NER as a structured multi-label classification. Instead of starting and ending indices, they represent each entity mention using the set of token positions that belong to the mention. An example of this representation, with each token tagged using an I/O schema is shown in Figure~\ref{figure-complex-ner-multilabel}. The advantage of this method is that the representation is very flexible as it allows entity mentions consisting of discontinuous tokens and does not require mentions to exclude each other. Using this representation, the NER problem is converted into the multi-label classification problem of finding up to $k$ correct labels among all possible $(T+1)^n$ labels, where $k$ is a hyper-parameter of the model, $T$ is the number of entity categories, and $n$ is the length of the sentence. Note that labels do not come from a pre-defined category but depend on the sentence being processed. \citeauthor{McDonald:Crammer:EMNLP:2005} use large-margin online learning algorithms to train the model, so that the scores of the correct labels (entity mentions) are higher than those of all other possible incorrect mentions. Another advantage of this method is that the outputs of the model are unambiguous for all kinds of complex entity mentions and easy to be decoded. However, the method suffers from a $O(n^3T)$ inference complexity. 

\citet{Finkel:Manning:EMNLP:2009} use a discriminative constituency parser to recognise nested entity mentions. They represent each sentence as a constituency tree, where each mention corresponds to a phrase in the tree. In addition, each node needs to be annotated with its parent and grandparent labels, so that the parser can learn how entity mentions nest. \citet{Ringland:PhD:2016} also employ a joint model using the Berkeley parser \citep{Petrov:Barrett:ACL:2006}, and show that it performs well even without specialised NER features. However, one disadvantage of these parsing based models, as in~\citep{McDonald:Crammer:EMNLP:2005}, is that their time complexity is cubic in the number of tokens in the sentence. Furthermore, the high quality parse training data, which is not always available, plays a crucial role in the success of the joint model~\citep{Li:Dong:EMNLP:2017}. 

\begin{figure}[tb] 
\centering
\includegraphics[width=0.55\textwidth]{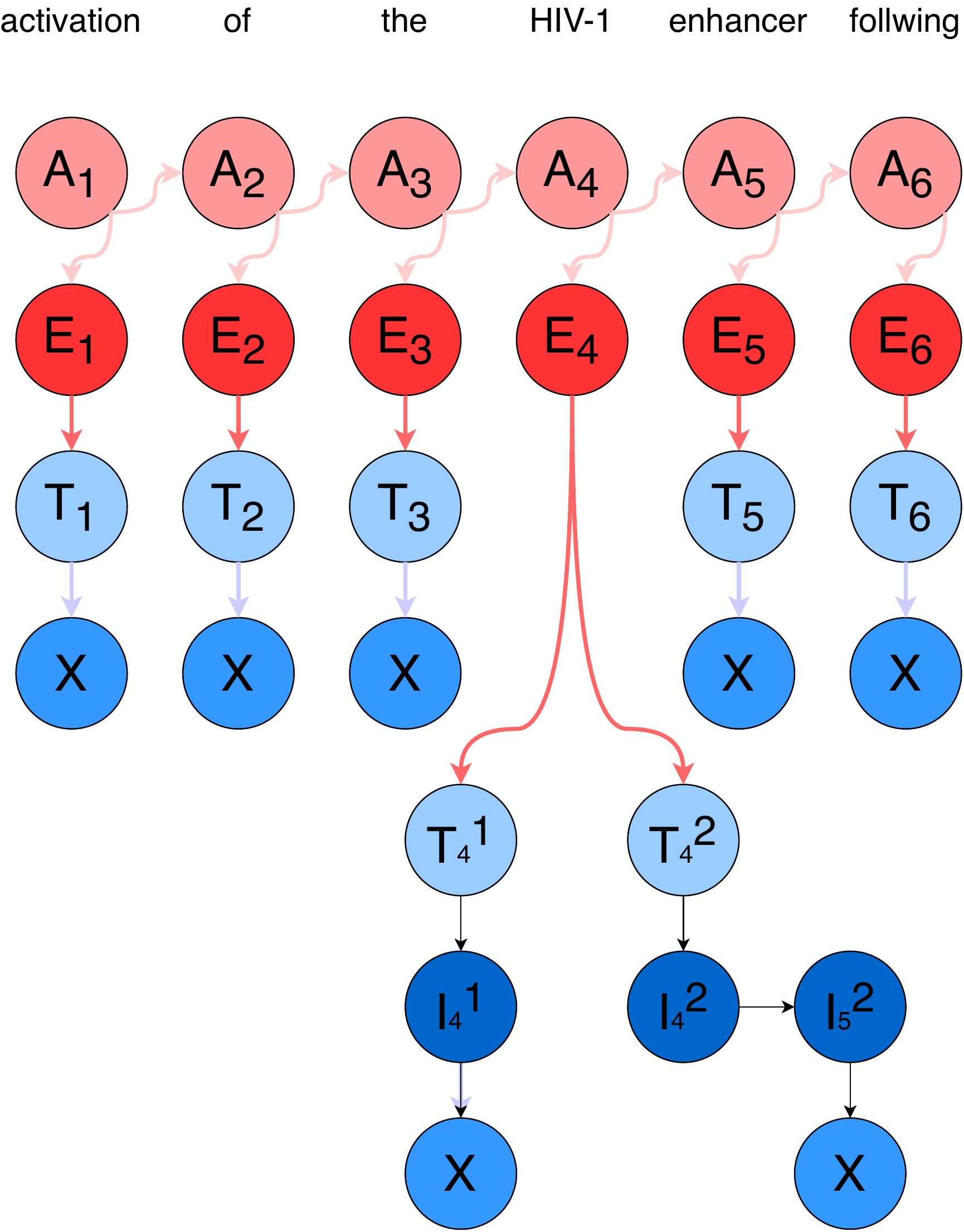}
\caption{\, An example sub-hypergraph with two nested entity mentions: `\textit{HIV-1}' (\textsc{Virus}) and `\textit{HIV-1 enhancer}' (\textsc{DNA}). Here, one mention corresponds to a path consisting of (AETI+X) nodes. Specifically, the path (A$_4$E$_4$T$_4^1$I$_4^1$X) corresponds to the mention \textit{`HIV-1'}, and the path (A$_4$E$_4$T$_4^2$I$_4^2$I$_5^2$X) corresponds to the mention \textit{`HIV-1 enhancer'}.\label{figure-complex-ner-hypergraph}}
\end{figure}

\citet{Lu:Roth:EMNLP:2015} propose a novel hypergraph to represent exponentially many possible nested mentions in one sentence, and one sub-hypergraph of the complete hypergraph can therefore be used to represent a combination of mentions in the sentence. The mention hypergraph consist of five types of nodes:
\begin{description}
\item[$A^k$ nodes] represent all mentions whose left boundaries are exactly at or after the $k$-th token;
\item[$E^k$ nodes] represent all mentions whose left boundaries are exactly at the $k$-th token;
\item[$T^k_j$ nodes] represent all mentions whose left boundaries are exactly at the $k$-th token and have the mention type $j$;
\item[$I^k_j$ nodes] represent all mentions which contain the $k$-th token and have the mention type $j$; and
\item[$X$ nodes] indicate the completion of a path.
\end{description}
Hyper-edges, each of which consists of a parent node and an ordered list of child nodes, are used to connect nodes. Figure~\ref{figure-complex-ner-hypergraph} is an example of such a sub-hypergraph, which represents two nested entity mentions. 

The training objective of this hypergraph-based model is to maximise the log-likelihood of training instances consisting of the sentence and mention-encoded hypergraph. During inference, the model first predicts a sub-hypergraph among all possible sub-hypergraphs of the complete hypergraph, and predicted mentions can be decoded from the output sub-hypergraph. Different to~\citet{Lu:Roth:EMNLP:2015} who build hand-crafted features defined over the input sentence and the output hypergraph structure, \citet{Katiyar:Cardie:NAACL:2018} learn the hypergraph representation using features extracted from a recurrent network. 

Although this hypergraph-based model enjoys a time complexity that is linear in the number of tokens in the input sentence, it suffers from some degree of ambiguity during decoding stage. For example, when one mention is contained by another mention with the same entity category and their boundaries are all different, the hypergraph can be decoded in different ways. This ambiguity comes from the fact that, if one node has multiple parent nodes and multiple child nodes, there is no mechanism to decide which of the parent node is paired with which child node. Therefore, \citet{Wang:Lu:EMNLP:2018-hypergraph} propose an extension of the I node where they use $I^k_{i,n}$ nodes to represent all mentions of type $k$ which contain the $j$-th token and start with the $i$-th token. 

\begin{figure}[tb] 
\centering
\includegraphics[width=0.6\textwidth]{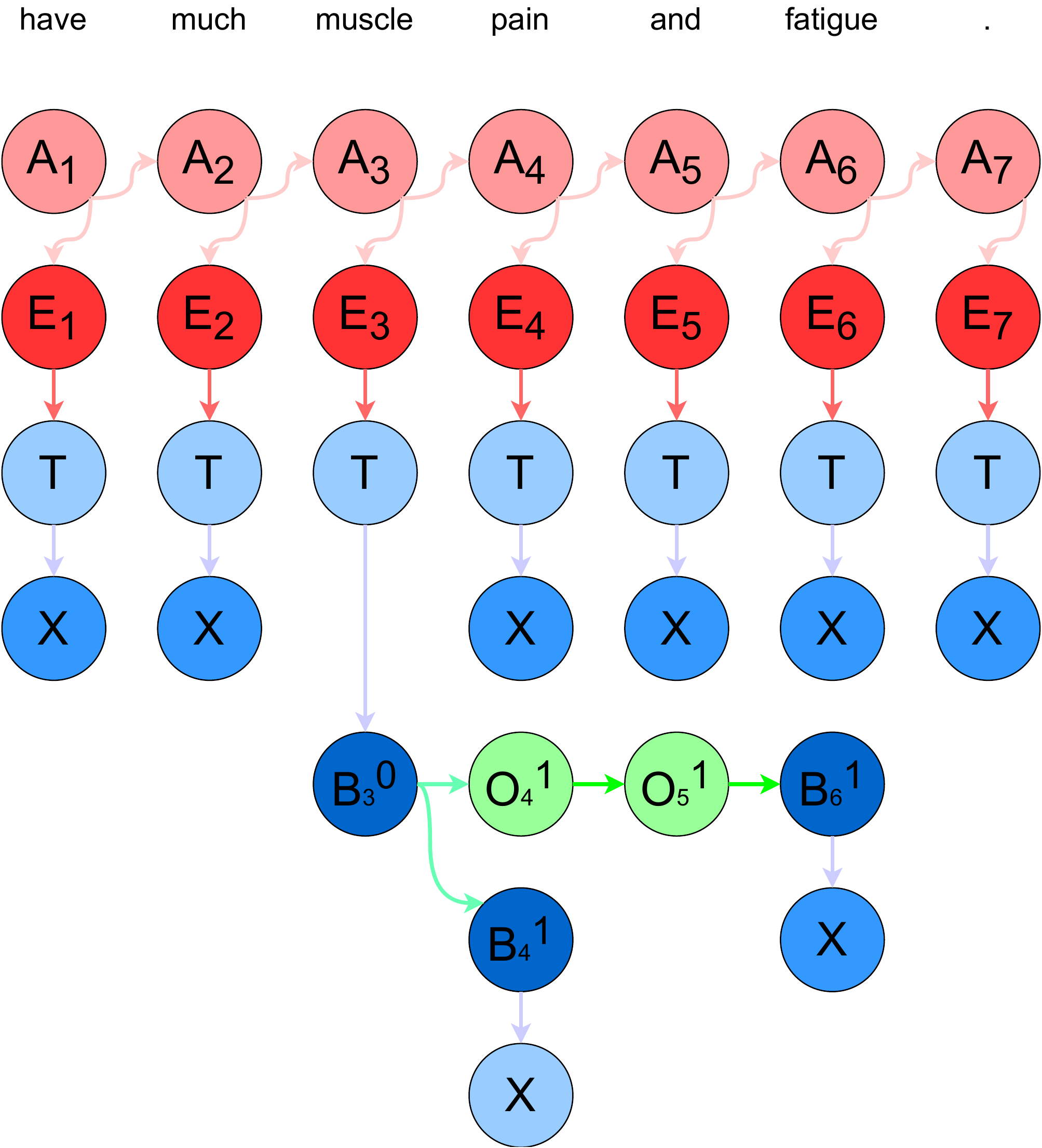}
\caption{\, An example sub-hypergraph with two entity mentions: \textit{`muscle pain'} and \textit{`muscle fatigue'}. \citet{Muis:Lu:EMNLP:2016} extend the hypergraph representation proposed by~\citet{Lu:Roth:EMNLP:2015} to capture discontinuous mentions through two new node types: B$_k^i$ representing the $k$-th token is part of the $i$-th component of an entity mention, and O$_k^i$ representing the $k$-th token appears in between $(i-1)$-th and $i$-th components of an entity mention. In this example, the path (A$_3$E$_3$TB$_3^0$B$_4^1$X) corresponds to the mention \textit{`muscle pain'} and the path (A$_3$E$_3$TB$_3^0$O$_4^1$O$_5^1$B$_6^1$X) corresponds to the discontinuous mention \textit{`muscle fatigue'}.\label{figure-complex-ner-discontinuous-hypergraph}}
\end{figure}

To represent discontinuous mentions, \citet{Muis:Lu:EMNLP:2016} expand the node types in the hypergraph representation to capture discontinuous mentions. That is, they add two new node types: B for tokens within the mention, and O for tokens belonging to part of the gap. Figure~\ref{figure-complex-ner-discontinuous-hypergraph} is an example of the sub-hypergraph, which encodes two mentions: \textit{`muscle pain'} and \textit{`muscle fatigue'}. \citet{Wang:Lu:EMNLP:2019} propose a two-stage approach that all spans are first identified using the hypergraph representation and then a classifier is used to predict whether two spans form a discontinuous mention. 

\subsection{Summary}
Despite the potential applications of complex NER recognition, there is comparatively few studies on recognising discontinuous and overlapping mentions. The span-based approach focuses on building effective span representations which are used to predict whether the span is an entity mention and its entity category. However, it cannot be directed used for discontinuous NER, because discontinuous mentions consist of multiple spans. 

We observe that there is usually a trade-of between the expressiveness power and the modelling difficulty. In other words, the more flexible (less constraints) the representation is, the more interactions are ignored, therefore the model might be more difficult to train. For example, the multi-label representation proposed by~\citet{McDonald:Crammer:EMNLP:2005} is the most flexible representation (Figure~\ref{figure-complex-ner-multilabel}); however, it does not take any interactions between different mentions into consideration. \citet{Tang:Hu:Wireless:2018} empirically show that a less flexible representation---BIO variant tagging based model---can outperform the multi-label representation when the training data are limited. 

Methods belonging to token-level and sentence-level approaches first predict intermediate representations, and then the intermediate representations are decoded into entity mentions. That is, token level representations use a sequence of tags~\citep{Tang:Cao:BMC:2013,Metke-Jimenez:Karimi:BMDID:2016} as the intermediate representation, and sentence-level approach uses a graph structure~\citep{Lu:Roth:EMNLP:2015,Katiyar:Cardie:NAACL:2018}. Inspired by these approaches, in Chapter~\ref{chapter-discontinuous-ner}, we propose a transition-based model for discontinuous NER. Similar to token-level and sentence-level approaches, our transition-based model first predicts the intermediate representation, i.e., a sequence of actions, but greatly reduce the ambiguity problem in these two approaches. Our method also employs effective methods to build span representations, which are inspired by span-based approach. 

%% file: src/ch4-data-augmentation.tex
Data augmentation, expanding the training set by transforming training instances without changing their labels, is heavily studied in the field of computer vision and sentence level NLP tasks. Inspired by these efforts, we design several easy to use data augmentation methods for NER. Through experiments on two English datasets from the biomedical domain, we demonstrate that our proposed augmentation methods can boost performance over a strong baseline where large scale pre-trained models are used, especially when the original labelled training set is small.

\section{Overview}
Modern deep learning techniques typically require a large amount of labelled data for training~\citep{Bowman:Angeli:EMNLP:2015,Conneau:Schwenk:EACL:2017}. However, in real world applications, such large labelled data sets are not always available. This is especially true in some specific domains, such as the biomedical domain, where annotating data requires expert knowledge and is usually time-consuming~\citep{Karimi:Metke-Jimenez:JBI:2015,Nye:Li:ACL:2018}. 

Different approaches have been investigated to solve this \emph{low-resource} problem~\citep{Hedderich:Lange:arXiv:2020}. For example, transfer learning pre-trains language representations on self-supervised or rich-resource \emph{source} tasks and then adapts these representations to the \emph{target} task~\citep{Ruder:PhD:2019,Gururangan:Marasovic:ACL:2020}. Data augmentation expands the training set by applying transformations to training instances without changing their original labels~\citep{Shorten:Khoshgoftaar:BigData:2019}. 

Recently, there has been an increased interest on applying data augmentation techniques on sentence-level NLP tasks, such as text classification~\citep{Wei:Zou:EMNLP:2019,Xie:Dai:arXiv:2019}, natural language inference~\citep{Min:McCoy:ACL:2020}, and machine translation~\citep{Wang:Pham:EMNLP:2018}. Augmentation methods explored for these tasks include creating augmented instances by manipulating a few words in the original instance, such as word replacement~\citep{Zhang:Zhao:NIPS:2015,Wang:Yang:EMNLP:2015,Cai:Chen:ACL:2020}, random deletion~\citep{Wei:Zou:EMNLP:2019}, and word position swap~\citep{Sahin:Steedman:EMNLP:2018,Min:McCoy:ACL:2020}; or creating entirely artificial instances via generative models, such as variational autoencoders~\citep{Yoo:Shin:AAAI:2019,Mesbah:Yang:EMNLP:2019} and back-translation models~\citep{Yu:Dohan:ICLR:2018,Iyyer:Wieting:NAACL:2018}. 

Different from these sentence-level NLP tasks, NER is usually regarded as a token-level NLP task. That is, for each token in the sentence, an NER model predicts a label indicating whether the token belongs to an entity mention and which entity category the mention belongs to. Therefore, applying transformations to individual tokens may also change their labels. Due to such a difficulty, data augmentation for NER is relatively less studied. In this chapter, we describe our efforts to fill this gap by exploring data augmentation techniques for NER, solved as a sequence tagging problem. 


\section{Proposed Data Augmentation Methods}
\label{section-data-augmentation-details}
We surveyed the existing data augmentation techniques for sentence-level NLP tasks in Section~\ref{section-low-resource}. Inspired by these efforts, we design several easy to use data augmentation methods for NER. Note that our proposed methods do not rely on any external trained models, such as machine translation models~\citep{Yu:Dohan:ICLR:2018,Iyyer:Wieting:NAACL:2018} or syntactic parsing models~\citep{Sahin:Steedman:EMNLP:2018}, which are by themselves difficult to train in low-resource domain specific scenarios. 

Given an original training instance, consisting of a sequence of tokens and the corresponding sequence of labels, we use the following transformations to create augmented instances. 

\begin{table}[tb] 
	\setlength{\tabcolsep}{2pt}
	\centering
	\begin{tabular}{c|ccc cccc}
		\toprule
		\bf Method & \multicolumn{6}{c}{\bf Instance} \\
		\midrule
		\multirow{5}{*}{None} & She & did & not & complain & of & headache & or \\
		& O & O & O & O & O & B-problem & O \\
		\\
		& any & other & neurological & symptoms & . \\
		& B-problem & I-problem & I-problem & I-problem & O \\
		\midrule
		\multirow{5}{*}{LwTR} & \textit{L.} & \textit{One} & not & complain & of & headache & \textit{he} \\
		& O & O & O & O & O & B-problem & O \\
		\\
		& any & \textit{interatrial} & neurological & \textit{current} & . \\
		& B-problem & I-problem & I-problem & I-problem & O \\
		\midrule
		\multirow{5}{*}{SR} & She & did & \textit{non} & complain & of & headache & or \\
		& O & O & O & O & O & B-problem & O \\
		\\
		& \textit{whatsoever} & \textit{former} & neurologic & symptom & . \\
		& B-problem & I-problem & I-problem & I-problem & O \\
		\midrule
		\multirow{5}{*}{MR} & She & did & not & complain & of & \textit{neuropathic} & \textit{pain} \\
		& O & O & O & O & O & \textit{B-problem} & \textit{I-problem} \\
		\\
		& \textit{syndrome} & or & \textit{acute} & \textit{pulmonary} & \textit{disease} & . \\
		& \textit{I-problem} & O & \textit{B-problem} & \textit{I-problem} & \textit{I-problem} & O \\
		\midrule
		\multirow{5}{*}{SiS} & \textit{not} & \textit{complain} & \textit{She} & \textit{did} & \textit{of} & headache & or \\
		& O & O & O & O & O & B-problem & O \\
		\\
		& \textit{neurological} & \textit{any} & \textit{symptoms} & \textit{other} & . \\
		& B-problem & I-problem & I-problem & I-problem & O \\
		\bottomrule 
	\end{tabular}
	\caption{An original training instance and different types of augmented instances. We highlight changes using \textit{italics}.}
	\label{table-data-augmentation-example}
\end{table}

\paragraph{Label-wise Token Replacement (LwTR)}
For each token which is not a stop word, we use a binomial distribution to randomly decide whether it should be replaced. If yes, we then use a label-wise token distribution, built from the original training set, to randomly select another token with the same label. Thus, we keep the original label sequence unchanged. Taking the instance in Table~\ref{table-data-augmentation-example} as an example, there are five tokens replaced by other tokens which share the same label as the original tokens. 

\paragraph{Synonym Replacement (SR)}
Our second approach is similar to LwTR, except that we replace the token with one of its synonyms retrieved from \WORDNET~\citep{Miller:Beckwith:1990}. Note that the retrieved synonym may consist of more than one token. Its BIO labels can be derived using a straightforward rule: If the replaced token is the first token within a mention (i.e., the corresponding label is `B-Entity'), we assign the same label to the first token of the retrieved multi-word synonym, and `I-Entity' to the other tokens. 

\paragraph{Mention Replacement (MR)}
For each mention in the instance, we use a binomial distribution to randomly decide whether it should be replaced. If yes, we randomly select another mention from the original training set which has the same entity category as the replacement. The corresponding sequence of BIO labels can be changed accordingly. For example, in Table~\ref{table-data-augmentation-example}, the mention `\textit{headache} [B-problem]' is replaced by another problem mention `\textit{neuropathic pain syndrome} [B-problem I-problem I-problem]'. 

\paragraph{Shuffle within Segments (SiS)}
We first split the token sequence into segments of the same entity category. Thus, each segment corresponds to either an entity mention or a sequence of tokens that does not belong to any mention. For example, the original instance in Table~\ref{table-data-augmentation-example} is split into five segments: `\textit{She did not complain of} [Out-of-Mention]', `\textit{headache} [Problem]', `\textit{or} [Out-of-Mention]', `\textit{any other neurological symptoms} [Problem]', `\textit{.} [Out-of-Mention]'. Then for each segment, we use a binomial distribution to randomly decide whether it should be shuffled. If yes, the order of the tokens within the segment is shuffled, while the label order is kept unchanged. 

\paragraph{All}
We also explore the augmentation of the training set using all aforementioned augmentation methods. That is, for each training instance, we create multiple augmented instances, one per transformation. 

\section{Evaluation}
We present an empirical analysis of the data augmentation methods described in Section~\ref{section-data-augmentation-details} on two English datasets from the biomedical domain~\footnote{In~\citep{Dai:Adel:COLING:2020}, we also evaluate these methods on a dataset from the materials science domain.}: \textsc{i2b2-2010}~\citep{Uzuner:South:AMIA:2011} and \NCBIDISEASE~\citep{Dogan:Leaman:JBI:2014}. 

We use a \textsc{BERT-CRF} model~\citep{Beltagy:Lo:EMNLP:2019,Baevski:Edunov:EMNLP:2019} as the backbone model, and we investigate the impact of applying data augmentation on training data of different sizes.

\subsection{Datasets}

\begin{table}[tb] 
	\centering
	\begin{tabular}{r | rrr | rrr }
		\toprule
		& \multicolumn{3}{c|}{\bf \textsc{i2b2-2010}} & \multicolumn{3}{c}{\bf \NCBIDISEASE} \\
		\hline
		& Train & Dev & Test & Train & Dev & Test \\
		\midrule
		\# Sentences & 13,868 & 2,447 & 27,625 & 5,424 & 923 & 940 \\
		\# Tokens & 129,087 & 20,454 & 267,249 & 135,701 & 23,969 & 24,497 \\
		\# Mentions & 14,376 & 2,143 & 31,161 & 5,134 & 787 & 960 \\
		\bottomrule
	\end{tabular}
	\caption{The descriptive statistics of the two English datasets from the biomedical domain: \textsc{i2b2-2010}~\citep{Uzuner:South:AMIA:2011} and \NCBIDISEASE~\citep{Dogan:Leaman:JBI:2014}.}
	\label{table-data-augmentation-data-statistics}
\end{table}

\textsc{i2b2-2010} focuses on the identification of three entity types of problem, treatment and test from patient reports. We use the train-test split from its corresponding shared task and randomly select $15$\% of sentences from the training set as the development set. \NCBIDISEASE contains scholarly articles annotated with disease names. We use the train-dev-test split provided by the authors. Descriptive statistics of these two datasets is listed in Table~\ref{table-data-augmentation-data-statistics}. 

To simulate a low-resource setting, we select the first $50$, $150$, $500$ sentences which contain at least one mention from the complete training set to create the corresponding small, medium, and large subsets (denoted as S, M, L in Table~\ref{table-data-augmentation-main-results}, whereas the complete training set is denoted as F). Note that we apply data augmentation only on the training set, without changing the development and test sets. 

\subsection{Backbone model}
We regard the NER task as a token-level sequence tagging problem, where each token in the sentence is assigned a tag. The tag can be used to infer whether the token is the first token within a mention, inside a mention or does not belong to any mention. 

\begin{figure}[tb] 
	\centering
	\includegraphics[width=0.95\textwidth]{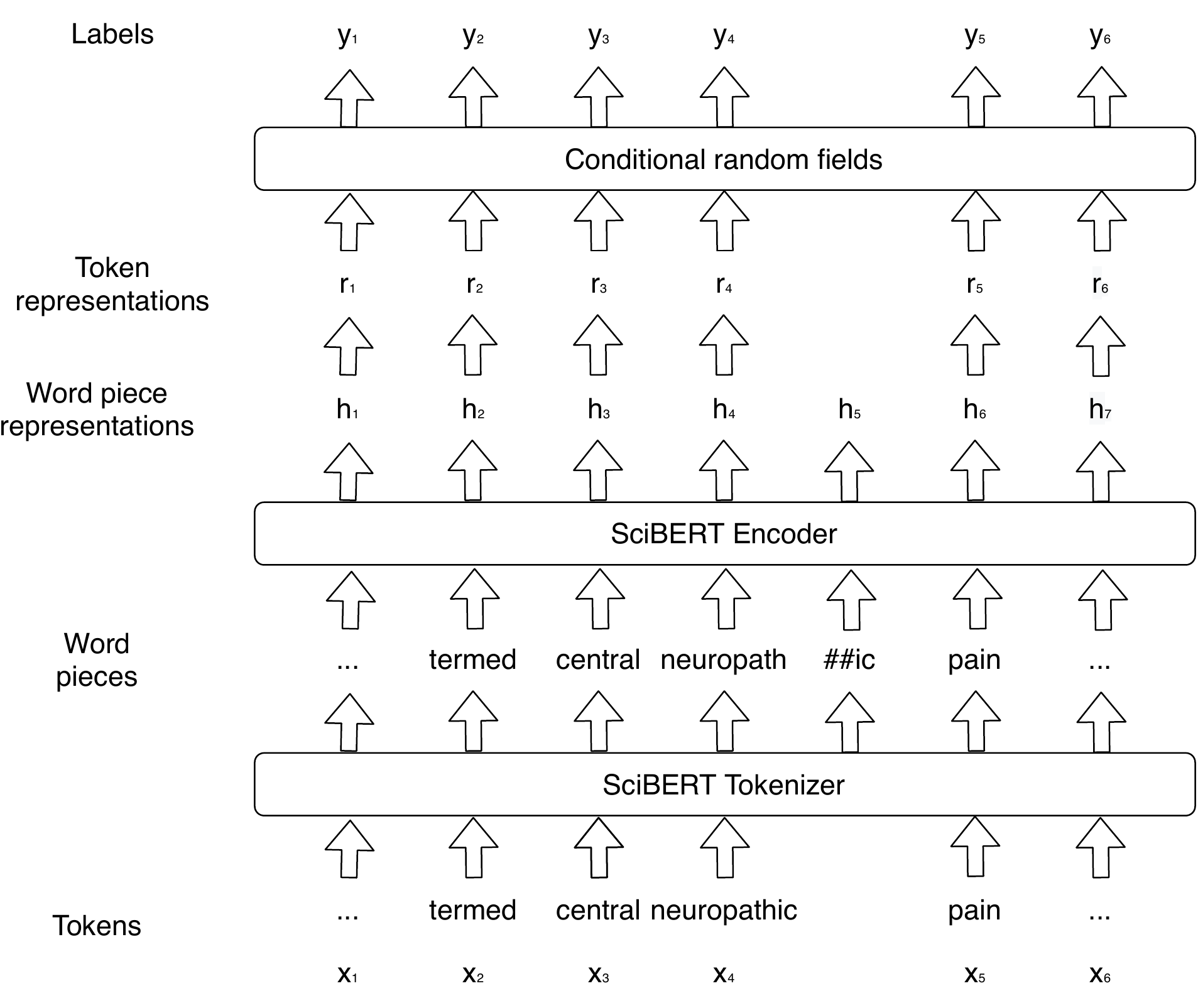}
	\caption{High level overview of the \textsc{BERT-CRF} model.}
	\label{figure-data-augmentation-backbone}
\end{figure}

The backbone model, illustrated in Figure~\ref{figure-data-augmentation-backbone}, is a \textsc{BERT-CRF} model~\citep{Beltagy:Lo:EMNLP:2019,Baevski:Edunov:EMNLP:2019}. It takes advantage of large scale pre-trained language models---using BERT-based encoder to create contextual representations for each token, and a probabilistic graphical model---using conditional random fields~\citep{Sutton:McCallum:2007} to capture dependencies between neighbouring tags. 

\paragraph{BERT-based encoder}
Given a sentence, the tokenizer, coupled with the pre-trained BERT-based model, first converts each token in the sentence into word pieces. That is, if the original token does not exist in the vocabulary, it will be segmented into several pieces from the vocabulary~\citep{Sennrich:Haddow:ACL:2015}. Then the word pieces are mapped to dense vectors---token embeddings---via a lookup table. Finally, the sum of token embeddings and positional embeddings, which indicate the position of each token in the sequence, are fed into a stack of multi-head self-attention and fully-connected feed-forward layers~\citep{Vaswani:Shazeer:NIPS:2017}. Following the study in~\citep{Devlin:Chang:NAACL:2019}, we use the final outputs corresponding to the first word piece within each token as the token representation. 

Recent studies on domain-specific BERT models show that effectiveness on downstream tasks can be improved when the BERT models are further pre-trained on in-domain data~\citep{Gururangan:Marasovic:ACL:2020}. We thus choose SciBERT~\citep{Beltagy:Lo:EMNLP:2019}, which is pre-trained on full text of scholarly articles about biology and computer science, and fine-tune it on the target NER task. In our preliminary experiments, we observe that SciBERT achieves significant better results than vanilla BERT~\citep{Devlin:Chang:NAACL:2019} and slightly better results than BioBERT~\citep{Lee:Yoon:Bioinformatics:2020}. 

\paragraph{Conditional random fields (CRF)}
Instead of assigning a tag to each token independently, we model them jointly using a conditional random fields. That is, given a sequence of token representations $\mathbf{R}=\left(\mathbf{r}_{1}, \mathbf{r}_{2}, \ldots, \mathbf{r}_{n}\right)$, we aim to predict a sequence of tags $\mathbf{y}=\left(y_{1}, y_{2}, \ldots, y_{n}\right)$ which has the maximum probability over all possible tag sequences. This conditional probability can be calculated using: 
\[
p(\mathbf{y} \mid \mathbf{R})=\frac{e^{s(\mathbf{R}, \mathbf{y})}}{\sum_{\widetilde{\mathbf{y}} \in \mathbf{Y}_{\mathbf{R}}} e^{s(\mathbf{R}, \widetilde{\mathbf{y}})}}
\]
and 
\[
s(\mathbf{R}, \mathbf{y})=\sum_{i=0}^{n} A_{y_{i}, y_{i+1}}+\sum_{i=1}^{n} P_{i, y_{i}},
\]
\noindent
where $A_{i,j}$ is the compatibility score of a transition from the tag i to tag j, and $P_{i, j}$ is the score of the tag $j$ given token representation $r_i$.  

The parameters, of both the SciBERT encoder and CRFs, are trained jointly to maximise the conditional probability of gold tag sequence given the training sentences.  

\subsection{Experimental results}
The evaluation results on the effectiveness of data augmentation methods are shown in Table~\ref{table-data-augmentation-main-results}. We use the Micro-average string match $F_1$ score to evaluate the effectiveness of the models. The model which is most effective on the development set, measured using the $F_1$ score, is finally evaluated on the test set. All experiments are repeated five times with different random seeds. Mean values and standard deviations are reported. The $\Delta$ row shows the averaged improvement due to data augmentation, comparing against the baseline without using data augmentation. In general, we find that all data augmentation methods improve over the baseline, and synonym replacement outperforms other augmentation on average. 

\begin{table}[tb] 
		\centering
		\begin{tabular}{r | c | c | cccc | c}
			\toprule
			Corpus & Size & Baseline & LwTR & SR & MR & Sis & All \\
			\hline
\multirow{4}{*}{i2b2-2010} & S & 34.6 \scriptsize{$\pm$ 1.4} & \bf \underline{39.9 \scriptsize{$\pm$ 0.7}} & \bf \underline{43.8 \scriptsize{$\pm$ 2.0}} & \bf \underline{39.6 \scriptsize{$\pm$ 1.3}} & \bf \underline{39.2 \scriptsize{$\pm$ 1.8}} & \bf \underline{42.6 \scriptsize{$\pm$ 1.4}} \\
 & M & 62.9 \scriptsize{$\pm$ 1.0} & \bf 64.0 \scriptsize{$\pm$ 1.3} & \bf \underline{64.5 \scriptsize{$\pm$ 0.5}} & \bf 63.5 \scriptsize{$\pm$ 0.8} & \bf 63.1 \scriptsize{$\pm$ 1.4} & \bf 63.9 \scriptsize{$\pm$ 1.8} \\
 & L & 69.6 \scriptsize{$\pm$ 0.3} & \bf 70.5 \scriptsize{$\pm$ 1.6} & \bf 70.7 \scriptsize{$\pm$ 1.1} & \bf 70.6 \scriptsize{$\pm$ 0.9} & \bf 71.0 \scriptsize{$\pm$ 1.2} & \bf 70.5 \scriptsize{$\pm$ 1.0} \\
 & F & 87.6 \scriptsize{$\pm$ 0.3} & 87.3 \scriptsize{$\pm$ 0.2} & \bf 87.7 \scriptsize{$\pm$ 0.2} & 87.6 \scriptsize{$\pm$ 0.1} & 87.1 \scriptsize{$\pm$ 0.2} & 86.8 \scriptsize{$\pm$ 0.3} \\
\midrule
\multirow{4}{*}{NCBI-disease} & S & 59.9 \scriptsize{$\pm$ 2.6} & \bf 62.9 \scriptsize{$\pm$ 1.4} & \bf 63.6 \scriptsize{$\pm$ 2.3} & \bf \underline{65.1 \scriptsize{$\pm$ 1.3}} & \bf 63.0 \scriptsize{$\pm$ 1.1} & \bf \underline{63.8 \scriptsize{$\pm$ 1.3}} \\
 & M & 71.6 \scriptsize{$\pm$ 1.4} & \bf 73.2 \scriptsize{$\pm$ 1.5} & \bf \underline{74.7 \scriptsize{$\pm$ 1.0}} & \bf 73.6 \scriptsize{$\pm$ 1.1} & \bf 73.4 \scriptsize{$\pm$ 1.1} & \bf 73.3 \scriptsize{$\pm$ 1.2} \\
 & L & 81.0 \scriptsize{$\pm$ 0.4} & 80.4 \scriptsize{$\pm$ 1.1} & \bf 82.2 \scriptsize{$\pm$ 1.0} & 80.5 \scriptsize{$\pm$ 1.3} & 80.6 \scriptsize{$\pm$ 1.0} & \bf 81.3 \scriptsize{$\pm$ 0.5} \\
 & F & 87.6 \scriptsize{$\pm$ 0.3} & 85.7 \scriptsize{$\pm$ 1.1} & \bf 87.9 \scriptsize{$\pm$ 0.7} & \bf 88.1 \scriptsize{$\pm$ 0.7} & 87.4 \scriptsize{$\pm$ 0.4} & 86.0 \scriptsize{$\pm$ 1.0} \\
\hline
\multicolumn{3}{c|}{$\Delta$} & 1.1 & 2.5 & 1.7 & 1.2 & 1.6 \\
			\bottomrule
		\end{tabular}
		\caption{Evaluation results in terms of span-level $F_1$ score. \textbf{S}mall set contains 50 training instances; \textbf{M}edium contains 150 instances; \textbf{L}arge contains 500 instances; \textbf{F}ull uses the complete training set. Results that are better than the baseline model without using data augmentation are highlighted in bold. \underline{underline}: the result is significantly better than the baseline model without data augmentation (paired student's t-test, p: $0.05$)}
		\label{table-data-augmentation-main-results}
\end{table}

Another observation is that the data augmentation methods are more effective when the original training sets are small. For example, all data augmentation methods achieve improvements when the training set contains only $50$ training instances. In contrast, when the complete training sets are used, only synonym replacement and mention replacement achieve improvements. This has also been observed in previous work on applying data augmentation on other NLP tasks~\citep{Fadaee:Bisazza:ACL:2017,Sahin:Steedman:EMNLP:2018,Xia:Kong:ACL:2019}. 




\section{Analysis}
After demonstrating the effectiveness of proposed data augmentation methods, we present an analysis of the best two performing transformations: synonym replacement and mention replacement. We aim to provide practical suggestions on hyperparameter settings as well as understandings about how they improve the performance. 

\subsection{The impact of hyperparameters}
For each augmentation method, we tune the number of augmented instances per original training instance from a list of numbers: \{1, 3, 6, 10\}. We also tune the $p$ value of the binomial distribution which is used to decide whether a token or a mention should be replaced. It is searched over the range from 0.1 to 0.7, with an incremental step of 0.2. We perform grid search to find the best combination of these two hyperparameters on the development set. 

The main question we aim to answer is how much augmentation is enough? More specifically, how the number of augmented instances per original training instance affects performance, and how the ratio a token, or a mention, is replaced affects performance? 

\begin{figure}[tb] 
	\begin{subfigure}{.48\textwidth}
		\centering
		\includegraphics[width=.98\linewidth]{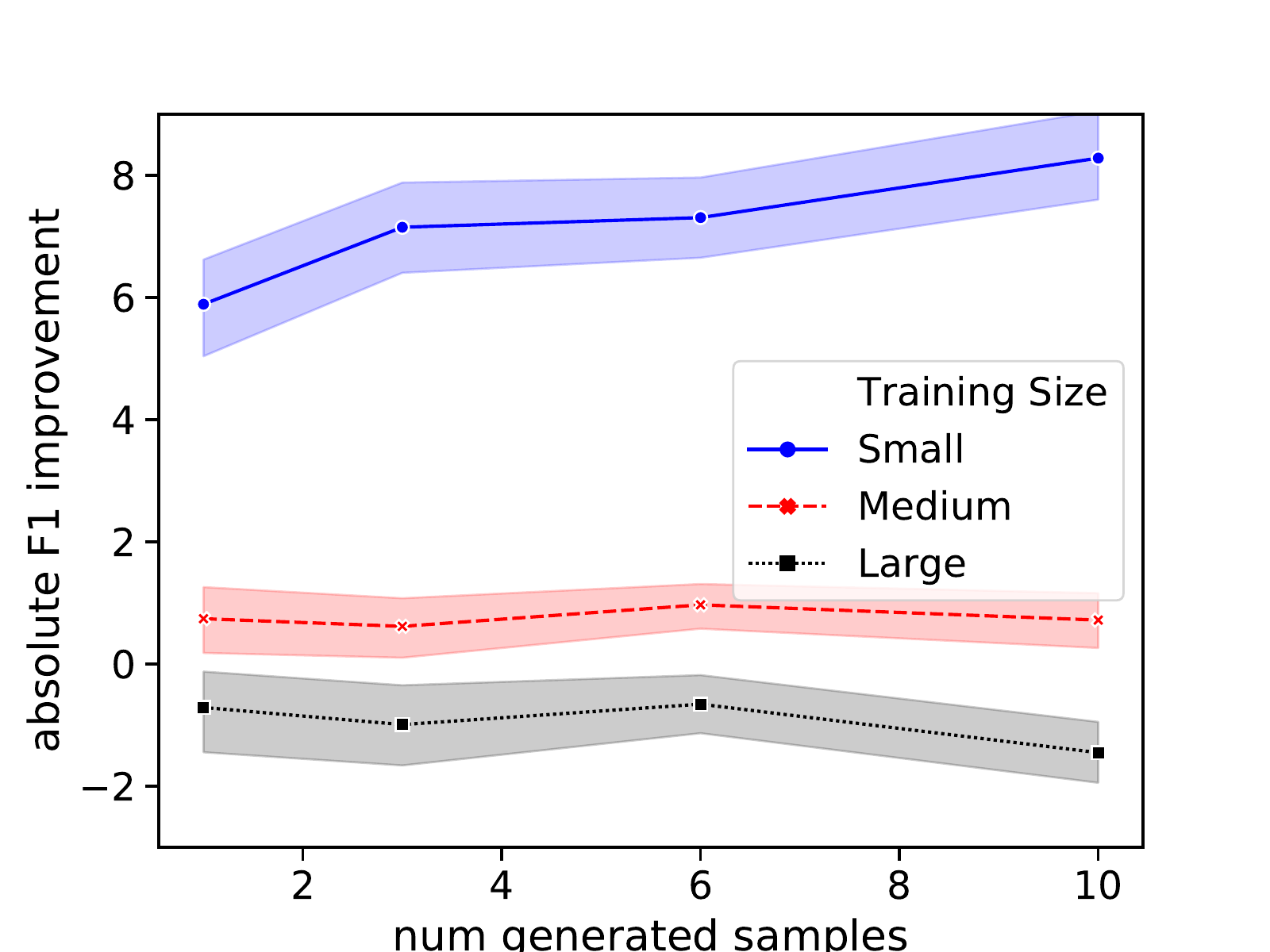}
		\caption{\textsc{i2b2-2010}, SR}
	\end{subfigure}%
	\begin{subfigure}{.48\textwidth}
		\centering
		\includegraphics[width=.98\linewidth]{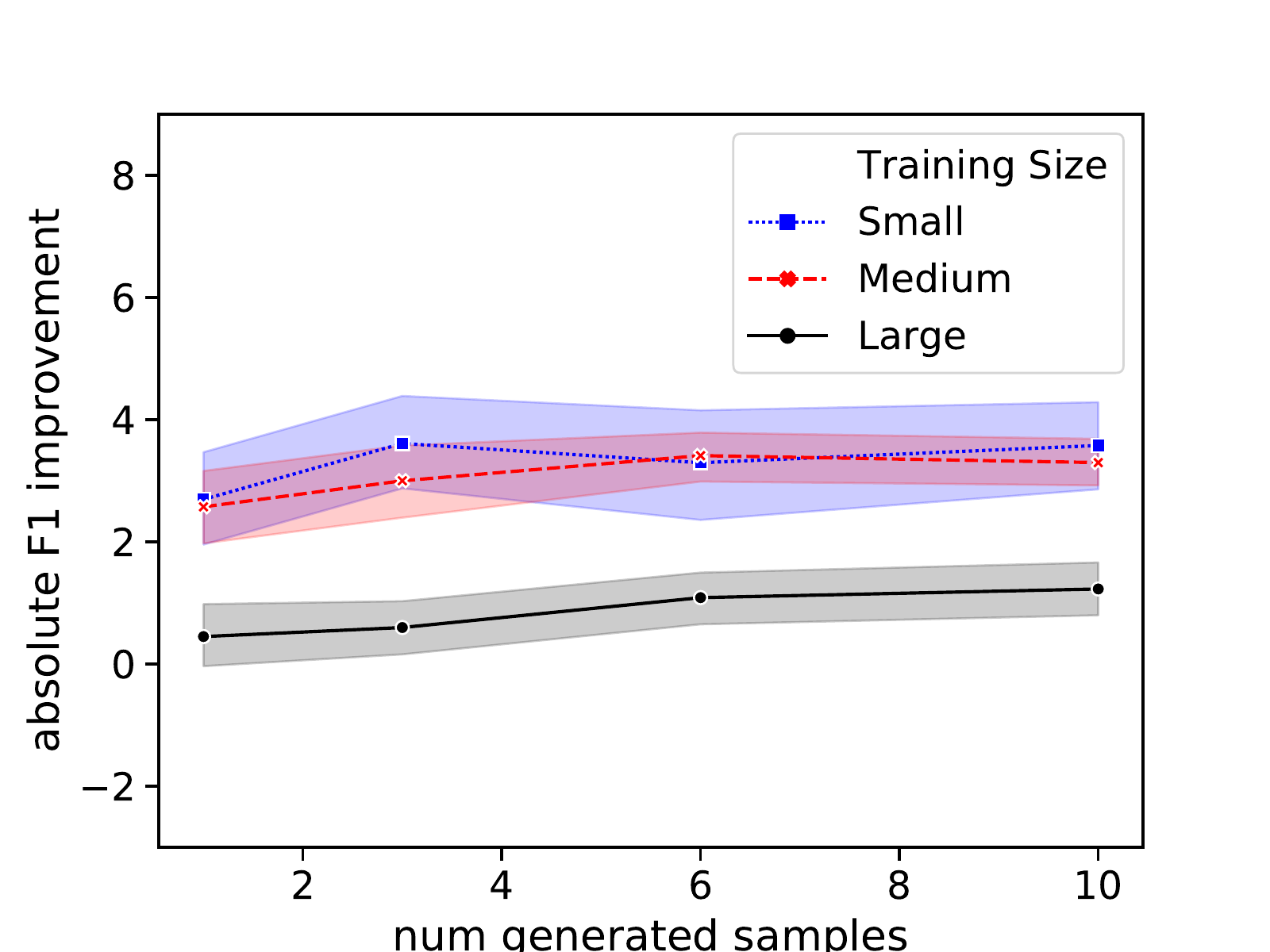}
		\caption{\NCBIDISEASE, SR}
	\end{subfigure}%
	
	\begin{subfigure}{.48\textwidth}
		\centering
		\includegraphics[width=.98\linewidth]{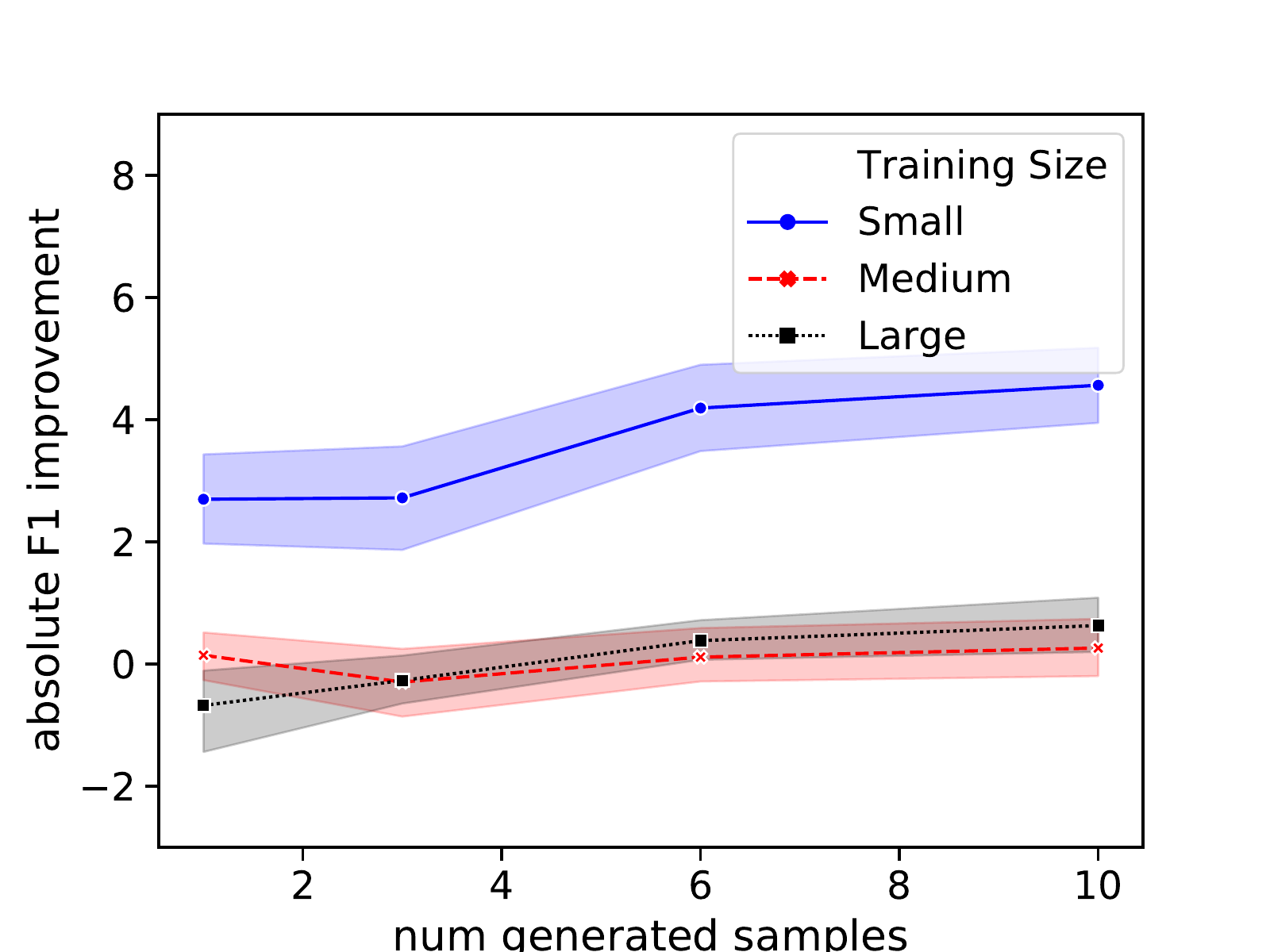}
		\caption{\textsc{i2b2-2010}, MR}
	\end{subfigure}%
	\begin{subfigure}{.48\textwidth}
		\centering
		\includegraphics[width=.98\linewidth]{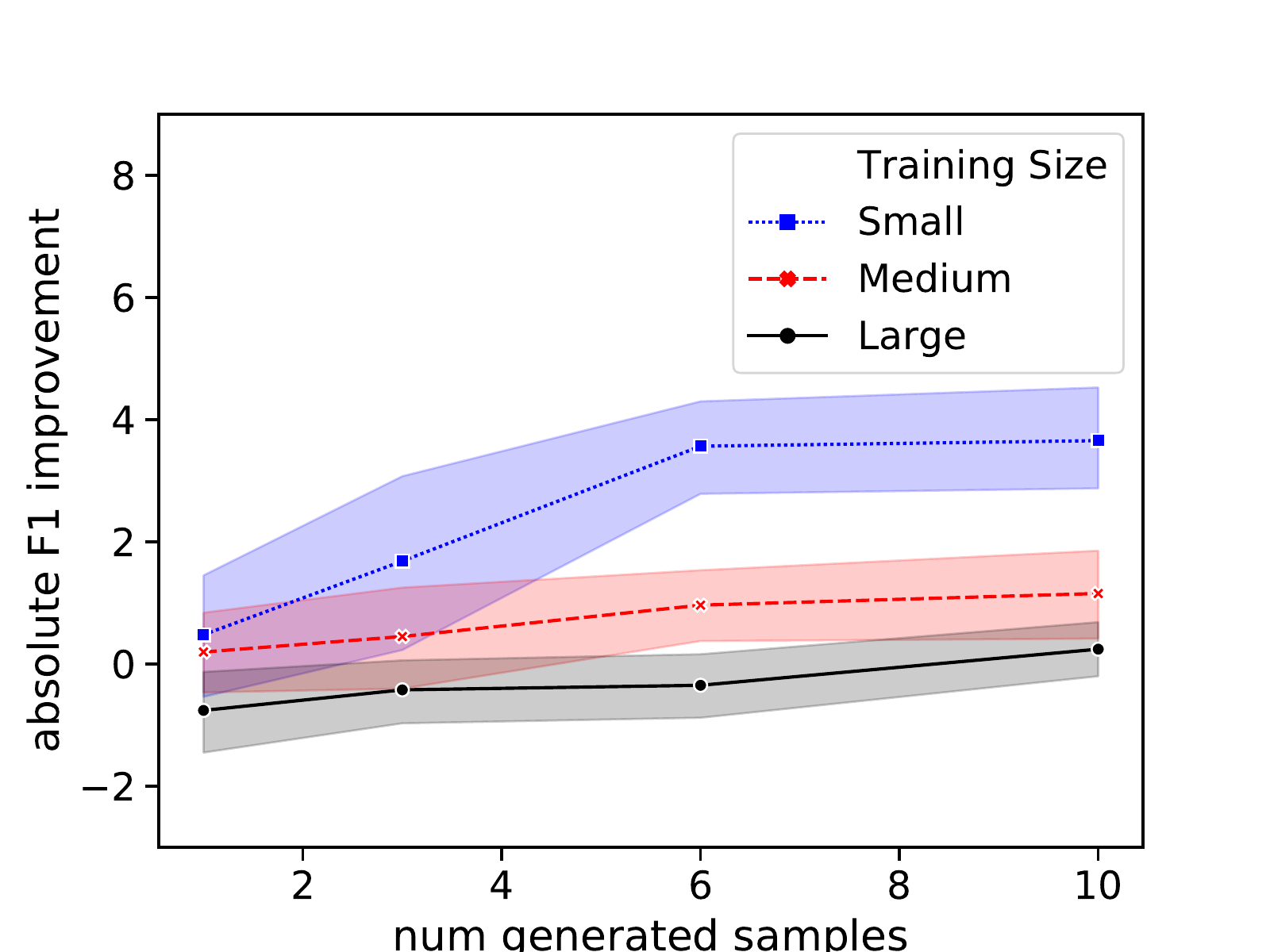}
		\caption{\NCBIDISEASE, MR}
	\end{subfigure}
	\caption{Impact of the number of augmented instances per original training instance on the effectiveness of data augmentation. SR: synonym replacement. MR: mention replacement.~\label{figure-data-augmentation-number-of-augmented-examples}}
\end{figure}

Figure~\ref{figure-data-augmentation-number-of-augmented-examples} shows the impact of the number of augmented instances per original training instance on the performance gain of synonym replacement and mention replacement. We use the improvement of absolute $F_1$ score over the baseline without using data augmentation as the performance gain. 

In general, we find that \emph{larger number of augmented instances can bring larger performance gain, especially when the training sets are small (i.e., $50$ training instances)}. However, the performance gain becomes relatively small when the number of augmented instances per original training instance is greater than $6$. The second observation is that when the training sets are medium (i.e., $150$ training instances) or large (i.e., $500$ training instances), the benefits of more augmented instances become small. On \textsc{i2b2-2010}, creating more augmented instances using synonym replacement on large training set even decreases the performance. 

\begin{figure}[tb] 
	\begin{subfigure}{.48\textwidth}
		\centering
		\includegraphics[width=.98\linewidth]{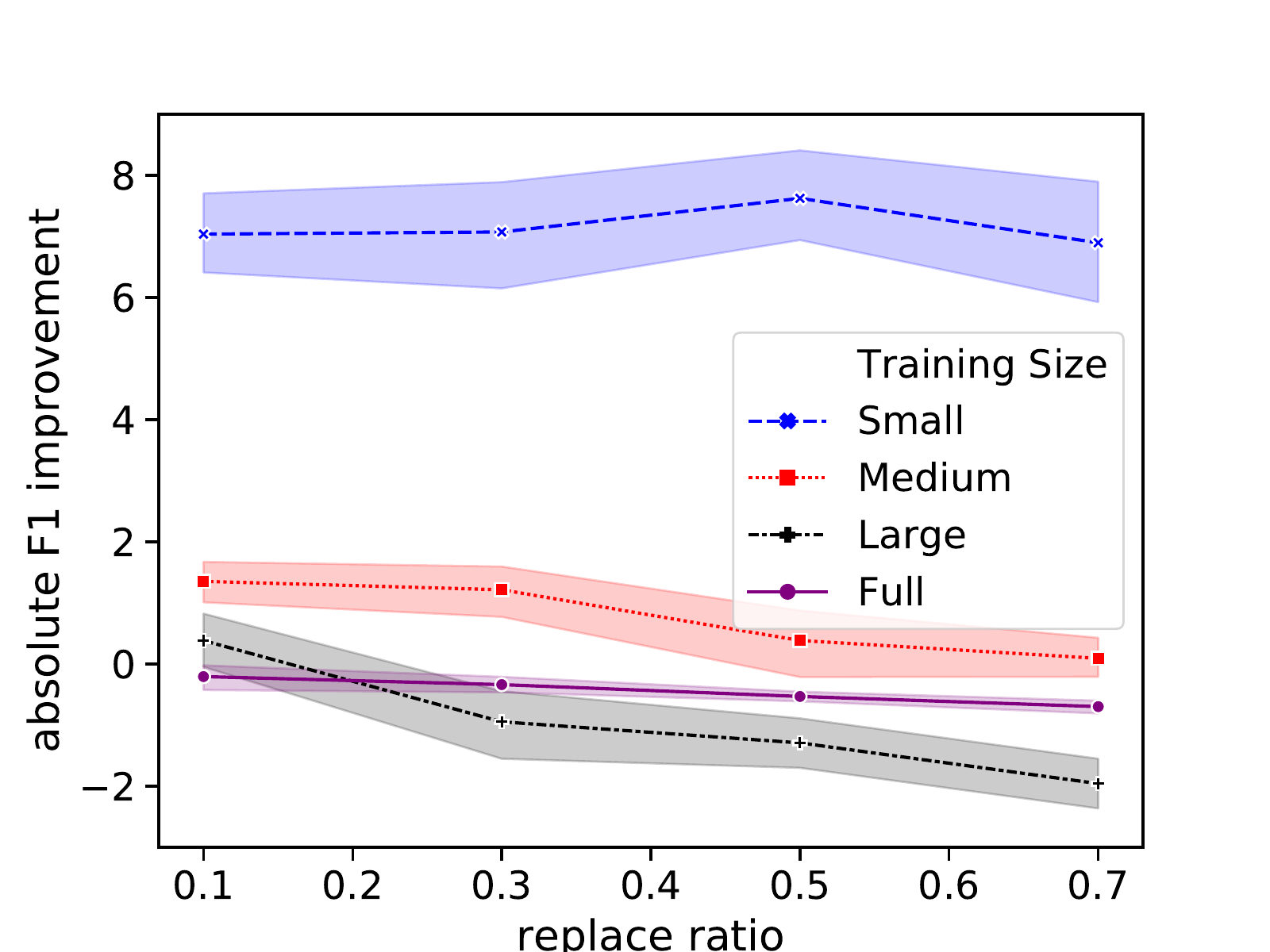}
		\caption{\textsc{i2b2-2010}, SR~\label{figure-data-augmentation-replace_ratio-sr-i2b2}}
	\end{subfigure}%
	\begin{subfigure}{.48\textwidth}
		\centering
		\includegraphics[width=.98\linewidth]{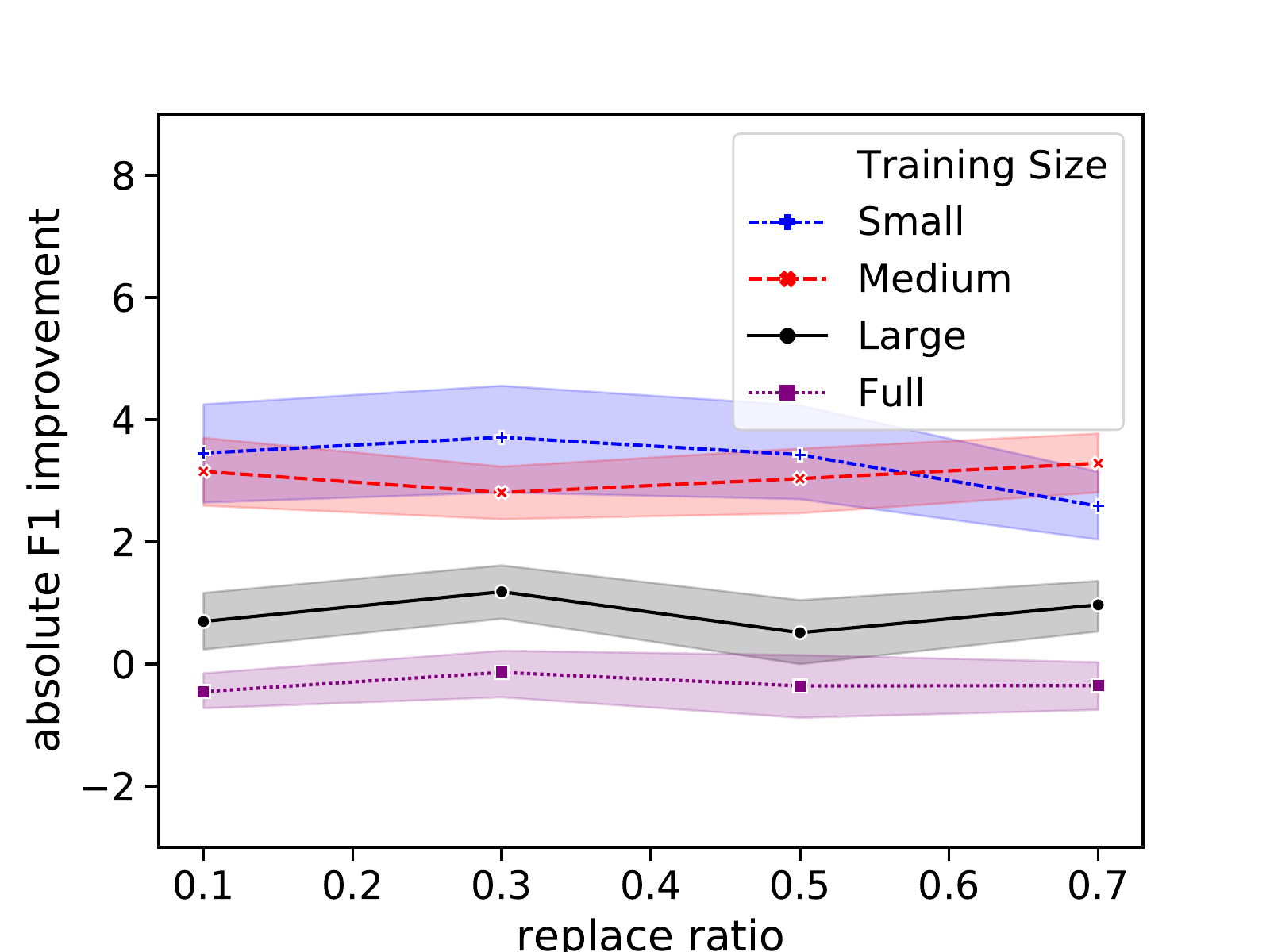}
		\caption{\NCBIDISEASE, SR~\label{figure-data-augmentation-replace_ratio-sr-ncbi}}
	\end{subfigure}%
	
	\begin{subfigure}{.48\textwidth}
		\centering
		\includegraphics[width=.98\linewidth]{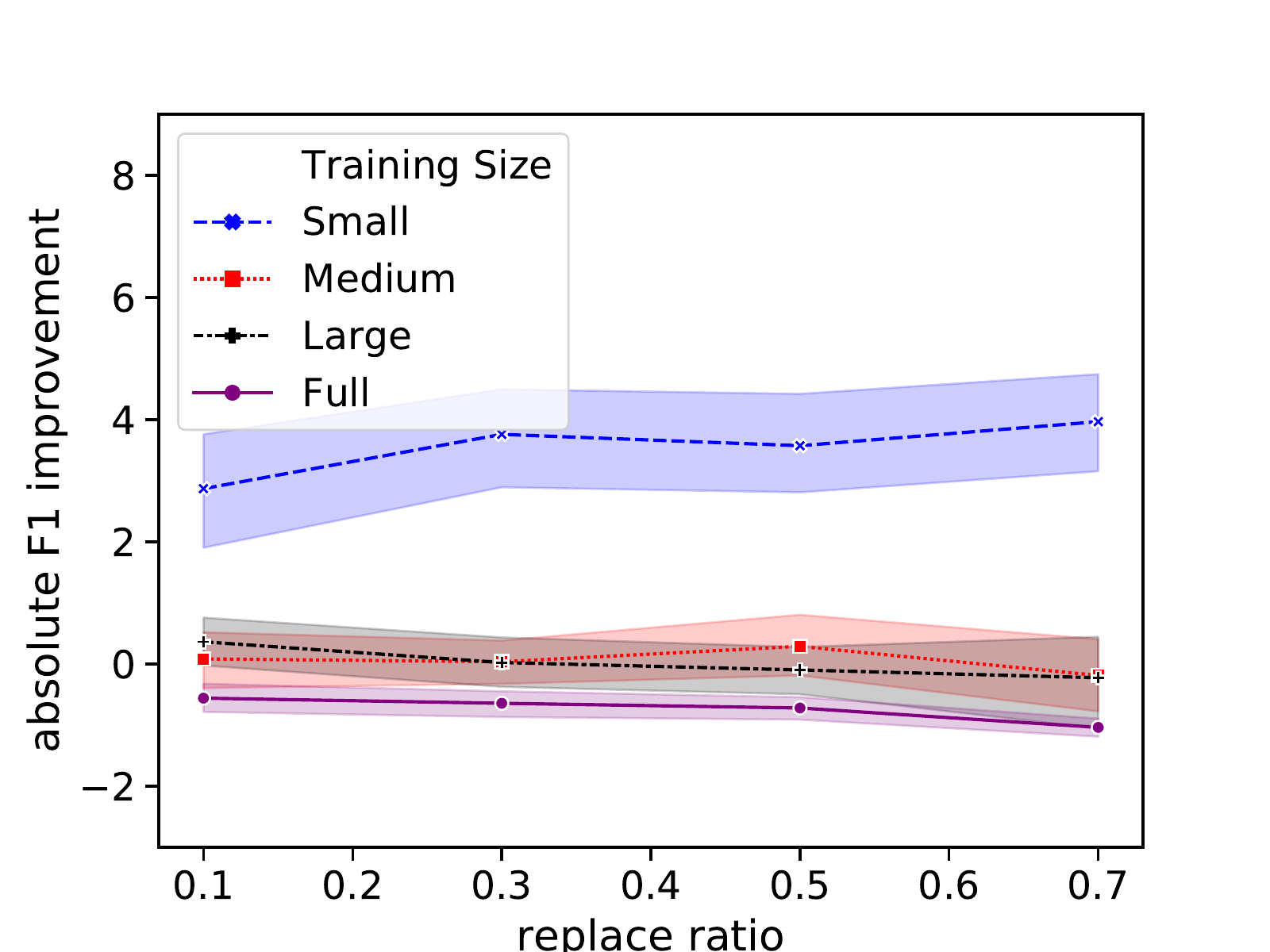}
		\caption{\textsc{i2b2-2010}, MR~\label{figure-data-augmentation-replace_ratio-mr-i2b2}}
	\end{subfigure}%
	\begin{subfigure}{.48\textwidth}
		\centering
		\includegraphics[width=.98\linewidth]{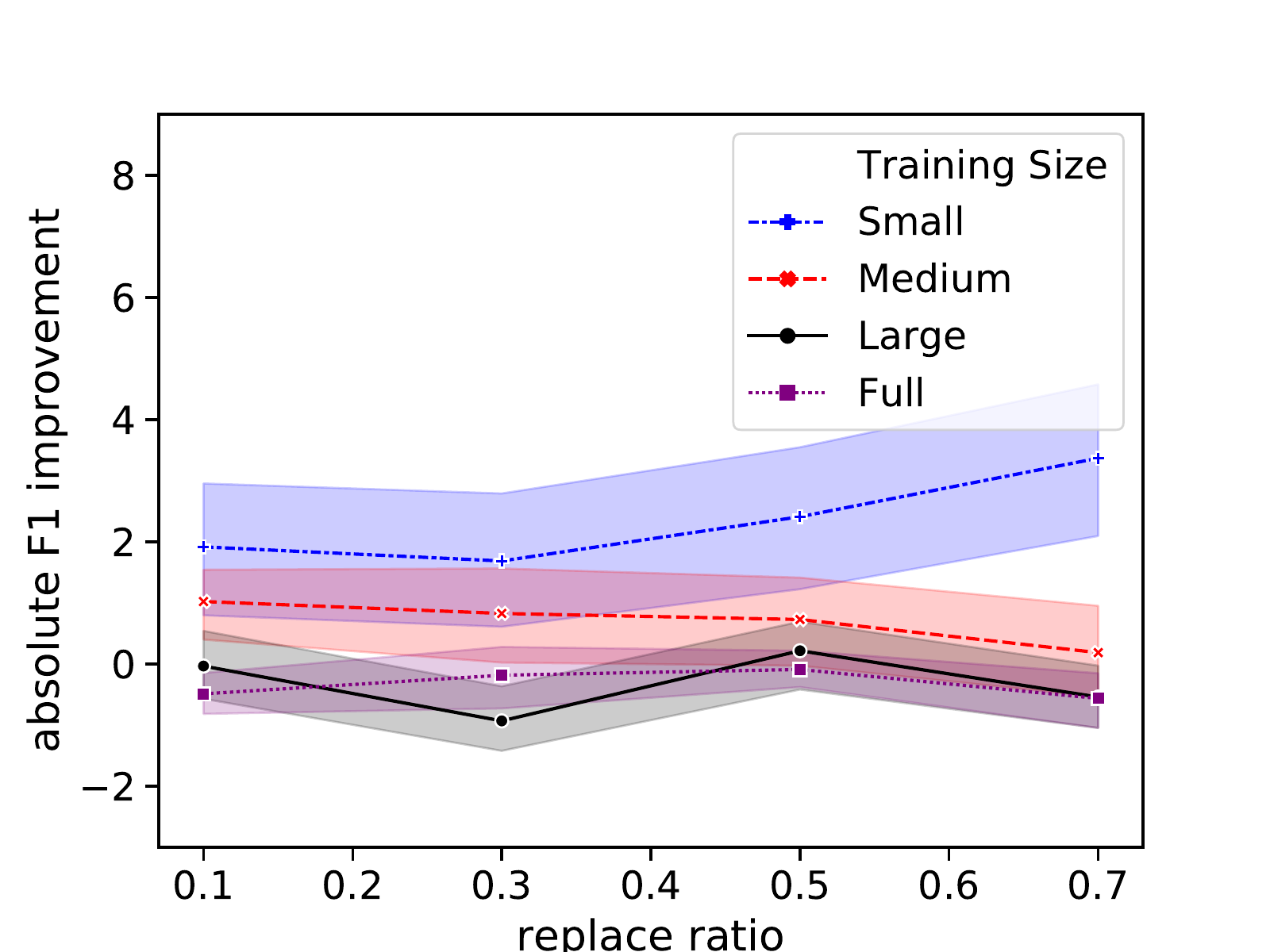}
		\caption{\NCBIDISEASE, MR~\label{figure-data-augmentation-replace_ratio-mr-ncbi}}
	\end{subfigure}
	\caption{The impact of the ratio a token or a mention is replaced on the effectiveness of data augmentation. SR: synonym replacement. MR: mention replacement.~\label{figure-data-augmentation-replace_ratio}}
\end{figure}

Figure~\ref{figure-data-augmentation-replace_ratio} A and B shows the impact of the ratio a token is replaced with one of its synonyms on the performance gain. We note a moderate ratio (e.g., $0.3$ or $0.6$) performs well across different setups. If the ratio is too small, the augmented instances may be very similar to the original one. Training on these augmented instances may have a similar effect as training on the original training instances for more epochs. In contrast, a large ratio is more likely to create syntactically invalid instance. These syntactically invalid instances are noisy, and training on such a combination of small amount of clean data and large amount of noisy data may underperform training on clean data only (more discussions in Section~\ref{section-low-resource-distant-supervision}). 

The pattern with mention replacement is different from the one with synonyms replacement (Figure~\ref{figure-data-augmentation-replace_ratio} C and D). When the training sets are small, increasing the ratio a mention is replaced with another mention of the same entity category always enlarges the performance gain. We find that most of these entity mentions have similar part of speech patterns. That is, most of them are either nouns or noun phrases. Because of this feature, the risk of creating syntactically invalid instance is much lower using mention replacement than using synonyms replacement. It can be the reason why \emph{a moderate ratio for synonyms replacement is best, whereas mention replacement can benefit from a large replace ratio}. 

\subsection{A closer look at errors}
The next question we aim to answer is how data augmentation improves the performance. Put another way, are data augmentation methods guaranteed to fix some particular errors predicted by the baseline model without using data augmentation, and if yes, which types of errors are more likely to get rectified. 

\begin{table}[tb]
    \centering
    \begin{tabular}{c|c|c|c|c}
    \toprule
    \multirow{2}{*}{} & \multicolumn{2}{c|}{\textsc{i2b2-2010}} & \multicolumn{2}{c}{\NCBIDISEASE} \\
     \cmidrule{2-3}\cmidrule{4-5}
     & FPs & FNs & FPs & FNs \\
    \midrule
Baseline $\cap$ SR $\cap$ MR & 12574 (26.9) & \bf 13500 (68.8) & 170 (20.5) & \bf 197 (45.8) \\
Baseline $\cap$ SR $\cap$ $\neg$ MR & \phantom{0}2424 (\phantom{0}5.2) & \phantom{0}2001 (10.2) & \phantom{0}53 (\phantom{0}6.4) & \phantom{0}50 (11.6) \\
Baseline $\cap$ $\neg$ SR $\cap$ MR & \phantom{0}4261 (\phantom{0}9.1) & \phantom{00}658 (\phantom{0}3.4) & 105 (12.7) & \phantom{0}18 (\phantom{0}4.2) \\
Baseline $\cap$ $\neg$ SR $\cap$ $\neg$ MR & \phantom{0}6674 (14.3) & \phantom{0}1613 (\phantom{0}8.2) & 122 (14.7) & \phantom{0}58 (13.5) \\
$\neg$ Baseline $\cap$ SR $\cap$ MR & \phantom{0}2949 (\phantom{0}6.3) & \phantom{00}335 (\phantom{0}1.7) & \phantom{0}72 (\phantom{0}8.7) & \phantom{0}42 (\phantom{0}9.8) \\
$\neg$ Baseline $\cap$ SR $\cap$ $\neg$ MR & \phantom{0}5034 (10.8) & \phantom{0}1107 (\phantom{0}5.6) & \phantom{0}84 (10.1) & \phantom{0}46 (10.7) \\
$\neg$ Baseline $\cap$ $\neg$ SR $\cap$ MR & \bf 12767 (27.3) & \phantom{00}421 (\phantom{0}2.1) & \bf 223 (26.9) & \phantom{0}19 (\phantom{0}4.4) \\
    \bottomrule
    \end{tabular}
    \caption{The comparison of different types of errors---FPs (false positives) and FNs (false negatives)---made by the baseline model without using data augmentation and models using Synonym Replacement (SR) and Mention Replacement (MR) data augmentation methods. $\cap$ indicates the intersection of two sets, and $\neg$ indicates the negative set. For example, the `FPs' column corresponding to the `Baseline $\cap$ SR $\cap$ $\neg$ MR' row shows the number of false positives predicted by both the baseline model and the model using SR data augmentation, but not by the one using MR data augmentation. ~\label{figure-data-augmentation-types-of-errors-fixed}}
\end{table}

To answer this question, we train three models---one baseline model without using data augmentation, two models using synonym replacement and mention replacement, respectively---and then compare the error predictions by these three models. 

From Table~\ref{figure-data-augmentation-types-of-errors-fixed}, we find \emph{data augmentations are more likely to reduce false positives than false negatives}. In other words, if the baseline model fails to recall some entity mentions, the model trained using data augmentation usually fails to recall them as well. However, data augmentation can fix those mistakenly predicted entity mentions. On one hand, we believe this improvement can be linked to the \emph{over-fitting} problem. That is, the model trained without using data augmentation may overfit some patterns observed in the small training set, and data augmentation can relieve this problem, by creating a new combination of mention and context. On the other hand, training model using data augmentation provides very little improvement on fixing those false negatives. Note that data augmentation may also make large amount of new false positives. In other words, there is no guarantee data augmentation can fix some particular errors predicted by models without using data augmentation, since they may provide a mechanism to prevent the training from over-fitting, but not help the learning algorithm to discover new regularities. 

\section{Summary}
We design several easy to use data augmentation methods for NER: label-wise token replacement, synonym replacement, mention replacement, and shuffle within segments. Through experiments on two datasets from the biomedical domain, we find that all proposed data augmentation methods can improve over the strong baseline, where large scale pre-trained models are used, and synonym replacement outperforms other augmentation on average. 

%% file: src/ch5-select-pretraining-data.tex
Pre-training language representation models on unlabelled data and then adapting them to downstream supervised tasks has become a standard practice in NLP. However, the selection of pre-training data usually resorts to intuition, which varies across NLP practitioners. We make use of similarity measures to nominate in-domain pre-training data. Experimental results suggest that simple similarity measures are good predictors of the usefulness of pre-trained language representation models on downstream NER tasks.

\section{Overview}
Sequential transfer learning---which pre-trains a model from a \emph{source} task and then adapts it to a different \emph{target} task---has demonstrated its effectiveness on a range of NLP tasks~\citep{Pan:Yang:TKDE:2009,Weiss:Khoshgoftaar:BigData:2016,Ruder:PhD:2019}. There are two stages in this procedure: pre-training, and adaptation. Researchers who work on low-resource NLP usually spend a considerable amount of efforts and resources on choosing useful external data sources and investigating how to transfer knowledge to their target tasks. 

\citet{Mikolov:Sutskever:NIPS:2013,Peters:Neumann:NAACL:2018,Devlin:Chang:NAACL:2019} make the most of limited labelled data by incorporating language representation models which are pre-trained on a large amount of unlabelled data. This benefits a range of NLP tasks where appropriate unlabelled data is available, and has become a standard practice in NLP. 

However, there is still a lack of systematic study on how to select appropriate data to pre-train language representation models. We observe two heuristic strategies in the literature: 
\begin{enumerate}
    \item collecting as large as possible generic data, such as news~\citep{Mikolov:Sutskever:NIPS:2013,Peters:Neumann:NAACL:2018,Liu:Ott:arXiv:2019}, web crawl~\citep{Pennington:Socher:EMNLP:2014,Mikolov:Grave:LREC:2018}, and Wikipedia~\citep{Bojanowski:Grave:TACL:2017,Devlin:Chang:NAACL:2019}; and,
    \item selecting moderate size data focusing on a specific domain. The resulting pre-trained models are called \emph{domain-specific models}~\citep{Chiu:Crichton:BioNLP:2016,Karimi:Dai:BioNLP:2017,Chronopoulou:Baziotis:NAACL:2019,Nguyen:Vu:EMNLP:2020,Lee:Yoon:Bioinformatics:2020}.
\end{enumerate}
The advantage of the first strategy is that the pre-trained generic models can be re-used in various domains, however, the corresponding training cost is high and unbearable to many academic labs. For example, \citet{Liu:Ott:arXiv:2019} pre-train the RoBERTa model using 1024 V100 GPUs, which are only accessible by large companies. Therefore, we focus on studying the second strategy, and we aim to pre-train domain-specific models, optimising the performance on downstream biomedical NER datasets. 

Studies on domain-specific language representation models empirically show that target task performance can be improved, when in-domain data is used for pre-training~\citep{Alsentzer:Murphy:ClinicalNLP:2019,Lee:Yoon:Bioinformatics:2020,Beltagy:Lo:EMNLP:2019}. These publicly available domain-specific models are valuable to the NLP community. However, the selection of in-domain data usually resorts to intuition, which varies across NLP practitioners (Section~\ref{section-select-pretraining-data-human-intuition}). According to~\citet{Halliday:Hasan:1989}, the context specific usage of language is affected by three factors: \emph{field} (the subject matter being discussed), \emph{tenor} (the relationship between the participants in the discourse and their purpose) and \emph{mode} (communication medium, such as `spoken' or `written'). Generally, the selection of pre-training data in existing domain-specific models is mainly based on the field rather than the tenor. For example, BioBERT~\citep{Lee:Yoon:Bioinformatics:2020} and SciBERT~\citep{Beltagy:Lo:EMNLP:2019} are both pre-trained on scholar articles, but on different fields (biology and computer science). 

We first show in Section~\ref{section-select-pretraining-data-human-intuition} that human intuition regarding selecting pre-training data varies across practitioners, motivating our work on employing quantitative measures to nominate in-domain pre-training data. We then describe several measures which can quantify similarity between two datasets in Section~\ref{section-select-pretraining-data-similarity-measure}. We pre-train several domain-specific language representation models on different sources, and investigate their effectiveness on various downstream NER datasets, respectively (Section~\ref{section-select-pretraining-data-experiments}). Finally, through correlation analysis, we show that simple similarity measures can be used to nominate in-domain pre-training data (Section~\ref{section-select-pretraining-data-analysis}). 

\section{What Human Intuition Indicates}
\label{section-select-pretraining-data-human-intuition}
\begin{figure}[p] 
	\centering
	\includegraphics[width=1\textwidth]{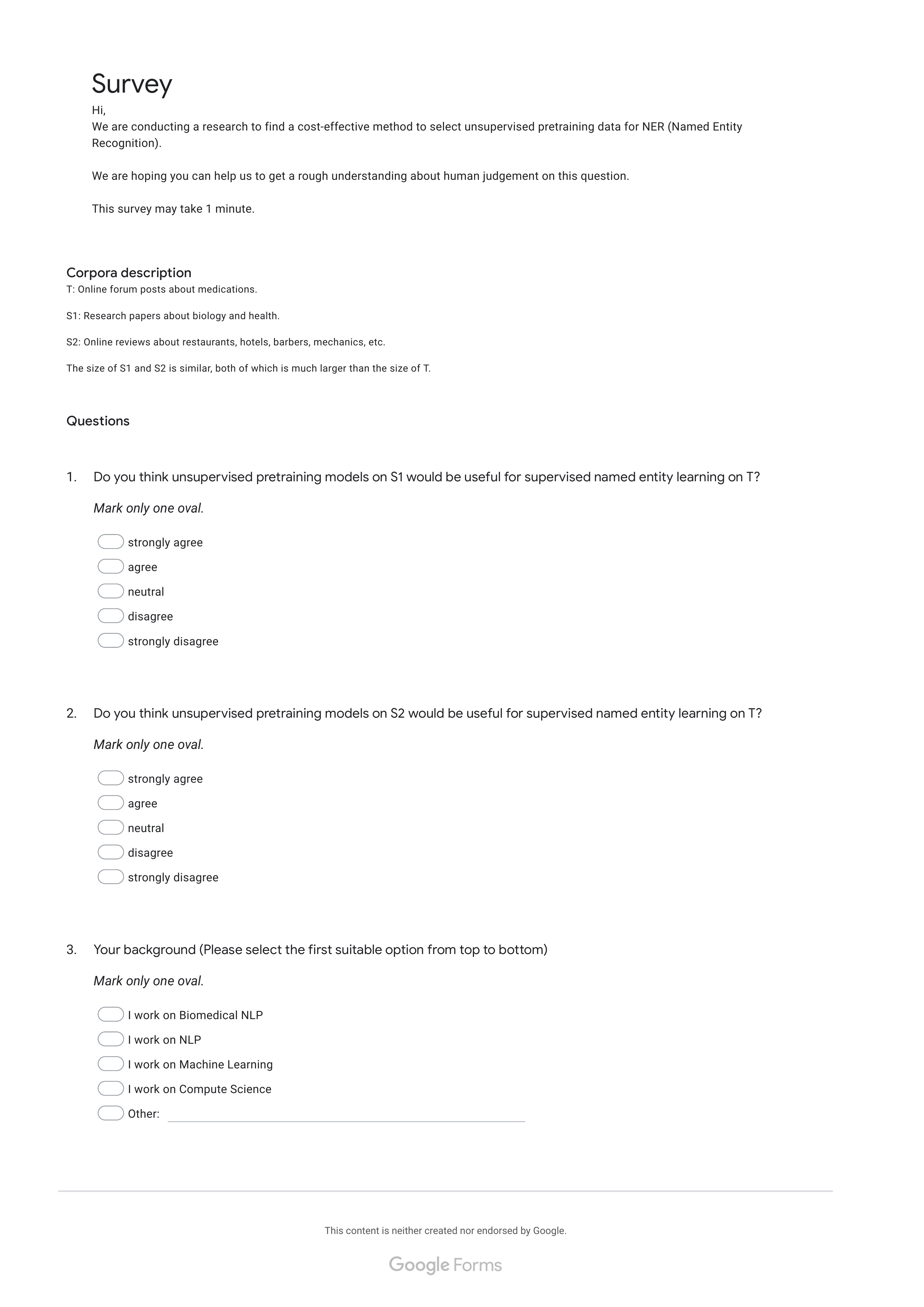}
	\caption{Survey questions regarding selection of pre-training data.~\label{figure-select-pretraining-data-survey-screen}}
\end{figure}

We surveyed $30$ NLP or machine learning practitioners to learn the human intuition regarding selection of pre-training data. Participants were provided short descriptions of the target data T, and two possible source data S1 and S2 as
\begin{itemize}
	\item T: Online forum posts about medications;
	\item S1: Research papers about biology and health; 
	\item S2: Online reviews about restaurants, hotels, barbers, mechanics, etc.
\end{itemize} 
A screenshot is shown in Figure~\ref{figure-select-pretraining-data-survey-screen}.

We constructed each of the descriptions as `$t$ about $f$' where $t$ is intended to indicate the tenor and $f$ the field. Each participant rated both sources on a five-point Likert, indicating agreement with the statement \textit{``Unsupervised pre-training on S would be useful for supervised named entity recognition learning on T''}. 

\begin{figure}[tb] 
	\centering
	\includegraphics[width=0.95\textwidth]{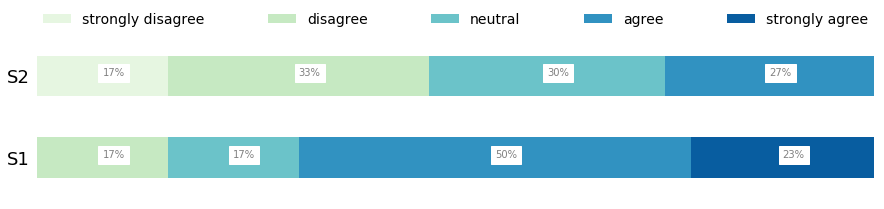}
	\caption{Likert scale ratings from NLP and ML practitioners ($N=30$) for the statement \textit{`Unsupervised pre-training on S would be useful for supervised named entity recognition learning on T.'} Target data T is described as \textit{`Online forum posts about medications,'} source data S1 as \textit{`Research papers about biology and health,'} and source data S2 as \textit{`Online reviews about restaurants, hotels, barbers, mechanics, etc.'}~\label{figure-select-pretraining-data-human-intuition}}
\end{figure}

Survey results show that $73$\% of the participants agreed or strongly agreed that S1---sharing similar \emph{biomedical} field with the target---would be useful. Only $27$\% agreed that S2---sharing similar \emph{social media} tenor with the target---would be useful (Figure~\ref{figure-select-pretraining-data-human-intuition}).  
On the one hand, a Wilcoxon signed-rank test indicates that scores are significantly higher for S1 than for S2 ($Z = 43.0, p < 0.001$). On the other hand, these results show the variety across practitioners, motivating our work on employing quantitative measures to nominate pre-training data. Our empirical investigations (detailed in Section~\ref{section-select-pretraining-data-experiments}) also suggest that human intuition maybe unreliable regarding selecting pre-training data. That is, practitioners favour field over tenor when selecting pre-training data, and this would be detrimental to accuracy of the target NER tasks. 

\section{Similarity Measures}\label{section-select-pretraining-data-similarity-measure}
Recall that the context specific usage of language is affected by three factors: field, tenor and mode~\citep{Halliday:Hasan:1989}. Researchers who select pre-training data from a similar field believe that, if the source data has a similar field to the target data, they tend to share similar topical vocabulary. Conversely, vocabularies are different from each other if source and target are from different fields. Imagine datasets about medications and restaurants. Those who select pre-training data from a similar tenor believe that tenor may impact the writing style of text. Imagine the participants in online reviews and scientific papers, their relationships to each other, their purposes and how these affect text style, including punctuation, lexical normalisation, politeness, emotiveness and so on~\citep{Lee:LLT:2001,Solano-Flores:2006}. We do not explicitly consider mode, because all of the datasets studied in this thesis are written text. 

Below, we detail different measures based on these intuitions to quantify different aspects of similarity between two datasets. 

\subsection{Target vocabulary covered}
The first measure is simply the percentage of the target vocabulary that is also present in the source data. An extremely dissimilar example is that of different languages. They have a totally different vocabulary and are considered dissimilar, even if they are written in a similar style and talking about the same subject. Note that our focus is on transferring through pre-trained models using one single source and we do not consider multilingual similarity. We propose \emph{Target Vocabulary Covered (TVC)} as a measure of field, calculated as
\[
TVC (D_S, D_T) = \frac{|V_{D_S} \cap V_{D_T}|}{|V_{D_T}|},
\]
where $V_{D_S}$ and $V_{D_T}$ are sets of unique content words (nouns, verbs, adjectives) in source and target datasets respectively. 

\subsection{Jaccard similarity of vocabularies}
Jaccard similarity coefficient~\citep{Agresti:2003}, is a statistic used for estimating the similarity and diversity of two sets. By calculating
\[
JSC (D_S, D_T) = \frac{|V_{D_S} \cap V_{D_T}|}{|V_{D_S} \cup V_{D_T}|},
\]
\emph{Jaccard Similarity of Vocabularies (JSV)} can be used to measure the similarity between source and target vocabularies, meanwhile, factoring out the source vocabulary size.

\subsection{Language model perplexity}
A language model~\citep{Schutze:Manning:2008} assigns a probability to any sequence of words $[w_1,\cdots,w_N]$ using chain rule of probability:
\[
p(w_1, w_2, \cdots, w_N) = \prod_{i=1}^N p(w_i|w_1^{w-1}),
\]
where $N$ is the length of the sequence and $w_1^{i-1}$ are all words before word $w_i$. In practice, this equation can be simplified by n-gram models based on Markov Assumption:
\[
p(w_1, w_2, \cdots, w_N) = \prod_{i=1}^N p(w_i|w_{i-n+1}^{i-1}),
\]
where $w_{i-n+1}^{i-1}$ represents only $n$ preceding words of $w_i$. To make the model generalise better, smoothing techniques can be used to assign non-zero probabilities to unseen events. We use Kneser-Ney smoothed 3-gram models~\citep{Heafield:SMT:2011} to measure the similarity between two datasets. Specifically, we first train the language model on the source data, then evaluate it on the target data using perplexity to represent the degree of similarity. The intuition is that, if the model finds a sentence very unlikely (dissimilar from the data where this language model is trained on), it will assign a low probability and therefore high perplexity. The summed up \emph{perplexity (PPL)} is then:
\[
PPL(D_S, D_T) = \sum_{i=1}^{m} P(D_T^i)^{-\frac{1}{N_i}},
\]
where $m$ is the number of sentences in the target data set, and $P(D_T^i)$ is the probability assigned by the language model trained on the source data to the $i$-th sentence from the target data set, whose sentence length is $N_i$. 

Similar to TVC, PPL is token-based but also captures surface structure. We therefore propose PPL as a proxy to measure tenor as well as field.

\subsection{Jensen-Shannon divergence}
Jensen-Shannon divergence (JSD), based on term distributions, has been successfully used for domain adaptation~\citep{Ruder:Plank:EMNLP:2017}. We first measure the probability of each term (up to 3-gram) in source $S$ and target data $T$, separately. Then, we use the Jensen-Shannon divergence~\citep{Fuglede:Topsoe:IEEE:2004} between
these two probability distributions
\[
JSD(S || T) = \frac{1}{2} KL (S || M) + \frac{1}{2} KL (T || M),
\]
where
\[
KL (P || Q) = \sum_{x \in X} P(x) \log (\frac{P(x)}{Q(x)})
\]
and 
\[
M = \frac{1}{2} (S + T)
\]
as a proxy to measure tenor as well as field.

\section{Datasets}
\label{section-select-pretraining-data-experiments}
We use six datasets as source data, covering a range of fields and tenors:
\begin{description}
\item[News] The original one billion word language model benchmark data~\citep{Chelba:Mikolov:arXiv:2013}, produced from News Crawl data. \emph{Popular reporting} usually involves one or several writers, and a large number of readers. The text is usually edited several times for easy understanding.
\item[Books] A corpus of 11,038 books in 16 different genres, e.g., Romance, Fantasy, Science fiction, etc., collected by~\citet{Zhu:Kiros:ICCV:2015}. There books are all free books written by yet unpublished authors. \emph{Fiction books} usually involves one writer, and a moderate size of readers. The text is usually edited many times, reflecting writer personality.
\item[MIMIC] A clinical database comprising over 58,000 hospital admissions for intensive care unit (ICU) patients~\citep{Johnson:Pollard:SD:2016}. \emph{Clinical notes} are usually written by doctors and nurses under time pressure, and read by their colleagues. The text is seldom carefully edited, so it is syntactically noisy and usually contains a lot of jargon for efficient communication.
\item[PubMed] Titles and abstracts of biomedical scholar articles. \emph{Scholar articles} are usually written by a small groups of writers, and the readers usually have similar knowledge background with authors. The text is usually edited many times for more comprehensible and less ambiguous.
\item[Yelp] Crowd-sourced reviews about local businesses, including restaurants, hotels, barbers, mechanics, etc. \emph{Online review} is usually written by a single customer, and read by a small group of people who are interested in the business. The text is usually edited once, and writer tends to use descriptive language to share their experiences.
\item[Wikipedia] A free online encyclopedia. \emph{Online encyclopedia} is created and edited by volunteers around the world, and it has around 250 million page views every day~\footnote{Siteviews Analysis: shorturl.at/ayQR5. Accessed data: 2021-Jan-31.}. Another important feature of Wikipedia as an open-collaborative website is that articles are edited all the time by human users and bots, reflecting the newest development of the world knowledge.
\end{description}

To isolate the impact of source data size, we randomly sample all source data to approximately 700 million tokens. The only exception is on MIMIC. Although all text from MIMIC data set has been used, it is still relatively small, comparing to other sources. The data statistics of source data is listed in Table~\ref{table-select-pretraining-data-sourcedata-statistics}. Based on the number of unique tokens and average sentence length, we can see these sources are roughly split into two categorise: formal text---PubMed, Wikipedia and News---with large vocabulary and long sentences, and informal text---Books, MIMIC, and Yelp---with small vocabulary and short sentences.

\begin{table}[tb]
    \centering
    \begin{tabular}{l|c|c|c|c}
    \toprule
    Source & \# sentences & \# tokens & \# unique tokens & Avg. sentence length \\
    \hline
Books & 53.9M & 0.69B & 0.5M & 12.8 \\
MIMIC & 61.9M & 0.61B & 0.5M & 9.8 \\
News & 27.5M & 0.70B & 2.1M & 25.4 \\
PubMed & 29.3M & 0.69B & 4.2M & 23.5 \\
Wikipedia & 31.1M & 0.69B & 3.3M & 22.1 \\
Yelp & 50.7M & 0.69B & 1.0M & 13.5 \\
    \bottomrule
    \end{tabular}
    \caption{\label{table-select-pretraining-data-sourcedata-statistics}Descriptive statistics of the source datasets.}
\end{table} 

Ten NER datasets are used as target data: BC2GM (BioCreative II Gene Mention Recognition)~\citep{Smith:Tanabe:2008}, BTC (Broad Twitter Corpus)~\citep{Derczynski:Bontcheva:COLING:2016}, CADEC (CSIRO Adverse Drug Event Corpus)~\citep{Karimi:Metke-Jimenez:JBI:2015}), CoNLL 2003~\citep{Sang:Meulder:CONLL:2003}, EBM (Evidence Based Medicine)~\citep{Nye:Li:ACL:2018}, i2b2 2010~\citep{Uzuner:South:AMIA:2011}, JNLPBA~\citep{Kim:Ohta:BioNLP:2004}, \NCBIDISEASE~\citep{Dogan:Leaman:JBI:2014}, SciERC~\citep{Luan:He:EMNLP:2018}, WetLab~\citep{Kulkarni:Xu:NAACL:2018}, and W-NUT 2016~\citep{Strauss:Toma:WNUT:2016}. 
Details of these target data are listed in Table~\ref{table-select-pretraining-data-targetdata}. 

\begin{table}[tb]
	\begin{small}
		\begin{center}
			\begin{tabular}{p{0.15\linewidth}p{0.4\linewidth}p{0.4\linewidth}}
				\toprule 
				\bf Target & \bf Entity Categories & \bf Description \\ \midrule
				BC2GM & Gene & Biomedical scholar articles \\ \midrule
				BTC & Person, Organisation, Location & Tweets sampled across different regions, temporal periods, and types of Twitter users \\ \midrule
				CADEC & Adverse Drug Event, Disease, Drug, Finding, Symptom & Posts taken from AskaPatient, which is a forum where consumers can discuss their experiences with medications. \\ \midrule
				EBM & Intervention, Outcome and Comparator  & Scholar articles about clinical trials \\ \midrule
				i2b2 2010 & Problem, Treatment and Test & Clinical notes about health \\ \midrule
				JNLPBA & Protein, DNA, RNA, Cell line and Cell
				type & Abstract of journal articles about biology. \\ \midrule
				NCBI-disease & Disease & Abstract of journal articles about health. \\ \midrule
				SciERC & Generic, Material, Method, Metric, Other-Scientific-Term, Task & Journal articles about Computer Science, Material Sciences and Physics \\ \midrule
				Wetlab & Action, 9 object-based (Amount, Concentration, Device, Location, Method, Reagent, Speed, Temperature, Time) entity types, 5 measure-based (Numerical, Generic-Measure, Size, pH, Measure-Type) and 3 other (Mention, Modifier, Seal) types & Protocols written by researchers about conducting biology and chemistry experiments. \\
				\bottomrule
			\end{tabular}
			\caption{\label{table-select-pretraining-data-targetdata}List of the target NER datasets and their specifications.}
		\end{center}
	\end{small}
\end{table}

\section{Experimental Results}
\label{section:select-pretraining-data-similarity_results}

\paragraph{Similarity Between Source and Target Datasets}
The results shown in Table~\ref{table-select-pretraining-data-similarity-values-1} and~\ref{table-select-pretraining-data-similarity-values-2} indicate that PubMed is the most similar source to most of these target datasets from the Biomedical domain. It achieves lower language model perplexity, higher target vocabulary covered, and Jensen-Shannon Divergence when evaluated against BC2GM, EBM, JNLPBA, NCBI-disease, SciERC and Wetlab compared to other sources. On one hand, it is expected that PubMed is similar to BC2GM, EBM, JNLPBA, NCBI-disease and Wetlab, since they are all scientific writing about biology and health, thus being similar in terms of both field and tenor. On the other hand, although SciERC does not have the same field as PubMed (computer science, material and physics versus biology and health), they are similar because they share a similar tenor (scholarly publications). On i2b2-2010 (clinical notes), only the target vocabulary covered measure indicates PubMed is the most similar source, whereas other three metrics indicate MIMIC as the most similar source.


\begin{table}[tb]
	\centering
\begin{tabular}{r l | c c c c }
\toprule
& & \multicolumn{4}{c}{\bf Similarity} \\ 
\bf Target & \bf Source & \bf TVC (\%) & \bf JSV (\%) \bf & \bf PPL & \bf JSD \\ 
\midrule
\multirow{6}{*}{BC2GM} & Books & 37.39 & 9.54 & 109.65 & 36.29 \\ 
& MIMIC & 41.95 & \bf 13.69 & 101.39 & 38.17 \\ 
& News & 48.12 & 6.15 & 101.34 & 38.02 \\ 
& PubMed & \bf 81.20 & 7.24 & \bf 75.43 & \bf 48.32 \\ 
& Wikipedia & 60.19 & 5.05 & 93.10 & 39.28 \\ 
& Yelp & 36.26 & 8.70 & 109.11 & 37.01 \\ 
\midrule
\multirow{6}{*}{BTC} & Books & 47.96 & \bf 10.09 & 61.63 & 38.44 \\ 
& MIMIC & 26.80 & 6.83 & 68.54 & 33.92 \\ 
& News & \bf 54.92 & 5.58 & \bf 59.49 & 36.61 \\ 
& PubMed & 41.97 & 2.84 & 71.44 & 33.48 \\ 
& Wikipedia & 54.46 & 3.57 & 61.19 & 35.00 \\ 
& Yelp & 47.58 & 9.40 & 60.58 & \bf 39.22 \\ 
 \midrule
\multirow{6}{*}{CADEC} & Books & 80.16 & 5.08 & 47.37 & 42.72 \\ 
& MIMIC & 78.21 & \bf 6.54 & 47.92 & 38.21 \\ 
& News & \bf 85.59 & 2.47 & 45.90 & 39.67 \\ 
& PubMed & 82.03 & 1.57 & 52.38 & 37.69 \\ 
& Wikipedia & 84.41 & 1.55 & 49.41 & 38.18 \\ 
& Yelp & 81.89 & 4.85 & \bf 45.52 & \bf 44.82 \\ 
 \midrule
\multirow{6}{*}{EBM} & Books & 29.68 & 10.29 & 146.01 & 36.03 \\ 
& MIMIC & 32.61 & \bf 14.06 & 130.42 & 37.70 \\ 
& News & 44.30 & 8.21 & 125.28 & 38.47 \\ 
& PubMed & \bf 70.66 & 9.30 & \bf 91.87 & \bf 51.85 \\ 
& Wikipedia & 47.20 & 5.79 & 125.61 & 39.00 \\ 
& Yelp & 29.94 & 9.85 & 142.15 & 36.82 \\ 
\bottomrule
\end{tabular}
		\caption{\label{table-select-pretraining-data-similarity-values-1}Similarity values measured between source and target datasets. TVC: Target Vocabulary Covered. JSC: Jaccarrd similarity of Vocabularies. PPL: language model perplexity. JSD: Jensen-Shannon Divergence based on term distributions.}
\end{table}

\begin{table}[tb]
	\centering
\begin{tabular}{r l | c c c c }
\toprule
& & \multicolumn{4}{c}{\bf Similarity} \\ 
\bf Target & \bf Source & \bf TVC (\%) & \bf JSV (\%) \bf & \bf PPL & \bf JSD \\ 
\midrule
\multirow{6}{*}{i2b2 2010} & Books & 44.65 & 5.74 & 45.55 & 37.75 \\ 
& MIMIC & 58.36 & \bf 9.95 & \bf 29.29 & \bf 48.99 \\ 
& News & 56.28 & 3.42 & 43.32 & 37.30 \\ 
& PubMed & \bf 64.86 & 2.64 & 38.92 & 38.23 \\ 
& Wikipedia & 59.59 & 2.32 & 42.24 & 37.94 \\ 
& Yelp & 45.76 & 5.52 & 44.75 & 37.92 \\ 
 \midrule
\multirow{6}{*}{JNLPBA} & Books & 31.25 & 5.87 & 105.66 & 35.84 \\ 
& MIMIC & 32.30 & \bf 7.74 & 101.82 & 36.25 \\ 
& News & 40.46 & 3.72 & 97.64 & 37.31 \\ 
& PubMed & \bf 71.49 & 4.52 & \bf 65.00 & \bf 47.80 \\ 
& Wikipedia & 46.46 & 2.78 & 91.20 & 38.29 \\ 
& Yelp & 30.46 & 5.38 & 107.22 & 36.55 \\ 
  \midrule
\multirow{6}{*}{NCBI-disease} & Books & 54.78 & 5.05 & 95.03 & 36.07 \\ 
& MIMIC & 57.02 & \bf 6.90 & 90.02 & 37.24 \\ 
& News & 65.84 & 2.82 & 86.97 & 37.43 \\ 
& PubMed & \bf 87.15 & 2.50 & \bf 64.48 & \bf 44.46 \\ 
& Wikipedia & 76.17 & 2.08 & 80.70 & 38.39 \\ 
& Yelp & 52.78 & 4.55 & 95.63 & 36.82 \\ 
  \midrule
\multirow{6}{*}{SciERC} & Books & 70.19 & 4.34 & 88.42 & 36.11 \\ 
& MIMIC & 57.06 & \bf 4.60 & 93.64 & 35.31 \\ 
& News & 77.67 & 2.19 & 80.88 & 37.03 \\ 
& PubMed & \bf 84.14 & 1.58 & \bf 71.03 & \bf 39.72 \\ 
& Wikipedia & 81.23 & 1.46 & 77.72 & 37.56 \\ 
& Yelp & 67.37 & 3.88 & 89.39 & 36.89 \\ 
  \midrule
\multirow{6}{*}{Wetlab} & Books & 48.82 & 3.97 & 60.62 & 36.43 \\ 
& MIMIC & 44.69 & \bf 4.74 & 58.02 & 36.52 \\ 
& News & 58.06 & 2.19 & 57.47 & 36.11 \\ 
& PubMed & \bf 68.54 & 1.72 & \bf 52.11 & \bf 37.15 \\ 
& Wikipedia & 61.24 & 1.47 & 55.99 & 36.41 \\ 
& Yelp & 50.29 & 3.83 & 57.84 & 37.12 \\ 
\bottomrule
\end{tabular}
		\caption{\label{table-select-pretraining-data-similarity-values-2}Similarity values measured between source and target datasets (continued). TVC: Target Vocabulary Covered. JSC: Jaccarrd similarity of Vocabularies. PPL: language model perplexity. JSD: Jensen-Shannon Divergence based on term distributions.}
\end{table}

The second observation is that tenor might be reflected more than field by these measures. Source data Yelp is more similar to CADEC than PubMed and MIMIC from both language model perplexity and Jensen-Shannon Divergence perspectives. CADEC is a data set focusing on recognising drugs, diseases and adverse drug events. The field of CADEC is therefore more similar to PubMed which includes journal articles in health discipline and MIMIC which contains clinical notes. However, CADEC is written by patients, and can be considered as `drug reviews'. The tenor is therefore closer to the one in Yelp, where customers use informal language to describe their experiences. Target vocabulary covered nominates News as the most similar source. Note that News has a moderate size vocabulary, 2.1 millions unique tokens, whereas the vocabulary size of PubMed is 4.2 million.

The last observation is that using different measures can lead to almost the same answer regarding {\em which source is the most similar one to a given target}, except for the Jaccard similarity of vocabularies. Using Jaccard similarity of vocabularies measure, MIMIC source is nominated as the most similar source against target sets, except for BTC. This might be explained by the fact that the vocabulary size of MIMIC is the smallest one in all sources, and Jaccard similarity of vocabularies measure favours sources with small vocabulary size than the ones with large vocabulary size.


\paragraph{Effectiveness of domain-specific models on downstream NER tasks}\label{section:select-pretraining-data-impactpretrainingdata}
After we quantify the similarity between source and target datasets, the next step is to investigate the impact of source data on pre-trained language representation models. We pre-train ELECTRA~\citep{Clark:Luong:ICLR:2020}---a sample-efficient variant of BERT---on different sources separately, then observe how the effectiveness of these pre-trained models varies in different downstream NER datasets. 

The most common approach of training domain-specific models is \emph{continue pre-training}, which is used by BioBERT~\citep{Lee:Yoon:Bioinformatics:2020}, ClinicalBERT~\citep{Alsentzer:Murphy:ClinicalNLP:2019}, and BERTweet~\citep{Nguyen:Vu:EMNLP:2020}. Continue pre-training approach starts from an existing pre-trained model---usually pre-trained on large size generic data set---and continues training on a domain-specific corpus. The main advantage of continue pre-training is that they can inherit knowledge from language representation models pre-trained on generic data, and thus be considered as capturing both generic domain and domain-specific knowledge. However, we aim to investigate the impact of pre-training data, therefore, we use the \emph{learning from scratch} approach, eschewing the potential impact of generic data. We follow the hyper-parameter setting in~\citep{Clark:Luong:ICLR:2020}, shown in Table~\ref{table-select-pretraining-data-pretraining-hyperparameters} to train the domain-specific models. Training of each model took four days using 1 Nvidia Tesla v100 GPU.
 
 \begin{table}[tb]
    \centering
    \begin{tabular}{l|r}
    \toprule
    \bf Hyper-parameter & \bf Value \\
    \midrule
    Number of layers & 12 \\
    Hidden size & 256 \\ 
    Intermediate size & 1024 \\
    Attention heads & 4 \\
    Attention head size & 64 \\
    Embedding size & 128 \\
    Learning rate & 5e-4 \\
    Train steps & 800K \\
    Vocab size & 30,994 \\
    \bottomrule
    \end{tabular}
    \caption{Pre-train hyper-parameters, which follow the practice of training ELECTRA-SMALL in~\citep{Clark:Luong:ICLR:2020}.}
    \label{table-select-pretraining-data-pretraining-hyperparameters}
\end{table}
 
Evaluation results using these domain-specific models on downstream NER tasks show that the effectiveness varies in different target datasets (Table~\ref{table-select-pretraining-data-ner-main-results}). In other words, no single source is suitable for all target NER datasets. It is worthy note that most of these results (for example on BC2GM, BTC, CADEC, SciERC) are lower than state-of-the-art results on these datasets with large margin. This is mainly because our pre-trained models are smaller---smaller hidden size, less number of attention heads, smaller embedding size---than the ones in other studies.

The best performing models on each target data are all pre-trained on the most similar source which is nominated by at least one similarity measure, except for SciERC  (Table~\ref{table-select-pretraining-data-similarity-values-1} and~\ref{table-select-pretraining-data-similarity-values-2}). 

\begin{table}[tb]
	\centering
		\begin{tabular}{r|c|c|c|c|c|c}
			\toprule
			& \bf Book & \bf MIMIC & \bf News & \bf PubMed & \bf Wiki & \bf Yelp \\ \midrule
BC2GM & 72.4 (0.2) & 72.7 (0.5) & 74.4 (0.4) & \bf 80.7 (0.2) & 75.1 (0.4) & 72.4 (0.3) \\
BTC & 70.4 (0.2) & 63.1 (0.3) & \bf 75.2 (1.3) & 67.1 (0.7) & 74.9 (0.4) & 70.8 (0.5) \\
CADEC & 64.3 (0.6) & \bf 67.5 (0.3) & 65.1 (0.3) & 66.1 (0.5) & 65.5 (0.5) & 66.1 (0.6) \\
EBM & 39.8 (0.4) & 41.0 (0.4) & 41.1 (0.6) & \bf 43.5 (0.3) & 40.7 (0.4) & 40.8 (0.5) \\
i2b2-2010 & 78.7 (0.3) & \bf 87.7 (0.2) & 79.8 (0.9) & 85.1 (0.3) & 79.6 (0.4) & 79.4 (0.3) \\
JNLPBA & 69.7 (0.2) & 70.1 (0.3) & 70.1 (0.7) & \bf 73.1 (0.2) & 70.7 (0.2) & 70.0 (0.2) \\
NCBI-Disease & 77.7 (0.5) & 80.3 (0.7) & 77.9 (4.8) & \bf 85.8 (0.5) & 80.8 (0.3) & 79.2 (0.8) \\
SciERC & 37.2 (4.1) & 23.8 (1.3) & 25.9 (2.2) & 41.0 (20.6) & \bf 47.3 (1.4) & 38.7 (3.1) \\
WetLab & 78.2 (0.1) & 78.1 (0.2) & 78.1 (0.3) & \bf 78.7 (0.2) & 78.3 (0.1) & 78.0 (0.1) \\
			\bottomrule
		\end{tabular}
		\caption{\label{table-select-pretraining-data-ner-main-results}The effectiveness of domain-specific pre-trained models on downstream NER tasks. We report the mention level $F_1$ scores.}
\end{table}

\subsection{Predictiveness of similarity measures~\label{section-select-pretraining-data-analysis}}

To analyse how proposed similarity measures can be used to nominate the best pre-training data option, we investigate the correlation between these similarity values and the effectiveness of pre-trained models on target tasks. Specifically, we employ Spearman rank-order  correlation coefficient to measure the relationship between the ranking of similarity values and NER results. For example, given the target data set NCBI-disease, the rank of sources is PubMed, Wikipedia, MIMIC, Yelp, News, and Book, if they are sorted based on the effectiveness of different domain-specific models. Similarity, if they are sorted based on the target vocabulary covered measure, the rank of sources is PubMed, Wikipedia, News, MIMIC, Books, and Yelp. The Spearman rank-order correlation coefficient between these two rankings is 0.71.

\begin{figure}[tb]
    \centering
    \includegraphics[width=0.6\textwidth]{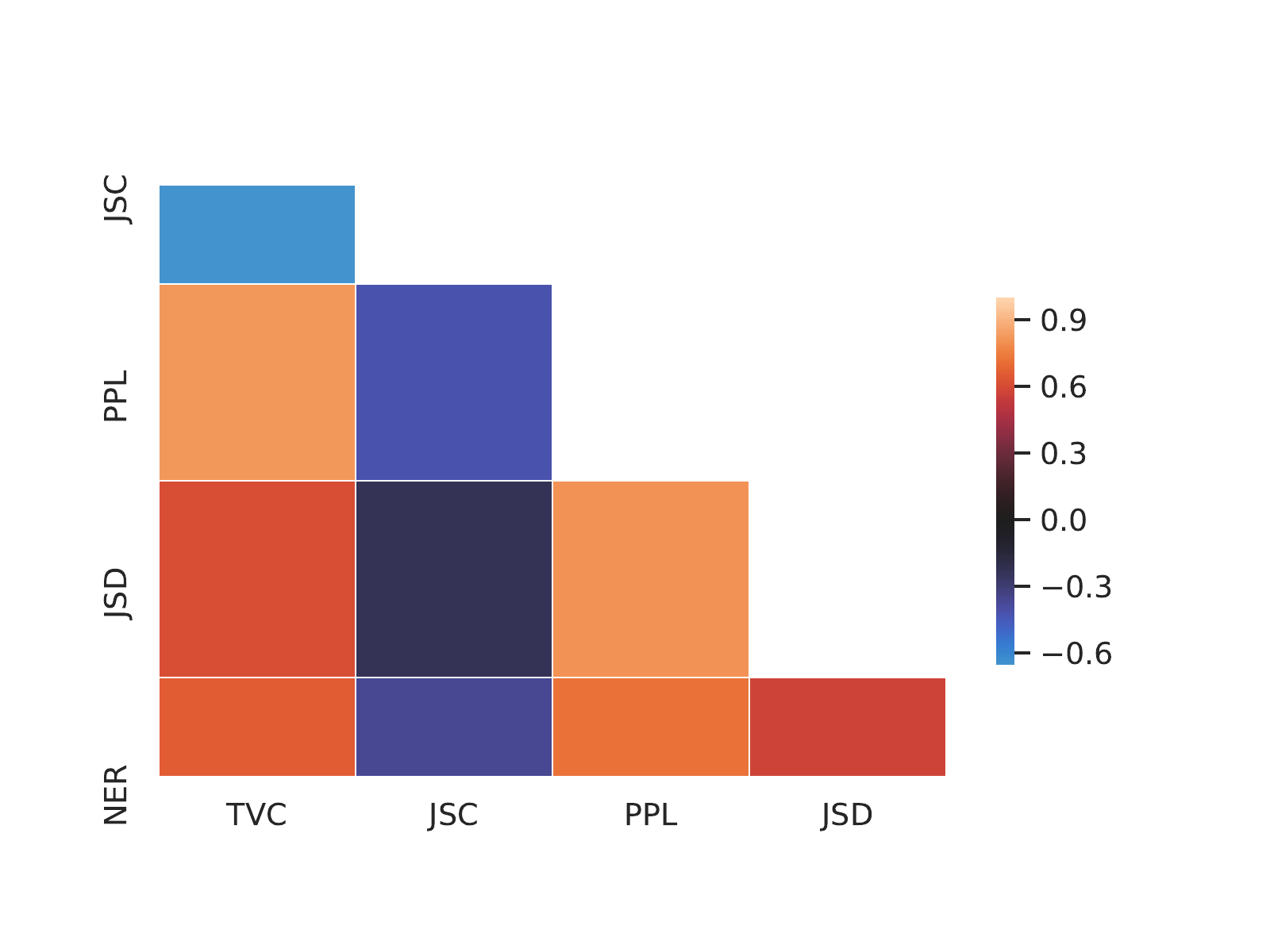}
    \caption{Correlation between different similarity measures and the effectiveness of domain-specific pre-trained models.}
    \label{figure-select-pretraining-data-predictiveness}
\end{figure}

The results in Figure~\ref{figure-select-pretraining-data-predictiveness} show that these proposed similarity measures are predictive of the effectiveness of the pre-training data, except for Jaccard similarity of vocabularies. 

\subsection{Comparison to publicly available pre-trained models}
Literature shows substantial improvements are sometimes possible when pre-training on large generic corpora~\citep{Liu:Ott:arXiv:2019,Baevski:Edunov:EMNLP:2019}. Given that pre-trained models are freely available, is it even necessary to pre-train on similar data as proposed above? We compare to publicly available ELECTRA models trained on 3.3 Billion tokens generic data. Note that the publicly available model we choose has the same model size. It is also pre-trained using the same hyper-parameters as ours, except that it is pre-trained longer than ours (1.45M vs.0.8M).

\begin{table}[tb]
	\centering
		\begin{tabular}{r|c|c}
			\toprule
			& \bf Domain-specific model & \bf Generic domain model \\ \midrule
BC2GM & \bf 80.7 (0.2) & 63.6 (0.7) \\
BTC & 75.2 (1.3) & \bf 75.3 (0.2) \\
CADEC & \bf 67.5 (0.3) & 57.9 (0.3) \\
EBM & \bf 43.5 (0.3) & 41.6 (0.2) \\
i2b2-2010 & \bf 87.7 (0.2) & 82.5 (0.1) \\
JNLPBA & \bf 73.1 (0.2) & 71.2 (0.4) \\
NCBI-Disease & \bf 85.8 (0.5) & 81.8 (0.6) \\
SciERC & \bf 47.3 (1.4) & \bf 47.3 (0.5) \\
WetLab & 78.7 (0.2) & \bf 78.8 (0.1) \\
			\bottomrule
		\end{tabular}
		\caption{\label{table-select-pretraining-data-ner-compare-to-public-models}Comparison between our best performing domain-specific models and the publicly available generic domain model.}
\end{table}

These results, shown in Table~\ref{table-select-pretraining-data-ner-compare-to-public-models} indicate that a small similar source reduces the computational cost without sacrificing the performance. This is especially important in practice, because collecting data and pre-training models are expensive, in terms of both computational and environmental cost~\citep{Schwartz:Dodge:arXiv:2019}.

\section{Summary}
This chapter focuses on whether there are cost-effective methods to nominate datasets to pretrain language representation models that are building blocks of NER models. 
We propose using different measures to measure different aspects of similarity between source and target data. We investigate how these measures correlate with the effectiveness of pre-trained models for NER tasks. 
While different NLP tasks may rely on different aspects of language, our study is a step towards systematically guiding researchers on their choice of data for pre-training, and models pre-trained on small size domain-specific corpus can outperform the one pre-trained on large size generic domain data.

%% file: src/ch6-discontinuous-ner.tex
Discontinuous mentions represent compositional concepts, for example disorders or symptoms, that differ from concepts represented by individual components, for example body locations or general feelings. In downstream applications such as pharmacovigilance and summarization, recognising these discontinuous mentions is more useful than recognising separate components. We propose a transition-based model that can effectively recognise discontinuous mentions without sacrificing the accuracy on continuous mentions. 

\section{Overview}

NER is a critical component of biomedical text mining applications. In pharmacovigilance, it can be used to identify adverse drug events in consumer reviews in online medication forums, alerting medication developers, regulators, and clinicians~\citep{Leaman:Wojtulewicz:BioNLP:2010,Sarker:Ginn:JBI:2015,Karimi:Wang:Survey:2015}. In clinical settings, NER can be used to extract and summarise key information from electronic medical records such as conditions hidden in unstructured doctors' notes~\citep{Feblowitz:Wright:JBI:2011,Wang:Wang:JBI:2018}. These applications require identification of complex entity mentions, discontinuous and overlapping mentions, not seen in generic domains. 

\begin{figure}[tb] 
	\centering
	\begin{subfigure}[b]{0.5\textwidth}
		\includegraphics[width=0.8\textwidth]{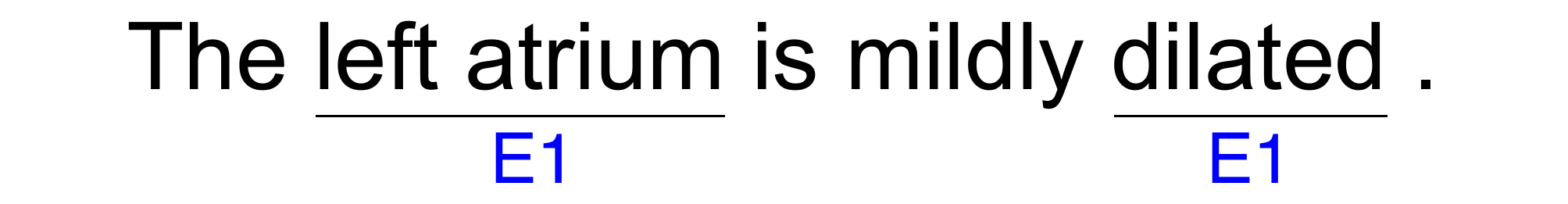}
	\end{subfigure}
	\\
	\vspace{1em}
	\begin{subfigure}[b]{0.5\textwidth}
		\includegraphics[width=0.8\textwidth]{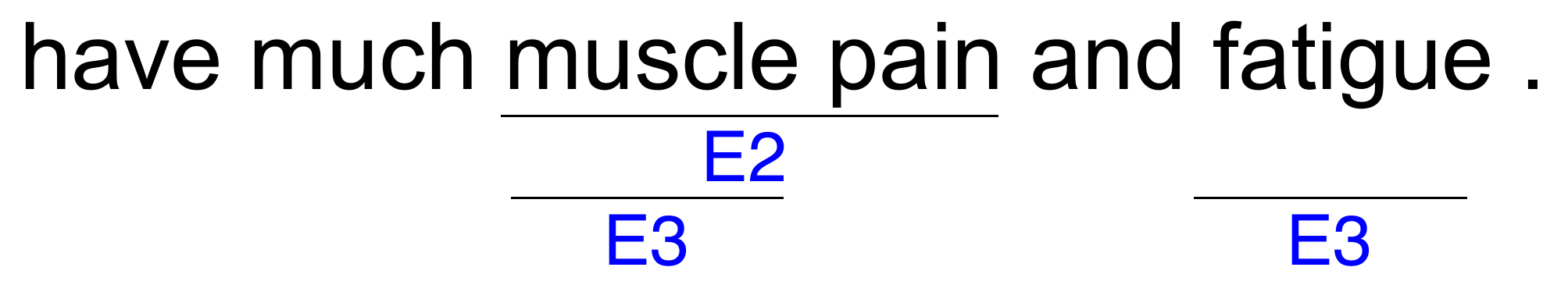}
	\end{subfigure}
	\caption{Examples involving discontinuous mentions, taken from the \SHARECLEF 13~\citep{Pradhan:Elhadad:CLEF:2013} and CADEC~\citep{Karimi:Metke-Jimenez:JBI:2015} datasets, respectively. The first example contains a discontinuous mention \textit{`left atrium dilated'}, the second example contains two mentions that overlap: \textit{`muscle pain'} and \textit{`muscle fatigue'} (discontinuous).~\label{figure-discontinuous-ner-example}}
\end{figure}

Widely used sequence tagging techniques encode two assumptions that do not always hold: (1) mentions do not overlap, therefore each token can belong to at most one mention; and, (2) mentions comprise continuous sequences of tokens. Nested entity recognition addresses violations of the first assumption (more discussions in Section~\ref{section-complex-ner}). However, the violation of the second assumption is comparatively less studied and requires handling discontinuous mentions (see examples in Figure~\ref{figure-discontinuous-ner-example}). 

In contrast to continuous mentions which are often short spans of text, discontinuous mentions consist of \emph{components} that are separated by \emph{intervals}. Recognising discontinuous mentions is particularly challenging as exhaustive enumeration of possible mentions, including discontinuous and overlapping spans, is exponential to sentence length. Existing approaches for discontinuous NER either suffer from high time complexity~\citep{McDonald:Crammer:EMNLP:2005} or ambiguity in translating intermediate representations into mentions~\citep{Tang:Cao:BMC:2013,Metke-Jimenez:Karimi:BMDID:2016,Muis:Lu:EMNLP:2016}. In addition, current arts use traditional approaches that rely on manually designed features, which are tailored to recognise specific entity categories. Also, these features usually do not generalise well in different types of text~\citep{Leaman:Khare:JBI:2015}. 

\paragraph{Motivations}
The main motivation for recognising discontinuous mentions is that they usually represent \emph{compositional concepts} that differ from concepts represented by individual components. For example, the mention \textit{`left atrium dilated'} in the first example of Figure~\ref{figure-discontinuous-ner-example} describes a disorder which has its own CUI (Concept Unique Identifier) in UMLS (Unified Medical Language System), whereas both \textit{`left atrium'} and \textit{`dilated'} also have their own CUIs. In downstream applications such as pharmacovigilance and summarization, recognising these discontinuous mentions that refer to disorders or symptoms is more useful than recognising separate components which may refer to body locations or general feelings. 

Another motivation for discontinuous NER is that discontinuous mentions usually overlap, and separating these overlapping mentions rather than identifying them as a single mention is important for downstream tasks, such as entity linking where the assumption is that the input mention refers to one entity~\citep{Shen:Wang:TKDE:2014}. 

In this chapter, we first characterise three datasets, from the biomedical domain, with a substantial number of discontinuous mentions (Section~\ref{section-discontinuous-ner-task-data}). Then, we introduce a transition-based model that can recognise discontinuous mentions (Section~\ref{section-discontinuous-ner-transition-model}). Through experiments, we show that our model can effectively recognise discontinuous mentions without sacrificing the accuracy on continuous mentions (Section~\ref{section-discontinuous-ner-experiment-results}). Analysis also suggests that our model is better than existing discontinuous NER models at handling long mentions and mentions that do not overlap or overlap at left, resulting in higher recall (Section~\ref{section-discontinuous-ner-analysis}). 

\section{Datasets~\label{section-discontinuous-ner-task-data}}
Although some text annotation tools, such as BRAT~\citep{Stenetorp:Pyysalo:EACL:2012}, allow discontinuous annotations, corpora annotated with large number of discontinuous mentions are still rare because they are hard to annotate. We describe three datasets from the biomedical domain that include a substantial number of discontinuous mentions: CADEC~\citep{Karimi:Metke-Jimenez:JBI:2015}, \SHARECLEF 2013~\citep{Pradhan:Elhadad:CLEF:2013} and \SHARECLEF 2014~\citep{Mowery:Velupillai:CLEF:2014}. We then motivate the discontinuous NER task. 

CADEC corpus is sourced from posts from \textit{AskaPatient}\footnote{https://www.askapatient.com/. Accessed data: \today}, a forum where patients can discuss their experiences with medications. The entity categories annotated in CADEC include drug, Adverse Drug Event (ADE), disease and symptom. In this work, we only consider the ADE annotations because only the ADEs involve discontinuous mentions. Note that ADEs in CADEC are defined as span of text that are clearly associated with a drug and should have the corresponding \textsc{MedDRA} (Medical Dictionary for Regulatory Activities) term. \SHARECLEF 2013, specifically Task 1 of the \SHARECLEF eHealth evaluation lab 2013, focuses on identification of disorder mentions in clinical reports. The corpus is sourced from de-identified clinical reports, including discharge summaries, electrocardiogram, echocardiogram, and radiology reports~\citep{Johnson:Pollard:SD:2016}. A disorder mention is defined as any span of text which can be mapped onto a concept in the Disorder semantic group of \textsc{SNOMED-CT} (Systematised Nomenclature of Medicine -- Clinical Terms). \SHARECLEF 2014, an extension of the \SHARECLEF 2013 task, focuses on template filling of disorder attributes. That is, given a disorder mention and its surrounding words, recognise the attributes of the disorder mention from its context, including subject class, severity indicator, uncertainty indicator. In this work, we frame \SHARECLEF 2014 as a disorder-NER dataset. That is, we consider only disorder annotations, without taking their attributes into consideration. 

\begin{table}[tb] 
\centering
\begin{tabular}{l c  c  c}
\toprule
& \multicolumn{3}{c}{\bf Dataset} \\ \midrule
& \bf CADEC & \bf \SHARECLEF 13 & \bf \SHARECLEF 14 \\ \midrule
Text type & online posts & clinical notes & clinical notes \\ 
Entity type & ADE & Disorder & Disorder \\ 
\# Documents & 1,250 & 298 & 433 \\ 
\# Tokens & 121K & 264K & 494K \\ 
\# Sentences & 7,597 & 18,767 & 34,618 \\
\# Mentions & 6,318 & 11,161 & 19,131 \\
\# Disc.M & 675 (10.6) & 1,090 (9.7) & 1,710 (8.9) \\ 
\midrule
Avg mention L. & 2.7 & 1.8 & 1.7 \\
Avg Disc.M L. & 3.5 & 2.6 & 2.5 \\
Avg interval L. & 3.3 & 3.0 & 3.2 \\ 
\midrule
\multicolumn{4}{c}{\bf Discontinuous Mentions} \\
\midrule
2 components & 650 (95.7) & 1,026 (94.3) & 1,574 (95.3) \\ 
3 components & \phantom{0}27 (\phantom{0}3.9) & \phantom{00}62 (\phantom{0}5.6) & \phantom{00}76 (\phantom{0}4.6) \\ 
4 components & \phantom{00}2 (\phantom{0}0.2) & \phantom{000}0 (\phantom{0}0.0) & \phantom{000}0 (\phantom{0}0.0) \\
\midrule
No overlap & \phantom{0}82 (12.0) & \phantom{0}582 (53.4) & \phantom{0}820 (49.6) \\
Overlap at left & 351 (51.6) & \phantom{0}376 (34.5) & \phantom{0}616 (37.3) \\
Overlap at right & 152 (22.3) & \phantom{0}102 (\phantom{0}9.3) & \phantom{0}170 (10.3) \\
Multiple overlaps & \phantom{0}94 (13.8) & \phantom{00}28 (\phantom{0}2.5) & \phantom{00}44 (\phantom{0}2.6) \\
\midrule
\multicolumn{4}{c}{\bf Continuous Mentions} \\
\midrule
Overlap & 326 (\phantom{0}5.7) & \phantom{0}157 (\phantom{0}1.5) & \phantom{0}228 (\phantom{0}1.3) \\
\bottomrule
\end{tabular}
\caption{The descriptive statistics of the datasets. ADE: adverse drug events; Disc.M: discontinuous mentions; Disc.M L.: discontinuous mention length, where intervals are not counted. Numbers in parentheses are the percentage of each category. Note that due to sentence segmentation issue, there are 13 and 64 mentions crossing multiple sentences in \SHARECLEF 2013 and \SHARECLEF 2014, respectively. We remove these mentions, as we frame the task as a sentence-level NER problem.~\label{table-discontinuous-ner-data-statistics}}
\end{table}

Descriptive statistics of these three datasets is listed in Table~\ref{table-discontinuous-ner-data-statistics}. On average, discontinuous mentions are longer than continuous mentions, because they consist of several components, and the intervals between different components make the total length of span even longer. Another important characteristic of discontinuous mentions is that they usually \emph{overlap}. That is, several mentions may share components that refer to the same body location (e.g., \textit{`muscle'} in \textit{`muscle pain and fatigue'}), or the same feeling (e.g., \textit{`Pain'} in \textit{`Pain in knee and foot'}). From this perspective, we also categorise discontinuous mentions into the following groups: 
\begin{itemize}
	\item No overlap: in such cases, the discontinuous mention can be intervened by severity indicators (e.g., \textit{`is  mildly'} in sentence \textit{`left atrium is mildly dilated'}), preposition (e.g., \textit{`on my'} in sentence \textit{`...rough on my stomach...'}) and so on. This category accounts for half of discontinuous mentions in the \SHARECLEF datasets but only 12\% in CADEC.
	\item Left overlap: the discontinuous mention shares one component with other mentions, and the shared component is at the beginning of the discontinuous mention. This is usually accompanied with coordination structure (e.g., the shared component \textit{`muscle'} in \textit{`muscle pain and fatigue'}). Conjunctions (e.g., \textit{`and'}, \textit{`or'}) are clear indicators of the coordination structure. However, clinical notes (\SHARECLEF datasets) are usually written by practitioners under time pressure. They often use commas or slashes rather than conjunctions. This category accounts for more than half of discontinuous mentions in CADEC and one third in \SHARECLEF.
	\item Right overlap: similar to left overlap, although the shared component is at the end. 
	For example, \textit{`hip/leg/foot pain'} contains three mentions that share the token \textit{`pain'}.
	\item Multi-overlap: the discontinuous mention shares multiple components with the others, which usually forms \emph{crossing compositions}. For example, the sentence \textit{`Joint and Muscle Pain / Stiffness'} contains four mentions: \textit{`Joint Pain'}, \textit{`Joint Stiffness'}, \textit{`Muscle Stiffness'} and \textit{`Muscle Pain'}, where each discontinuous mention share two components with the others.
\end{itemize} 

Although these three datasets -- CADEC, \SHARECLEF 2013 and \SHARECLEF 2014 -- share similar field (the subject matter of the content being discussed), the tenor (the participants in the discourse, their relationships to each other, and their purposes) of CADEC is very different from the \SHARECLEF datasets. Specially, laymen authors (CADEC) tend to use idioms or ungrammatical phrases to describe their feelings, whereas professional practitioners (\SHARECLEF) tend to use compact terms for efficient communications. This difference of tenor results in different features of mentions between these datasets. That is, the mentions in CADEC are overall longer than those in \SHARECLEF datasets, and larger ratio of discontinuous mentions in CADEC are involved in overlapping structure (Table~\ref{table-discontinuous-ner-data-statistics}). 

\section{Proposed Model~\label{section-discontinuous-ner-transition-model}}
We propose a transition-based model based on the shift-reduce parser~\citep{Watanabe:Sumita:ACL:2015,Lample:Ballesteros:NAACL:2016} that employs a \emph{stack} to store partially processed spans and a \emph{buffer} to store unprocessed tokens. The learning problem is then framed as: given the state of the parser, predict an action which is applied to change the state of the parser. This process is repeated until the parser reaches the end state, which is the stack and buffer are both empty. 

Similar to prior work~\citep{Metke-Jimenez:Karimi:BMDID:2016,Muis:Lu:EMNLP:2016} that first predict an intermediate representation of mentions, which are then decoded into the final mentions, our proposed transition-based model uses a sequence of actions as the intermediate representation (refer to Section~\ref{section-complex-ner}). 

The main difference between our model and the ones in~\citep{Watanabe:Sumita:ACL:2015,Lample:Ballesteros:NAACL:2016} is the set of transition actions. \citet{Watanabe:Sumita:ACL:2015} use SHIFT, REDUCE, UNARY, FINISH, and IDEA for the constituent parsing system. \citet{Lample:Ballesteros:NAACL:2016} use SHIFT, REDUCE, OUT for the flat NER system. Inspired by these models, we design a set of actions specifically for recognising discontinuous and overlapping structure. There is a total of six actions in our model:
\begin{itemize}
	\item SHIFT moves the first token from the buffer to the stack; it implies this token is part of an entity mention. 
	\item OUT pops the first token of the buffer, indicating it does not belong to any mention. 
	\item COMPLETE pops the top span of the stack, outputting it as an entity mention. If we are interested in multiple entity categories, we can extend this action to COMPLETE-$y$ which labels the mention with entity category $y$.
	\item REDUCE pops the top two spans $s_0$ and $s_1$ from the stack and concatenates them as a new span which is then pushed back to the stack.
	\item LEFT-REDUCE is similar to the REDUCE action, except that the span $s_1$ is kept in the stack. This action indicates the span $s_1$ is involved in multiple mentions. In other words, several mentions share $s_1$ which could be a single token or several tokens.
	\item RIGHT-REDUCE is the same as LEFT-REDUCE, except that $s_0$ is kept in the stack.
\end{itemize} 

\begin{figure}[tb] 
	\centering
	\includegraphics[width=0.8\linewidth]{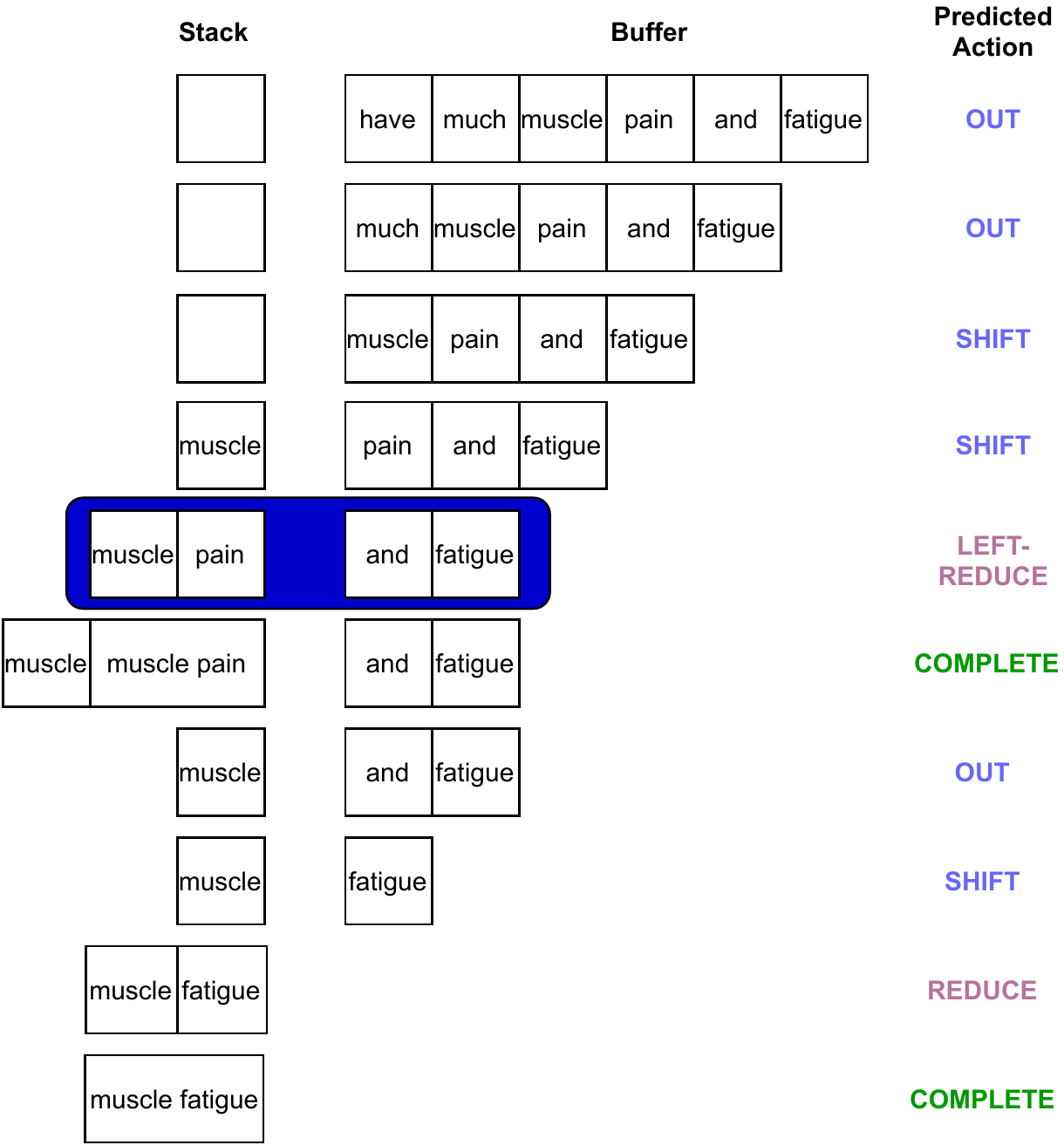}
	\caption{An example sequence of transitions. Given the states of stack and buffer (blue highlighted), as well as the previous actions, predict the next action (i.e., LEFT-REDUCE) which is then applied to change the states of stack and buffer.}
	\label{figure-discontinuous-ner-transition-example}
\end{figure}

Figure~\ref{figure-discontinuous-ner-transition-example} shows an example of how the proposed parser recognises entity mentions from a sentence. 

\subsection{Representation of the parser state~\label{section-discontinuous-ner-configuration-representation}}

Given a sequence of $N$ tokens, we first run a bi-directional \lstm~\citep{Graves:Mohamed:ICASSP:2013} to derive the contextual representation of each token. Specifically, for the $i$-th token in the sequence, its representation can be denoted as:
\[
\dvector{\tilde{c_i}} = \left [ \overrightarrow{\lstm} (\dvector{t_0}, \dots, \dvector{t_{i}}); \overleftarrow{\lstm} (\dvector{t_{i}}, \dots, \dvector{t_{N - 1}}) \right ],
\]
where $\mathbf{t_i}$ is the concatenation of the embeddings for the $i$-th token, its character level representation learned using a CNN network~\citep{Ma:Hovy:ACL:2016}. Pretrained contextual word representations have shown its usefulness on improving various NLP tasks. Here, we can also concatenate pretrained contextual word representations using ELMo~\citep{Peters:Neumann:NAACL:2018} with $\mathbf{\tilde{c_i}}$, resulting in:
\begin{equation}
\dvector{c_i} = \big[ \dvector{\tilde{c_i}}; \ELMO_i \big],
\end{equation}
where $\ELMO_i$ is the output representation of pretrained \ELMO models (frozen) for the $i$-th token. These token representations $\mathbf{c}$ are directly used to represent tokens in the buffer. 

Following the work in~\citep{Dyer:Ballesteros:ACL:2015}, we use \textsc{StackLSTM} to represent spans in the stack. That is, if a token is moved from the buffer to the stack, its representation is learned using:
\[
\dvector{s_0} = \textsc{StackLSTM} (\dvector{s_D} \dots \dvector{s_1};\dvector{c}_{\text{SHIFT}}),
\]
where $D$ is the number of spans in the stack. Once REDUCE related actions are applied, we use a multi-layer perceptron to learn the representation of the concatenated span. For example, the REDUCE action takes the representation of the top two spans in the stack: $\dvector{s_0}$ and $\dvector{s_1}$, and produces a new span representation:
\begin{equation}
    \dvector{\Tilde{s}} = \dmatric{W}^T \big[ \dvector{s_0}; \dvector{s_1} \big] + \dvector{b},
\end{equation}
where $\dmatric{W}$ and $\dvector{b}$ denote the parameters for the composition function. 
The new span representation $\dvector{\Tilde{s}}$ is pushed back to the stack to replace the original two spans: $\dvector{s_0}$ and $\dvector{s_1}$. 

\subsection{Capturing discontinuous dependencies}
We hypothesise that the interactions between spans in the stack and tokens in the buffer are important factors in recognising discontinuous mentions. Considering the example in Figure~\ref{figure-discontinuous-ner-transition-example}, a span in the stack (e.g., \textit{`muscle'}) may need to combine with a future token in the buffer (e.g., \textit{`fatigue'}). To capture this interaction, we use multiplicative attention~\citep{Luong:Pham:EMNLP:2015} to let the span in the stack $\dvector{s_i}$ learn which token in the buffer to attend, and thus a weighted sum of the representation of tokens in the buffer $\dmatric{B}$:
\begin{equation}
\begin{split}
\dvector{s^a_i} & = \textsc{Attention} (\dvector{s_i}, \dmatric{B}, \dmatric{B}) \\
& = \textsc{softmax} (\dvector{s_i^TW^a_iB})\dmatric{B}.
\end{split}
\end{equation}
We use distinct $\dmatric{W}_i^a$ for spans in different positions $\dvector{s_i}$ separately. 

\subsection{Selecting an action}
Finally, we build the parser representation as the concatenation of the representation of top three spans from the stack ($\dvector{s_0}, \dvector{s_1}, \dvector{s_2}$) and its attended representation ($\dvector{s^a_0}$, $\dvector{s^a_1}$, $\dvector{s^a_2}$), as well as the representation of the previous action $\dvector{a}$, which is learned using a simple unidirectional \lstm. If there are less than 3 spans in the stack or no previous action, we use randomly initialised vectors $\dvector{s_{empty}}$ or $\dvector{a_{empty}}$ to replace the corresponding vector. This parser representation is used as input for the final softmax prediction layer to select the next action. 

Note that, given one parser state, not all types of actions are valid. For example, if the stack does not contain any span, only SHIFT and OUT actions are valid because all other actions involve popping spans from the stack. We employ hard constraints that we only select the most likely action from valid actions. 

\section{Experimental Results~\label{section-discontinuous-ner-experiment-results}}
To evaluate the effectiveness of our proposed model, we run experiments on previous described three datasets: CADEC, \SHARECLEF 2013 and \SHARECLEF 2014, and compare the effectiveness of our model against several baselines. 

\subsection{Baseline models}
We choose one flat NER model which is strong at recognising continuous mentions, and two discontinuous NER models as our baseline models: 

\paragraph{Flat model}
To train the flat model on our datasets, we use an off-the-shelf framework: \FLAIR~\citep{Akbik:Blythe:COLING:2018}, which achieves the state-of-the-art performance on \CONLL 2003 dataset. Recall that the flat model cannot be directly applied to datasets containing discontinuous mentions. Following the practice in~\citep{Stanovsky:Gruhl:EACL:2017}, we replace the discontinuous mention with the shortest span that fully covers it, and merge overlapping mentions into a single mention that covers both. Different from~\citep{Stanovsky:Gruhl:EACL:2017}, we apply these changes only on the training set, and not on the development and the test sets. 

\paragraph{BIO extension model}
The original implementation in~\citep{Metke-Jimenez:Karimi:BMDID:2016} used a CRF model with manually designed features. We report their results on CADEC in Table~\ref{table-discontinuous-ner-main-result} and re-implement a \bilstm-CRF-\ELMO model using their tag schema (denoted as `BIO extension' in Table~\ref{table-discontinuous-ner-main-result}). 

\paragraph{Graph-based model}
The original paper of~\citep{Muis:Lu:EMNLP:2016} only reported the evaluation results on sentences which contain at least one discontinuous mention. We use their implementation to train the model and report evaluation results on the whole test set (denoted as `Graph' in Table~\ref{table-discontinuous-ner-main-result}). We argue that it is important to see how a discontinuous NER model works not only on the discontinuous mentions but also on all the mentions, especially since, in real datasets, the ratio of discontinuous mentions cannot be made a priori. 

\subsection{Experimental setup}
As CADEC does not have an official train-test split, we follow~\citep{Metke-Jimenez:Karimi:BMDID:2016} and randomly assign 70\% of the posts as the training set, 15\% as the development set, and the remaining posts as the test set. The train-test splits of \SHARECLEF 13 and 14 are both from their corresponding shared task settings, except that we randomly select 10\% of documents from each training set as the development set. The original \SHARECLEF 14 task focuses on template filling of disorder attributes: that is, given a disorder mention, recognise the attribute from its context. In this work, we use its mention annotations and frame the task as a discontinuous NER task. Micro average strict match $F_1$ score is used to evaluate the effectiveness of the model. The trained model which is most effective on the development set, measured using the $F_1$ score, is used to evaluate the test set. All experiments are repeated five times using different random seeds and averaged results are reported. 

\subsection{Results}
When evaluated on the whole test set, our model outperforms three baseline models, as well as over previous reported results in the literature, in terms of recall and $F_1$ scores (Table~\ref{table-discontinuous-ner-main-result}). 

\begin{table}[tb] 
	\begin{small}
		\setlength{\tabcolsep}{3pt} 
		\centering
		\begin{tabular}{cr  cccc  cccc  cccc}
			\toprule
			&&& \multicolumn{3}{c}{\bf CADEC} && \multicolumn{3}{c}{\bf ShARe 2013} && \multicolumn{3}{c}{\bf ShARe 2014} \\ 
			\cmidrule{4-6}\cmidrule{8-10} \cmidrule{12-14}
			&\bf Model && P & R & F && P & R & F && P & R & F \\ 
			\cmidrule{2-2} \cmidrule{4-6}\cmidrule{8-10} \cmidrule{12-14}
			& \cite{Metke-Jimenez:Karimi:BMDID:2016} && 64.4 & 56.5 & 60.2 && -- & -- & -- && -- & -- & -- \\
			& \cite{Tang:Hu:Wireless:2018} && 67.8 & 64.9 & 66.3 && -- & -- & -- && -- & -- & -- \\
			& \cite{Tang:Wu:CLEF:2013} && -- & -- & -- && 80.0 & 70.6 & 75.0 && -- & -- & -- \\
			\hline
			& Flat && 65.3 & 58.5 & 61.8 && 78.5 & 66.6 & 72.0 && 76.2 & 76.7 & 76.5 \\ 
			& BIO extension && 68.7 & 66.1 & 67.4 && 77.0 & 72.9 & 74.9 && 74.9 & 78.5 & 76.6 \\ 
			& Graph && \bf 72.1 & 48.4 & 58.0 && \bf 83.9  & 60.4  & 70.3 && \bf 79.1 & 70.7 & 74.7 \\ 
			& Ours && 68.9 & \bf 69.0 & \bf 69.0 && 80.5 & \bf 75.0 & \bf 77.7 && 78.1 & \bf 81.2 & \bf 79.6 \\ 
			\bottomrule 
		\end{tabular}
		\caption{Evaluation results in terms of precision (P), recall (R) and $F_1$ score (F).~\label{table-discontinuous-ner-main-result}}
	\end{small}
\end{table}

The graph-based model achieves highest precision, but with substantially lower recall, therefore obtaining lowest $F_1$ scores. In contrast, our model improves recall over flat and BIO extension models as well as previously reported results, without sacrificing precision. This results in more balanced precision and recall. Improved recall is especially encouraging for our motivating pharmacovigilance and medical record summarization applications, where recall is at least as important as precision. Note that most of these previous models are tailored for specific entity categories, and utilise domain-specific resources. For example, \citep{Tang:Wu:CLEF:2013}, the best-performing system participated in the \SHARECLEF 2013 shared task, utilise several external domain-specific resources, such as \textsc{MetaMap}, \textsc{cTAKES} and \textsc{UMLS}. We avoid these tailored resources in our model. We argue that this makes our model more generic and robust, especially since we apply the hyper-parameters tuned on \textsc{CADEC} directly to \SHARECLEF datasets and obtain similar improvements with respect to the benchmarks. 

\paragraph{Effectiveness on recognising discontinuous mentions}
Recall that only 10\% of mentions in these three datasets are discontinuous. To evaluate the effectiveness of our proposed model on recognising discontinuous mentions, we follow the evaluation approach in~\citep{Muis:Lu:EMNLP:2016} where we construct a subset of test set where only sentences with at least one discontinuous mention are included (Table~\ref{table-discontinuous-ner-result-on-sentence-with-disc}). We also report the evaluation results when only discontinuous mentions are considered (Table~\ref{table-discontinuous-ner-result-on-disc}). Note that sentences in the former setting usually contain continuous mentions as well, including those involved in overlapping structure (e.g., \textit{`muscle pain'} in the sentence \textit{`muscle pain and fatigue'}). Therefore, the flat model, which cannot predict any discontinuous mentions, still achieves $38\%$ $F_1$ on average when evaluated on these sentences with at least one discontinuous mention, but fails to recognise discontinuous mentions. 

\begin{table}[tb] 
	\begin{small}
		\centering
		\begin{tabular}{cr  cccc  cccc  cccc}
			\toprule
			&&& \multicolumn{3}{c}{\bf CADEC} && \multicolumn{3}{c}{\bf \SHARECLEF 2013} && \multicolumn{3}{c}{\bf \SHARECLEF 2014} \\ 
			\cmidrule{4-6}\cmidrule{8-10} \cmidrule{12-14}
			& \bf Model && P & R & F && P & R & F && P & R & F \\ 
			\cmidrule{2-2} \cmidrule{4-6}\cmidrule{8-10} \cmidrule{12-14}
			& Flat && 50.2 & 36.7 & 42.4 && 43.5 & 28.1 & 34.2 && 41.5 & 31.9 & 36.0 \\ 
			& BIO extension && 63.8 & 52.0 & 57.3 && 51.8 & 39.5 & 44.8 && 37.5 & 38.4 & 37.9 \\ 
			& Graph && \bf 69.5 & 43.2 & 53.3 && \bf 82.3 & 47.4 & 60.2 && 60.0 & 52.8 & 56.2 \\ 
			& Ours && 66.5 & \bf 64.3 & \bf 65.4 && 70.5 & \bf 56.8 & \bf 62.9 && \bf 61.9 & \bf 64.5 & \bf 63.1 \\ 
			\bottomrule 
		\end{tabular}
		\caption{Evaluation results on sentences that contain at least one discontinuous mention.~\label{table-discontinuous-ner-result-on-sentence-with-disc}}
	\end{small}
\end{table}

\begin{table}[tb] 
	\begin{small}
		\centering
		\begin{tabular}{cr  cccc  cccc  cccc}
			\toprule
			&&& \multicolumn{3}{c}{\bf CADEC} && \multicolumn{3}{c}{\bf \SHARECLEF 2013} && \multicolumn{3}{c}{\bf \SHARECLEF 2014} \\ 
			\cmidrule{4-6}\cmidrule{8-10} \cmidrule{12-14}
			& \bf Model && P & R & F && P & R & F && P & R & F \\ 
			\cmidrule{2-2} \cmidrule{4-6}\cmidrule{8-10} \cmidrule{12-14}
			& Flat && 0 & 0 & 0 && 0 & 0 & 0 && 0 & 0 & 0 \\ 
			& BIO extension && 5.8 & 1.0 & 1.8 && 39.7 & 12.3 & 18.8 && 8.8 & 4.5 & 6.0 \\ 
			& Graph && \bf 60.8 & 14.8 & 23.9 && 78.4 & 36.6 & 50.0 && 42.7 & 39.5 & 41.1 \\ 
			& Ours && 41.2 & \bf 35.1 & \bf 37.9 && \bf 78.5 & \bf 39.4 & \bf 52.5 && \bf 56.1 & \bf 43.8 & \bf 49.2 \\ 
			\bottomrule 
		\end{tabular}
		\caption{Evaluation results on discontinuous mentions only.~\label{table-discontinuous-ner-result-on-disc}}
	\end{small}
\end{table}

Our model again achieves the highest $F_1$ and recall in all three datasets under both settings. The comparison between these two evaluation results also shows the necessity of comprehensive evaluation settings. The BIO extension model outperforms the graph-based model in terms of $F_1$ score on CADEC, when evaluated on sentences with discontinuous mentions. However, it achieves only $1.8$ $F_1$ when evaluated on discontinuous mentions only. The main reason is that most of discontinuous mentions in CADEC are involved in overlapping structure ($88\%$, cf. Table~\ref{table-discontinuous-ner-data-statistics}), and the BIO extension model is better than the graph-based model at recognising these continuous mentions. On \SHARECLEF 2013 and 2014, where the portion of discontinuous mentions involved in overlapping is much less than on CADEC, the graph-based model clearly outperforms BIO extension model in both evaluation settings. 

Graph based model again achieves highest precision on all three datasets. It also outperforms BIO extension model on \SHARECLEF 2013 and 2014 in terms of $F_1$ score, but not on CADEC. Graph based model employs lots of handcrafted features for clinical notes (e.g., note type, section name, word-level semantic category extracted from \textsc{UMLS}). These handcrafted features usually lead to high precision but are not general enough to recall unseen mentions. In addition, they usually do not generalise well in different types of text (i.e., online posts in CADEC). In contrast, we avoid these handcrafted features in our model. 

\section{Analysis}
\label{section-discontinuous-ner-analysis}

\subsection{Impact of mention and interval length}
Discontinuous mentions usually represent compositional concepts that consist of multiple components. Therefore, these mentions are usually longer than continuous mentions (Table~\ref{table-discontinuous-ner-data-statistics}). In addition, intervals between components make the total length of span involved even longer. Previous work shows that flat NER performance degrades when applied on long mentions~\citep{Augenstein:Das:SemEval:2017,Xu:Jiang:ACL:2017,Lange:Dai:IberLEF:2020}. 

\begin{figure}[tb] 
	\begin{subfigure}{.48\textwidth}
		\centering
		\includegraphics[width=.98\linewidth]{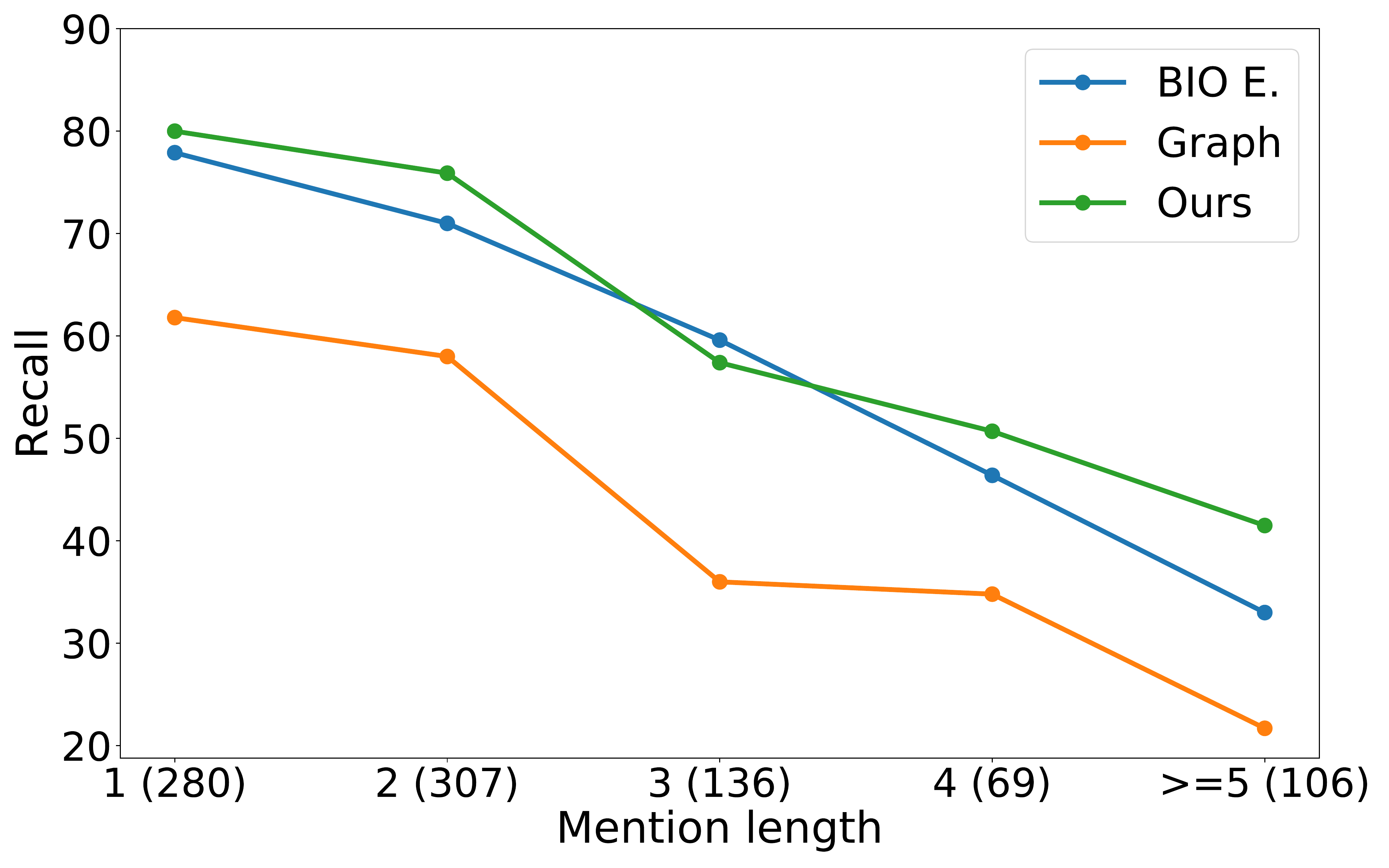}
		\caption{CADEC}
	\end{subfigure}%
	\begin{subfigure}{.48\textwidth}
		\centering
		\includegraphics[width=.98\linewidth]{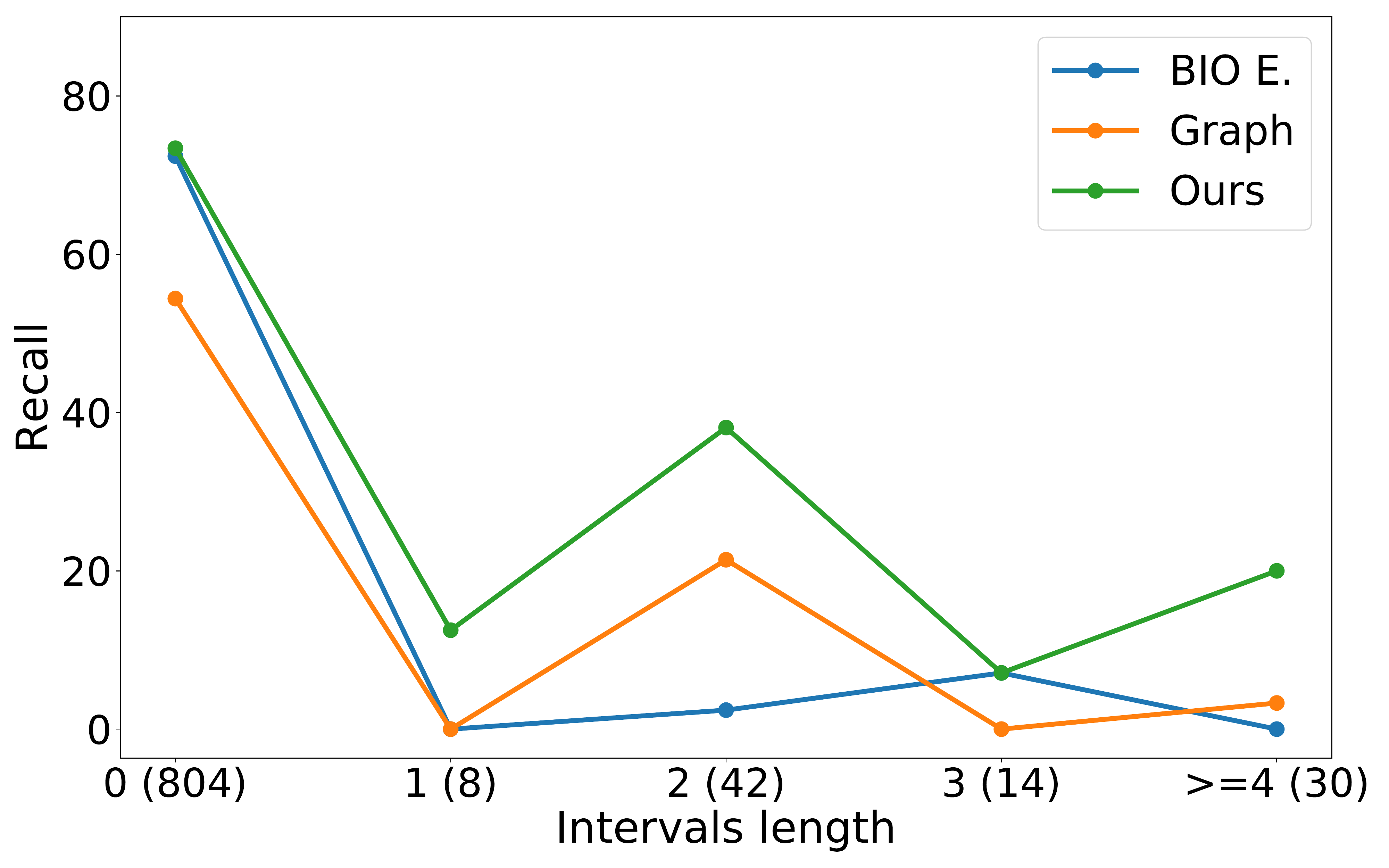}
		\caption{CADEC}
	\end{subfigure}%
	
	\begin{subfigure}{.48\textwidth}
		\centering
		\includegraphics[width=.98\linewidth]{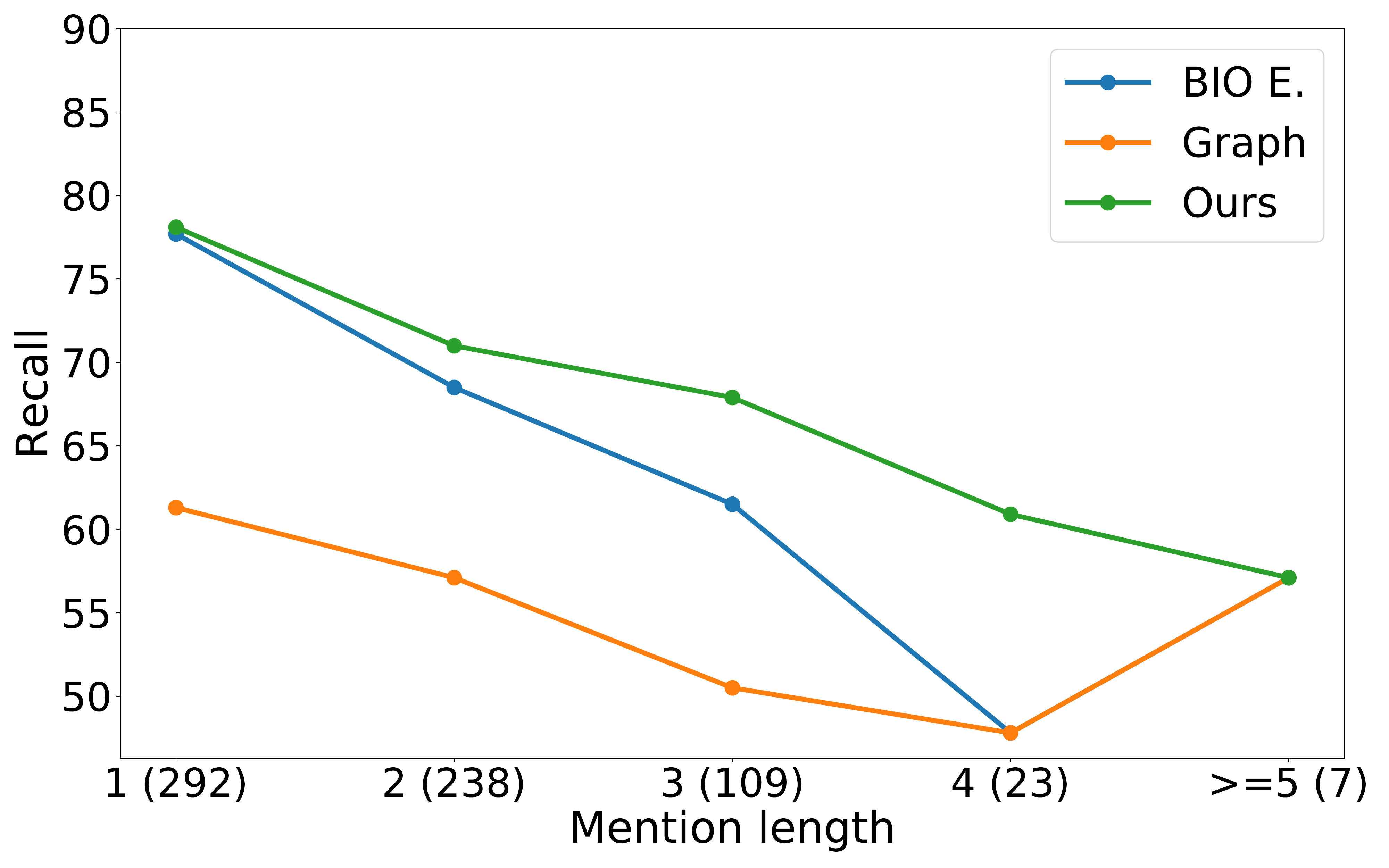}
		\caption{\SHARECLEF 2013}
	\end{subfigure}%
	\begin{subfigure}{.48\textwidth}
		\centering
		\includegraphics[width=.98\linewidth]{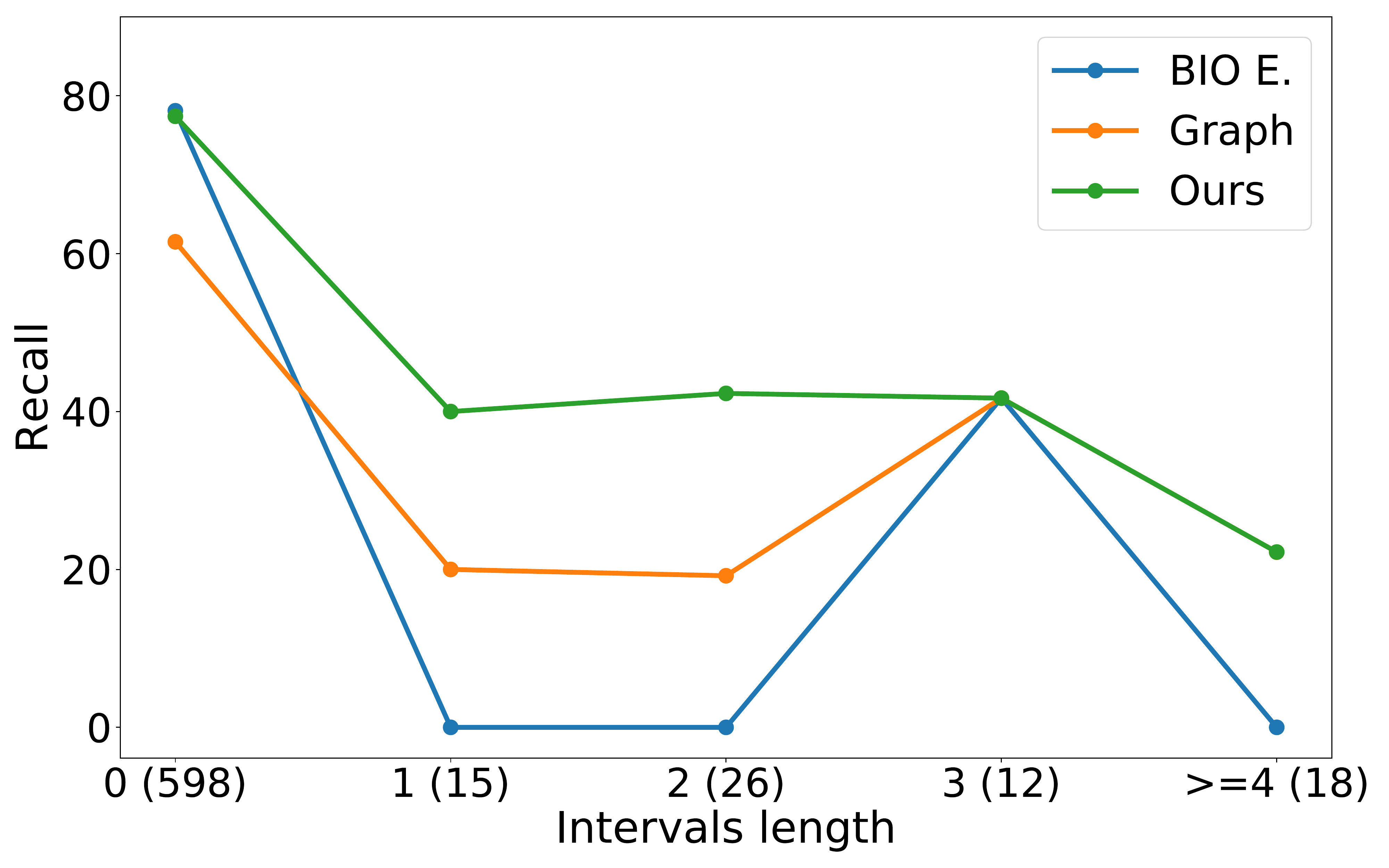}
		\caption{\SHARECLEF 2013}
	\end{subfigure}%
	
	\begin{subfigure}{.48\textwidth}
		\centering
		\includegraphics[width=.98\linewidth]{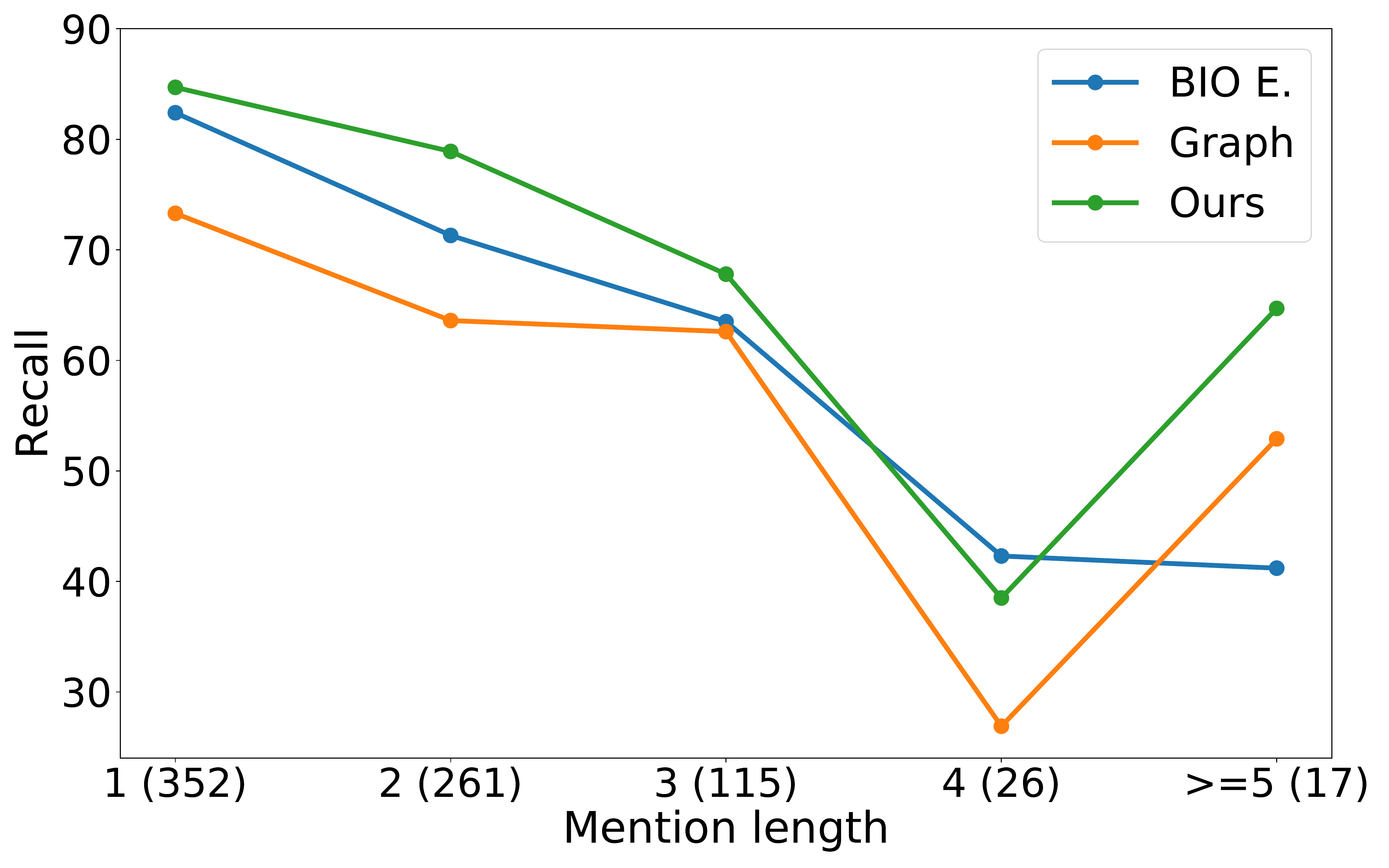}
		\caption{\SHARECLEF 2014}
		\label{fig:sfig1}
	\end{subfigure}
	\begin{subfigure}{.48\textwidth}
		\centering
		\includegraphics[width=.98\linewidth]{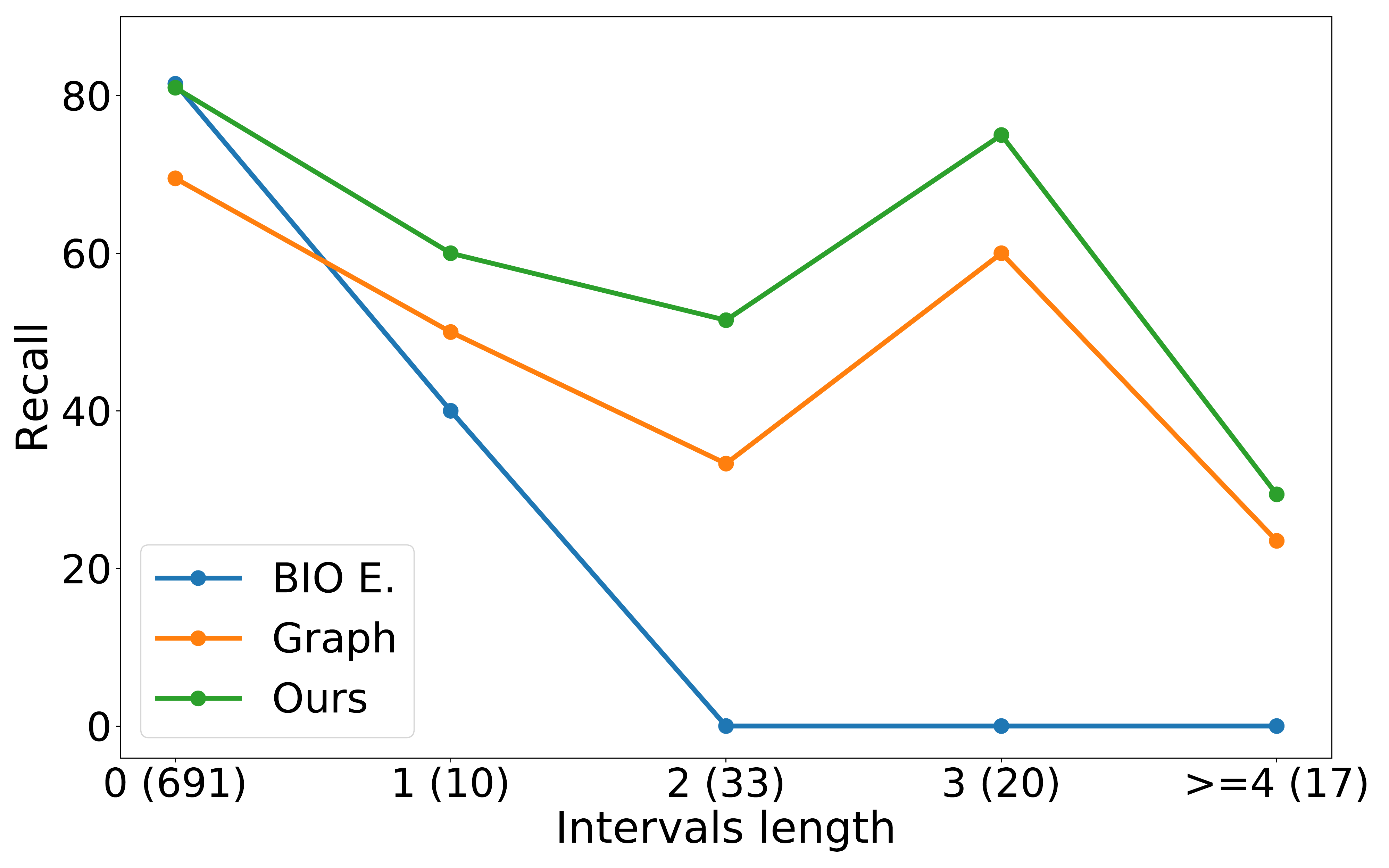}
		\caption{\SHARECLEF 2014}
	\end{subfigure}
	\caption{The impact of mention length and interval length on recall. Mentions with interval length of zero are continuous mentions. Numbers in parentheses are the number of gold mentions.~\label{figure-discontinuous-ner-length-vs-recall}}
\end{figure}

We experiment to measure the ability of different models on recalling mentions of different lengths, and to observe the impact of interval lengths. We find that the recall of all models decreases with the increase of mention length in general (Figure~\ref{figure-discontinuous-ner-length-vs-recall} (a -- c)), which is similar to previous observations in the literature on flat mentions~\citet{Lange:Dai:IberLEF:2020}. However, the impact of interval length is not straightforward. Mentions with very short interval lengths are as difficult as those with very long interval lengths to be recognised (Figure~\ref{figure-discontinuous-ner-length-vs-recall} (d -- f)). On CADEC, discontinuous mentions with interval length of two are easiest to be recognised (Figure~\ref{figure-discontinuous-ner-length-vs-recall} (d)), whereas those with interval length of three are easiest on \SHARECLEF 2013 and 2014. We hypothesise this also relates to annotation inconsistency, because very short intervals may be overlooked by annotators. 

Our method achieves highest recall among all models in most settings. This demonstrates our model is effective to recognise both continuous and discontinuous mentions with various lengths. In contrast, the BIO extension model is only strong at recalling continuous mentions (outperforming the graph-based model), but fails on discontinuous mentions (interval lengths larger than zero). 

\subsection{Impact of overlapping structure}
Another characteristic of discontinuous mentions is that they usually {\em overlap} (Section~\ref{section-discontinuous-ner-task-data}). Previous study shows that the intervals between components can be problematic for coordination boundary detection (\cite{Ficler:Goldberg:EMNLP:2016}). Conversely, we want to observe whether the overlapping structure may help or hinder discontinuous entity recognition. We categorise discontinuous mentions into different subsets, described in Section~\ref{section-discontinuous-ner-task-data}, and measure the effectiveness of different discontinuous NER models on each category. 

\begin{table}[tb] 
		\centering
		\begin{tabular}{ r r c  cc c cc c cc}
			\toprule
			& & & \multicolumn{2}{c}{\bf CADEC} && \multicolumn{2}{c}{\bf ShARe 2013} && \multicolumn{2}{c}{\bf ShARe 2014} \\
			\cmidrule{4-5}\cmidrule{7-8}\cmidrule{10-11}
			& \bf Model & & \# & F && \# & F && \# & F \\
			\cmidrule{2-2}\cmidrule{4-5}\cmidrule{7-8}\cmidrule{10-11}
			
			\multirow{3}{*}{No overlap} & BIO extension && \multirow{3}{*}{9} & 0.0 && \multirow{3}{*}{41} & 7.5 && \multirow{3}{*}{39} & 0.0 \\
			& Graph && & 0.0 &&  & 32.1 &&  & 45.2 \\
			& Ours &&  & 0.0 && & \bf 36.1 && & \bf 57.1 \\
			\hline
			
			\multirow{3}{*}{Overlap at left} & BIO extension && \multirow{3}{*}{54} & 6.0 && \multirow{3}{*}{11} & 25.0 && \multirow{3}{*}{30} & 15.7 \\
			& Graph && & 9.2 &&  & \bf 45.5 &&  & 37.7 \\
			& Ours &&  & \bf 28.6 && & 33.3 && & \bf 49.2 \\
			\hline
			
			\multirow{3}{*}{Overlap at right} & BIO extension && \multirow{3}{*}{16} & 0.0 && \multirow{3}{*}{19} & 0.0 && \multirow{3}{*}{5} & 0.0 \\
			& Graph && & \bf 45.2 &&  & \bf 21.4 &&  & 0.0 \\
			& Ours &&  & 29.3 && & 13.3 &&  & 0.0 \\
			\hline
			
			\multirow{3}{*}{Multiple overlaps} & BIO extension && \multirow{3}{*}{15} & 0.0 && \multirow{3}{*}{0} & -- && \multirow{3}{*}{6} & 0.0 \\
			& Graph && & 0.0 &&  & -- &&  & 0.0 \\
			& Ours &&  & 0.0 && & -- && & 0.0 \\
			\bottomrule
		\end{tabular}
		\caption{Evaluation results on different categories of discontinuous mentions. `\#' columns show the number of gold discontinuous mentions in development set of each category.\label{discontinuous-ner-table-impact-of-overlapping}}
\end{table}

From Table~\ref{discontinuous-ner-table-impact-of-overlapping}, we find that our model achieves better results on discontinuous mentions belonging to \emph{No overlap} category on \SHARECLEF 2013 and 2014, and \emph{Overlap at left} category on CADEC and \SHARECLEF 2014. Note that \emph{No overlap} category accounts for half of discontinuous mentions in \SHARECLEF 2013 and 2014, whereas \emph{Overlap at left} accounts for half in CADEC (Table~\ref{table-discontinuous-ner-data-statistics}). Graph-based model achieves better results on \emph{Overlap at right} category. On the \emph{Multiple overlaps} category, no models is effective~\footnote{Our model cannot recognise all mentions belonging to this category in theory. For example, if two mentions overlap at both the left and the right, our model can predict only one of them.}, which emphasises the challenges of dealing with this syntactic phenomena (examples can be found in Section~\ref{section-discontinuous-ner-example-predictions}). We note, however, the portion of discontinuous mentions belonging to this category is very small in all three datasets. 

Although our model achieves better results on \emph{No overlap} category on \SHARECLEF 2013 and 2014, it does not predict correctly any discontinuous mention belonging to this category on CADEC. The ineffectiveness of our model, as well as other discontinuous NER models, on CADEC \emph{No overlap} category can be attributed to two reasons: 1) the number of discontinuous mentions belonging to this category in CADEC is small (around 12\%), rending the learning process more difficult. 2) the gold annotations belonging to this category are inconsistent from a linguistic perspective. For example, severity indicators are annotated as the interval of the discontinuous mention sometimes, but not often. Note that this may be reasonable from a medical perspective, as some symptoms are roughly grouped together no matter their severity, whereas some symptoms are linked to different concepts based on their severity and severe adverse drug reactions are especially on the radar. 

\subsection{Example predictions~\label{section-discontinuous-ner-example-predictions}}
We find that previous models often fail to identify discontinuous mentions that involve long and overlapping spans. For example, the sentence \textit{‘Severe joint pain in the shoulders and knees.’} contains two mentions: \textit{‘Severe joint pain in the shoulders’} and \textit{‘Severe joint pain in the knees’}. Graph-based model does not identify any mention from this sentence, resulting in a low recall. The BIO extension model predicts most of  these tags (8 out of 9) correctly, but fails to decode into correct mentions (predict \textit{‘Severe joint pain in the’}, resulting in a false positive, while it misses \textit{‘Severe joint pain in the shoulders’}). In contrast, our model correctly identifies both of these two mentions. 

\begin{table}[tb] 
	\begin{small}
		\centering
		\begin{tabular}{r|p{12cm}}
			\hline
			Sentence & Walked like that for about six months with increasing pain , especially in right thigh which felt . \\
			Gold mentions & 1. \textbf{pain in right thigh} \\
			Predictions & 1. \st{pain} [BIO extension] \\
			& No prediction [Graph] \\
			& 1. \st{increasing pain} [Ours] \\ \hline
			Sentence & Stated with joint and pain and muscle weakness , depression , fatigue and cramps . \\
			Gold mentions & 1. \textbf{joint pain}; 2. \textit{muscle weakness}; 3. \textit{depression}; 4. \textit{fatigue}; 5. \textit{cramps} \\
			Predictions & 1. \st{joint and pain}; 2. \textit{muscle weakness}; 3. \textit{depression}; 4. \textit{fatigue}; 5. \textit{cramps} [BIO extension] \\
			& No prediction [Graph] \\
			& 1. \textit{muscle weakness}; 2. \st{pain weakness}; 3. \st{joint weakness}; 4. \textit{depression}; 5. \textit{fatigue}; 6. \textit{cramps} [Ours] \\ \hline
			Sentence & stopped taking them 4 years ago and still suffer terrible muscle pain and wasting . \\
			Gold mentions & 1. \textit{muscle pain}; 2. \textbf{muscle wasting} \\
			Predictions & 1. \st{terrible muscle pain}; 2. \st{wasting} [BIO extension] \\
			& No prediction [Graph] \\
			& 1. \textit{muscle pain}; 2. \st{wasting} [Ours] \\ \hline
			Sentence & Then I sated having hip / leg / foot pain and numbness . \\
			Gold mentions & 1. \textbf{hip pain}; 2. \textbf{leg pain}; 3. \textit{foot pain}; 4. \textit{numbness} \\
			Predictions & 1. \textit{foot pain}; 2. \st{pain}; 3. \textit{numbness} [BIO extension] \\
			& No prediction [Graph] \\
			& 1. \st{stated}; 2. \textit{hip pain}; 3. \textit{leg pain}; 4. \textit{foot pain}; 5. \textit{numbness} [Ours] \\ \hline
			Sentence & Severe joint pain in the shoulders and knees . \\
			Gold mentions & 1. \textit{Severe joint pain in the shoulders}; 2. \textit{Severe joint pain in the knees} \\
			Predictions & 1. \st{Severe joint pain in the}; 2. \textit{Severe joint pain in the knees} [BIO extension] \\
			& None [Graph] \\
			& 1. \textit{Severe joint pain in the shoulders}; 2. \textit{Severe joint pain in the knees} [Ours] \\ \hline
			Sentence & Joint and Muscle Pain / Stiffness .  \\
			Gold mentions & 1. \textbf{Joint pain}; 2. \textbf{Muscle Stiffness}; 3. \textit{Muscle Pain}; 4. \textbf{Joint Stiffness} \\
			Predictions & 1. \st{Joint}; 2. \textit{Muscle Pain}; 3. \st{Stiffness} [BIO extension] \\
			& 1. \textit{Joint pain}; 2. \st{Muscle Pain / Stiffness}; 3. \st{Stiffness} [Graph] \\
			& 1. \textit{Joint pain}; 2. \textit{Muscle Pain}; 3. \st{Stiffness} [Ours] \\ \hline
		\end{tabular}
	\end{small}
	\caption{Example sentences involving discontinuous entity mentions and predictions using different methods. These examples are taken from CADEC. Gold discontinuous mentions are highlighted in bold. We cross out the incorrect predictions (false positives) for easy understanding.~\label{table-discontinuous-ner-example-predictions}}
\end{table}

Another observation is that no model can fully recognise mentions which form crossing compositions. For example, the sentence \textit{‘Joint and Muscle Pain / Stiffness’} contains four mentions: \textit{‘Joint Pain’}, \textit{‘Joint Stiffness’}, \textit{‘Muscle Stiffness’} and \textit{‘Muscle Pain’}, all of which share multiple components with the others. Our model correctly predicts \textit{‘Joint Pain’} and \textit{‘Muscle Pain’}, but it mistakenly predicts \textit{‘Stiffness’} itself as a mention (Table~\ref{table-discontinuous-ner-example-predictions}). 

\subsection{Ablation studies}

To empirically evaluate the importance of attention and ELMo components, we test the performance of model variants where attention and ELMo are removed separately on CADEC and \SHARECLEF 2013 datasets. 

\begin{table}[tb] 
	\begin{small}
		\centering
		\begin{tabular}{ lc  c c c c c}
			\toprule
			&& \multicolumn{2}{c}{\bf CADEC} && \multicolumn{2}{c}{\bf \SHARECLEF 2013} \\ \cmidrule{3-4}\cmidrule{6-7}
			\bf Model&& All & Subset w. Disc. && All & Subset w. Disc. \\ \cmidrule{1-1}\cmidrule{3-4}\cmidrule{6-7}
			Full && \bf 68.4 & \bf 65.4 && \bf 77.2 &  \bf 64.3 \\ 
			-Attention && 68.4 & 63.3 && 76.8 & 62.3 \\ 
			- ELMo && 66.7 & 62.2 && 75.2 & 60.9 \\ \bottomrule
		\end{tabular}
		\caption{Ablation study to estimate the contribution of attention and ELMo components.~\label{table-discontinuous-ner-ablation-study}}
	\end{small}
\end{table}

The results in Table~\ref{table-discontinuous-ner-ablation-study} show that removing attention hurts the performance when evaluated on sentences with discontinuous mentions (\textit{w. Disc.} columns), but have little impact on the complete test set where continuous mentions are prevalent. Since we use BiLSTM to derive contextual representation for each token, we believe these contextual representations are effective at recognising continuous mentions, but have trouble identifying intervals within discontinuous mentions. Attention mechanism, via allowing tokens interacting with distant tokens, can capture additional discontinuous dependencies which are not captured by BiLSTM. In terms of the ELMo component, we find that it contributes approximately 2 $F_1$ score when evaluated on the complete test set and around 4 $F_1$ when evaluated on sentences with discontinuous mentions, demonstrating the usefulness of pretrained word representations. 

\section{Summary}
Recognising discontinuous mentions that represent compositional concepts is important for downstream applications such as pharmacovigilance. We propose an end-to-end transition-based model for discontinuous NER. It makes use of specialised actions and attention mechanism to determine whether a span is the component of a discontinuous mention or not. We evaluate our model on three biomedical datasets with a substantial number of discontinuous mentions and demonstrate that our model can effectively recognise discontinuous mentions without sacrificing the accuracy on continuous mentions. Analysis also suggests that our model is better than existing discontinuous NER models at handling long mentions, resulting in higher recall. 

%% file: src/ch7-conclusion.tex
Recognising biomedical names from scholarly articles, clinical notes, and social media data is a fundamental NLP task that can benefit many downstream biomedical NLP and information retrieval applications. However, due to the unique characteristics of biomedical names and the stylistic variation in biomedical language---used by biomedical researchers, practitioners, patients and other participants---biomedical NER needs to solve challenges comparatively less studied in the generic domain NER applications.

In this thesis, we first identified challenges of applying standard sequence tagger to recognise biomedical names. Although sequence tagging techniques have demonstrated their effectiveness in generic domain NER, achieving state-of-the-art performance in many benchmarks, they suffer from three problems when being applied in the biomedical domain:
\begin{itemize}
    \item Biomedical names may consist of non-consecutive spans and they may overlap with each other. The main reason of this complex structure in biomedical names is that many biomedical concepts are compositional. For example, a symptom description may consist of several components: body location, severity indicator, and general feeling, and these components may locate far away from each other.
    \item Training of neural based sequence taggers usually requires large training set, which is difficult to obtain in the biomedical domain. Annotating biomedical NER datasets usually requires domain-knowledge, and sometimes even unlabelled data are unavailable due to legal reasons.
    \item State-of-the-art sequence taggers are usually enhanced by language representation models pre-trained on large set of generic domain unlabelled data. Domain shift between these out-of-domain pre-training data and the target biomedical data usually results in a performance drop.
\end{itemize}

Targeting these three problems, we explored the corresponding research directions.

We proposed a transition-based model for discontinuous NER. The proposed model is an end-to-end model with generic neural encoding that allows us to leverage specialised actions and attention mechanism to determine whether a span is the component of a discontinuous mention or not. We evaluate our model on three biomedical datasets with a substantial number of discontinuous mentions and demonstrate that our model can effectively recognise discontinuous mentions without sacrificing the accuracy on continuous mentions.

We designed several easy to use data augmentation methods for the NER task: Label-wise token replacement, Synonym replacement, Mention replacement and Shuffle within segments. These augmentations do not rely on any externally trained models, such as machine translation models or syntactic parsing models, which are by themselves difficult to train in a low-resource domain-specific scenario. Through experiments on two biomedical datasets, we show that simple data augmentation can improve performance even over strong baselines, where large scale pre-trained language representation models are used. We leave the exploration of combining these data augmentation methods with other NER models, such as the transition-based model we proposed, for future work.

We analysed different aspects of similarity between domains, and employed cost-effective measures to quantify domain similarity. We demonstrated that these measures are good predictors of the usefulness of pre-trained language representation models on downstream NER task. We find that human intuition favour field (the subject matter being discussed) over tenor (the participants of the discourse and their purpose) when they select in-domain pre-training data. Results suggest that this intuition may be unreliable when the target data set locates in the intersection of several domains.

Based on the discoveries presented in this thesis, we see two future directions worth exploring. 
\begin{itemize}
    \item The first one is on incorporating existing biomedical knowledge base. In Chapter~\ref{chapter-data-augmntation}, we explored the data augmentation methods that make use of the original training set and a generic lexical database of English. Similar augmentation methods can also be applied to biomedical knowledge base. For example, a biomedical concept---defined using CUI in UMLS---can have several aliases from various vocabularies. These aliases can be used to create augmented sentences in the data augmentation settings. In Chapter~\ref{chapter-select-pretraining-data}, we investigated the impact of pre-training data on the effectiveness of pre-trained language representation models, where we pre-train models using only the unlabelled text. We believe more sophisticated pre-training tasks based on biomedical knowledge base can create pre-trained models that capture both language and biomedical knowledge. 
    \item The second direction is on investigating the impact of NER performance on downstream tasks, such as entity linking, relation extraction, or biomedical literature search. In Chapter~\ref{chapter-discontinuous-ner}, we showed that our proposed model can effectively recognise discontinuous biomedical names without sacrificing the performance of continuous ones. It worth investing how this improvement can benefit downstream tasks whose results are directly presented to end users.
\end{itemize}